\documentclass[twoside,11pt]{article}

\usepackage{blindtext}

\usepackage[abbrvbib, preprint]{jmlr2e}

\usepackage{enumitem}
\usepackage{float}

\usepackage{verbatim}
\usepackage{multirow}

\usepackage{xr}
\usepackage{balance}
\usepackage{xspace}

\usepackage{url}
\usepackage{amsmath}
\usepackage{amssymb}

\usepackage{mathtools}

\usepackage{dsfont}
\usepackage{upgreek}
\usepackage{bbm}			

\usepackage{graphicx}
\usepackage{epstopdf}
\graphicspath{ {./figures/} } 
\usepackage{grffile}
\usepackage{tikz}
\usepackage{float}
\usepackage{svg}

\usepackage{booktabs} 

\usepackage[labelfont=bf]{caption}
\usepackage{subfigure}

\usepackage{mathtools}
\usepackage{relsize}  

\usepackage{makecell}

\usepackage{expl3}

\newcommand{\Sec}[1]		{Sec.\,\ref{#1}}

\newcommand{\Appendix}[1]		{Appendix\,\ref{#1}}		
\newcommand{\Fig}[1]		{Fig.\,\ref{#1}}

\newcommand{\Eq}[1]			{Eq.\,\ref{#1}}
\newcommand{\Expr}[1]			{Expr.\,\ref{#1}}

\newcommand{\Tab}[1]		{Tab.\,\ref{#1}}
\newcommand{\Alg}[1]		{Alg.\,\ref{#1}}
\newcommand{\Theorem}[1]{Theorem\,\ref{#1}}

\newcommand{\Lemma}[1]{Lemma\,\ref{#1}}

\newcommand{\Assumption}[1]{Assumption\,\ref{#1}}

\newcommand{\Problem}[1]{Problem\,\ref{#1}}	
\newcommand{\vs}   			{vs.\@\xspace}
\newcommand{\ie}   			{i.e.\@\xspace}
\newcommand{\eg}   			{e.g.\@\xspace}
\newcommand{\etc}   		{etc.\xspace}
\newcommand{\wrt}   		{w.r.t.\@\xspace}
\newcommand{\iid}   		{iid\@\xspace}

\newcommand{\one}       {\mathbf{1}}     
\newcommand{\ones}[1]   {\one_{#1}}
\newcommand{\zero}      {\mathbf{0}}
\newcommand{\zeros}[1]   {\zero_{#1}}
\newcommand{\bigO} 		  {\mathcal{O}} 
\newcommand{\ind}       {\mathds{1}}
\newcommand{\Ind}[1]    { \ind{\{#1\}} }

\newcommand{\diag}    {\operatorname{diag}}
\newcommand{\spankern}    {\operatorname{span}}
\newcommand{\Vect}     {\operatorname{vec}}
\newcommand{\Block}    {\operatorname{block}}
\newcommand{\real}      {\mathbb{R}}

\newcommand{\pdfuncs}		{pdfs\xspace}
\newcommand{\pdf}		    {pdf\xspace}
\newcommand{\pdfs}	 	  {pdfs\xspace}

\newcommand{\Hilbert}      {\mathbb{H}}
\newcommand{\Ghilbert}      {\mathbb{G}}
\newcommand{\R}      {\mathbb{R}}
\newcommand{\N}      {\mathbb{N}}
\newcommand{\C}      {\mathbb{C}}
\newcommand{\PE}{P\!\!E}  

\newcommand{\KernelFunc} {\textup{K}} 

\newcommand{\conj}{\star}
\newcommand{\fdivpre} 	{$\phi$-divergence\xspace}
\newcommand{\fdivsymb} 	{\phi}
\newcommand{\fdiv} 	{$\phi$-divergence\xspace}
\newcommand{\fdivs} {$\phi$-divergences\xspace}
\newcommand{\KLdiv} {KL-divergence\xspace}
\newcommand{\PEdiv} {$\chi^2$-divergence\xspace}

\newcommand{\LRE}	  {LRE\xspace}

\newcommand{\VVRKHS}	  {VV-RKHS\xspace}
\newcommand{\RKHS}	  {RKHS\xspace}

\newcommand{\inlinetitle}[2]  {\vspace{4pt}\noindent\textbf{\emph{#1}{#2}}}

\renewcommand*{\top}{{\mkern-1.5mu\mathsf{T}}}

\newcommand{\Expecnv}{\mathbb{E}_{p_v(x)}}

\newcommand{\Expecjointv}{\mathbb{E}_{p_{z,v}}}

\newcommand{\ExpecJointv}[1]{\mathbb{E}_{p_{z,v}}[#1]}
\newcommand{\q}{q}
\newcommand{\ExpecAlpha}[1]{\mathbb{E}_{p^{\alpha}(y)}[#1]}
\newcommand{\ExpecAlphav}[1]{\mathbb{E}_{p_v^{\alpha}(y)}[#1]}
\newcommand{\Expecalphav}{\mathbb{E}_{p_v^{\alpha}(y)}}

\newcommand{\ExpecA}[1]{\mathbb{E}_{\q(x')}[#1]}
\newcommand{\Expecav}{\mathbb{E}_{\q_v(x')}}

\newcommand{\ExpecAm}[2]{\mathbb{E}_{\q_{#2}(x')}[#1]}

\newcommand{\norm}[1]{\left\lVert#1\right\rVert}

\newcommand{\dott}[2]{\langle #1,\,#2 \rangle}
\newcommand{\dotH}[2]{\langle #1,#2 \rangle_{\Hilbert}}

\newcommand{\Lpc}{\mathlarger{\mathcal{L}}}
\newcommand{\Gramm}{\mathlarger{\mathcal{K}}}

\newcommand{\abs}[1]{\left|#1\right|}
\newcommand{\Npre}{n}
\newcommand{\Npost}{n'}

\newcommand{\setprev}{X_v}
\newcommand{\setpostv}{{X'_v}}

\newcommand{\X}{{\mathcal{X}}}
\newcommand{\nghood}{\text{ng}} 
\newcommand{\Gdims}{N} 
\newcommand{\G}{G} 

\newcommand{\ThetaG}{\mathbf{\Theta}}

\newcommand{\hthetanode}[1]{\hat{\theta}_{#1}}

\newcommand{\eigmax}[1]{e_{\max}\!\left({#1}\right)}

\newcommand{\mydef}{=} 
\newcommand{\id}{I}

\newcommand{\likelihood}{likelihood\xspace}
\newcommand{\upright}[1]{\mathrm{#1}}

\newcommand{\CO}        {\Sigma}
\newcommand{\rhofunction}        {\varrho}

\newcommand{\padded}[1] {\,#1\,}
\newcommand{\void}[1] {\padded{\cdot}}

\newcommand{\mycomplement}{{\scriptscriptstyle\complement}}

\newtheorem{assumption1}{Assumption}

\usepackage{booktabs,arydshln}
\makeatletter
\def\adl@drawiv#1#2#3{%
        \hskip.5\tabcolsep
        \xleaders#3{#2.5\@tempdimb #1{1}#2.5\@tempdimb}%
                #2\z@ plus1fil minus1fil\relax
        \hskip.5\tabcolsep}
\newcommand{\cmidruledashed}[1]{%
  \noalign{\vskip\aboverulesep
           \global\let\@dashdrawstore\adl@draw
           \global\let\adl@draw\adl@drawiv}
  \cdashline{#1}%
	\noalign{\vskip-\belowrulesep}
	\noalign{\vskip-\belowrulesep}
}
\makeatother

\makeatletter
\newcommand{\pushright}[1]{\ifmeasuring@#1\else\omit\hfill$\displaystyle#1$\fi\ignorespaces}
\newcommand{\pushleft}[1]{\ifmeasuring@#1\else\omit$\displaystyle#1$\hfill\fi\ignorespaces}
\makeatother

\makeatletter
\let\KV@Gin@trim@old\KV@Gin@trim
\define@key{Gin}{trim}{%
  \begingroup
  \edef\x{\endgroup
    \noexpand\setkeys{Gin}{trim@old={#1}}%
  }\x
}
\makeatother
\makeatletter
\let\KV@Gin@viewport@old\KV@Gin@viewport
\define@key{Gin}{viewport}{%
  \begingroup
  \edef\x{\endgroup
    \noexpand\setkeys{Gin}{viewport@old={#1}}%
  }\x
}
\makeatother

\DeclareMathOperator*{\argmax}{\arg\!\max}
\DeclareMathOperator*{\argmin}{\arg\!\min}

\usepackage[normalem]{ulem} 
\newcommand{\RMV}[1] 	  {{\color{orange} \sout{#1}}}

\newcommand{\VOID}[1] 				{}

\newcommand{\mySqBullet}		{\raisebox{0.25em}{{\scriptsize$_\blacksquare$}}}

\newcounter{marginNoteCounter}

\graphicspath{ {../figures/} } 
\usepackage{algorithm}
\usepackage{algorithmic}
\usepackage{xfp}

\usetikzlibrary{decorations.pathreplacing, calligraphy}

\usepackage{lastpage}
\jmlrheading{xx}{202x}{1-\pageref{LastPage}}{xx/xx; Revised xx/xx}{xx/xx}{xx-xxxx}{Alejandro de la Concha, Nicolas Vayatis, and Argyris Kalogeratos}

\ShortHeadings{Collaborative likelihood-ratio Estimation}
{A. de la Concha, N. Vayatis, and A. Kalogeratos}
\firstpageno{1}

\begin{document}

\title{Collaborative likelihood-ratio estimation over graphs}
\author{\name Alejandro de la Concha \email alejandro.de\_la\_concha\_duarte@ens-paris-saclay.fr \\
\AND
\name Nicolas Vayatis \email nicolas.vayatis@ens-paris-saclay.fr 
 \\
\AND
\name Argyris Kalogeratos \email argyris.kalogeratos@ens-paris-saclay.fr\\
\addr{{Université  Paris-Saclay, ENS Paris-Saclay, CNRS,  Centre Borelli, France}}}

\editor{}

\maketitle

\begin{abstract}
Assuming we have \iid observations from two unknown probability density functions (\pdfuncs), $p$ and $q$, the \emph{likelihood-ratio estimation} (\LRE) is an elegant approach to compare the two \pdfuncs only by relying on the available data. In this paper, we introduce the first -to the best of our knowledge- graph-based extension of this problem, which reads as follows: Suppose each node $v$ of a fixed graph has access to observations coming from two unknown \emph{node-specific} \pdfuncs, $p_v$ and $q_v$, and the goal is to estimate
for each node the likelihood-ratio between both \pdfuncs by also taking into account the information provided by the graph structure. The node-level estimation tasks are supposed to exhibit similarities conveyed by the graph, which suggests that the nodes could collaborate to solve them more efficiently. We develop this idea in a concrete non-parametric method that we call \emph{Graph-based Relative Unconstrained Least-squares Importance Fitting} (GRULSIF). 
We derive convergence rates for our collaborative approach that highlights the role played by variables such as 
the number of available observations per node, the size of the graph, and how accurately the graph structure encodes the similarity between tasks. These theoretical results explicit the situations where collaborative estimation effectively leads to 
an improvement in performance 
compared to solving each problem independently.
Finally, in a series of experiments, we illustrate how GRULSIF infers the likelihood-ratios at the nodes of the graph more accurately compared to state-of-the art \LRE methods, which would operate independently at each node, and we also verify that the behavior of GRULSIF is aligned with our previous theoretical analysis. 
\end{abstract}

\begin{keywords}Unsupervised learning, f-divergence, likelihood-ratio estimation, kernel methods, graph regularization, multitask learning. 
\end{keywords}

\section{Introduction}{\label{sec:intro}}

A number of computational tasks and practical questions can be stated in the form of a comparison between probabilistic models. 
Quantifying the distance between two probability measures is an old problem in Statistics, which has led to the development of a unified framework for diverse problems, such as the maximum likelihood estimation, dimension reduction, two-sample hypothesis testing, and outlier detection \citep{Csiszar2004,Liese2006,Sugiyama2012,Basseville2013,Rubenstein2019}. A concept of major role in these advancements has been the \emph{\fdivpre}\footnote{In literature, \fdivpre is commonly met as $f$-divergence; here the choice is due to notation clarity.}, which is a similarity measure between two probability measures. Widely-used examples are Kullback–Leibler's \KLdiv \citep{Kullback59} and Pearson's \PEdiv \citep{Pearson1900}.

In Machine Learning, there is an increasing interest in the \fdivpre estimation relying only on data 
from two probability distributions
with probability density functions (\pdfs)
$p$ and $\q$ \citep{Nguyen2007,Perez-Cruz2008,Qing2009,Poczos2011,Sugiyama2012,Krishnamurthy2014,Moon2014a,Moon2014b,Rubenstein2019}. An effective strategy is the likelihood-ratio estimation (\LRE) that directly infers the real function $r : \mathcal{X} \rightarrow \R$, called \emph{likelihood-ratio} (or \emph{density-ratio}), $r(x)=\frac{q(x)}{p(x)}$
, via non-parametric techniques. The \fdivpre is then approximated in terms of empirical means of quantities defined in terms of $r(x)$ \citep{Nguyen2007,Nguyen2010,Sugiyama2012,Yamada2011}.

The study of the interplay between \fdiv and \LRE has led to several applications in fields where the likelihood-ratio is a central quantity, such as Transfer Learning, Hypothesis Testing, and Change-point Detection. Transfer Learning relaxes the classical hypothesis that the training and the test datasets are samples of the same distribution, and relies instead on importance weighting 
 that trains a predictive model by focusing on training losses that are weighted according to the test-over-training likelihood-ratio, $r(x)=\frac{p_{\text{test}}}{p_{\text{train}}}$ \citep{Huang2006,Sugiyama2007,Yamada2013_RULSIFjournal,Lu2023}.
In Hypothesis Testing, statistical tests based on \fdivpre have been proposed when there is no prior knowledge of the form of $q$ and $p$, and only two data samples from both distributions are available. The test statistic takes the form of an approximated \fdivpre via empirical averages on an estimated likelihood-ratio \citep{Sugiyama2011_LSTST,Sugiyama2012,Yamada2013_RULSIFjournal}. A similar approach is followed in non-parametric Change-point Detection, where the goal is to detect the moment at which a time-series changes its behavior from $p(x)$ to $q(x)$ , which are both of unknown form \citep{Liu2013,Ferrari2023}. 

The common point of the workflow in these applications is that they comprise two distinct stages: first comes the \LRE stage where the likelihood-ratio $r$ is estimated as best as possible, and then in the second stage this quantity is used to compute proper application-specific scores for solving the task of interest (\eg a weighting function or a test statistic). This explains the broad interest of the research community in generic \LRE approaches that can implement the first of the above stages. Important to note that, to the best of our knowledge, the existing research is focused on estimating a single likelihood-ratio. Nevertheless, modern challenges are posed in cases where there are multiple local data sources aiming at solving several similar tasks in applications like those mentioned earlier. An intriguing question emerges: how can these local data sources collaborate to solve their respective task with higher precision than if they were to tackle the task by their own? 

The main contribution of this paper focuses on the collaborative \LRE: in our framework local data sources are represented as nodes in a fixed graph, and we specifically intend to compare two node-specific \pdfuncs of each node $v$, namely $p_v$ and $\q_v$, using the \iid observations recorded at $v$ and the graph structure. The novelty of the framework lies on the fact that \emph{it allows nodes to estimate collaboratively the functions of interest}, $r_v$, instead of independently each node by itself. Our fundamental hypothesis is that the graph structure conveys valuable information about the `similarity' between nodes, \ie how similar are expected to be the estimation problems at any two nodes. 

There are many possible ways to define the notion of node similarity, yet leading to different estimation problems. For instance, node similarity can be assumed in the input space $\mathcal{X}$, by assuming that adjacent nodes will have similar observations (this is termed as \emph{graph signal smoothness} in Graph Signal Processing \citep{Ortega2018}); or in the output space, by seeking a single model that has similar outputs for similar inputs (as in Semi-supervised Learning \citep{Belkin2004, SSL2006,Cabannes2021}. In our setting, multiple models need to be learned at the same time, and hence one can assume node similarity in the model space, and more specifically in a metric functional space $\mathcal{F}$: two models, $f_u$, $f_v\in \mathcal{F}$, which solve the individual estimation problems at two adjacent nodes $u$ and $v$, are now expected to be close to each other with respect to a proper distance defined in $\mathcal{F}$. This is the notion of \emph{task similarity} appearing in Multitask Learning. Indeed, graph-based multitasking has brought improvements in the generalization performance for supervised problems \citep{Maurer2006,Maurer2006b,Yousefi2018,Zhang2021}. Such results, and the attractiveness of distributed data processing, have motivated various applications and the design of special optimization schemes \citep{Nassif2020a,Nassif2020b,Nassif2020c,Zhang2021}.

The collaborative non-parametric LRE framework, proposed for the graph-based setting, is novel. It assumes that the likelihood-ratios, $r_u$ and $r_v$, are elements of a Reproducing Kernel Hilbert Space $\Hilbert$ that is shared among all nodes, and if $u$ and $v$ are connected in the graph, then $r_u$ and $r_v$ are expected to be also close to each other in $\Hilbert$. Our approach capitalizes over the advances in Multitask Learning and Kernel Methods, achieves better performance than solving the problem at each graph node independently, and the gains become more evident the fewer data are available. This is validated experimentally, using synthetic experiments, against state-of-the-art non-parametric \LRE methods that operate at each node without taking into account the graph. Moreover, our approach is distributed, hence limits by design the data sharing
: nodes share actual data only with a central server in order to build a global dictionary, and then adjacent nodes exchange over the evaluations of their relative estimates $r_v$'s, which compare their associated $p_v$'s and $q_v$'s. An important by-product of our optimization scheme is the POOL variant that reduces GRULSIF by neutralizing the graph component (\ie solving multiple independent \LRE tasks), yet manages to produce much better estimates than state-of-the-art methods addressing the independent \LRE tasks. Therefore, POOL is a valuable result per se, which also helps as a baseline so that we can precise the share of the gain that comes from incorporating a graph into the \LRE process.

\inlinetitle{Organization of the paper}{.}~The rest of the paper is organized as follows. In \Sec{sec:background}, we introduce the problem, and we provide the building tools for our \LRE framework. 
In \Sec{sec:GRULSIF}, we present the main technical contribution of the paper, the collaborative \LRE method that we call Graph-based Relative Unconstrained Least Squares Importance Fitting (GRULSIF). Then, in \Sec{sec:theory}, we provide theoretical guarantees on the excess risk to illustrate the performance of GRULSIF and its sensibility to relevant parameters such as the number of available observations per node, the size of the graph, and the prior information provided by the graph structure. We discuss in \Sec{sec:practical_implementation} the elements allowing the  efficient implementation of GRULSIF in practice. Finally, in \Sec{sec:experiments}, we illustrate the performance of the proposed method in experiments involving synthetic data. 

\section{Preliminaries and problem statement} \label{sec:background}

\inlinetitle{General notations}{.} Let $a_i$ be the $i$-th entry of a vector $a$; when the vector is itself indexed by 
$j$, 
we refer to its $i$-th entry by $a_{j,i}$. 
$A_{ij}$ denotes the entry at the $i$-th row and $j$-th column of a matrix $A$, and $A_{i,:}$ is its $i$-th row. We denote by $\eigmax{A}$, the maximum eigenvalue of a given matrix A, and 
$A^{\dag}$ denotes the pseudoinverse. Given two matrices $A$ and $B$, we denote by $A\otimes B$ their Kronecker product.  
We denote by $\Block(A_1,...,A_n)$  a block diagonal matrix where each block corresponds to one of the square matrices $A_1,...,A_n$. $\Vect(a_1,..,a_n)$ denotes the concatenation of the input vectors $a_1,...,a_{n}$ in a single vector. Also, $\ones{M}$ represents the vector with $M$ ones (resp. $\zeros{M}$ for zeros), $\id_{M}$ is the $M \times M$ identity matrix, and $\Ind{\text{condition}} \in \{0,1\}$ is the indicator matrix. %
The Euclidean norm and the dot product are denoted by $\norm{\cdot}$ and $\dott{\cdot}{\cdot}$. When those are endowed to a functional space $\mathcal{F}$, we write them as 
$\norm{\cdot}_{\mathcal{F}}$ and $\dott{\cdot}{\cdot}_{\mathcal{F}}$, respectively.

Concerning graph structure, a fixed undirected weighted graph $\G=(V,E,W)$ is defined by the set $V$ containing $N$ nodes, and the set of edges $E$. Throughout the rest of the presentation, we suppose that the edges are positive-weighted and undirected, and that the nodes have no self-loops, \ie the entries of its weight matrix $W \in \real^{\Gdims\times \Gdims}$ are such that $W_{uu}=0$, $\forall u \in V$, and $W_{uv}=W_{vu} \geq 0$. The set with the neighbors of node $v$ is $\nghood(v) = \{u\,:\,W_{uv} \neq 0\}$, and in an undirected graph it holds $u \in \nghood(v) \Leftrightarrow v \in \nghood(u)$. Finally, the degree of node $v$ is denoted by $d_v$. %
In the rest, composite objects (vectors, matrices, sets, \etc) that refer to all the nodes of a graph, are denoted in bold font.

\subsection{Problem statement}{\label{sec:problem_statement}}

Let a fixed undirected and positive-weighted graph $\G=(V,E,W)$, and suppose each node $v \in V$ has \iid observations from two unknown \pdfs: $\Npre_v$ observations from $p_v$, and respectively $\Npost_v$ others from $\q_v$. The two sets are:

\begin{equation}\label{eq:windows}
\left\{
    \begin{array}{ll}
        \mathbf{X}\!\! &= \{\mathbf{X}_v\}_{v\in V}= \big\{\{x_{v,1},...,x_{v,\Npre_v}\}\big\}_{v \in V}, \ \ \ 
				\forall v,i: \ \ x_{v,i} \,\overset{\text{\iid}}{\sim}\, p_v;%
          \\
        \mathbf{X}' &= \{\mathbf{X}'_v\}_{v\in V} = \big\{\{x'_{v,1},...,x'_{v,\Npost_v}\}\big\}_{v \in V}, \ \ \ \forall v,i: \ \ x'_{v,i} \,\overset{\text{\iid}}{\sim}\, \q_v.
    \end{array}
    \right.
\end{equation}%

\begin{figure}[t!]
  \centering

\includegraphics[width=0.95\linewidth]{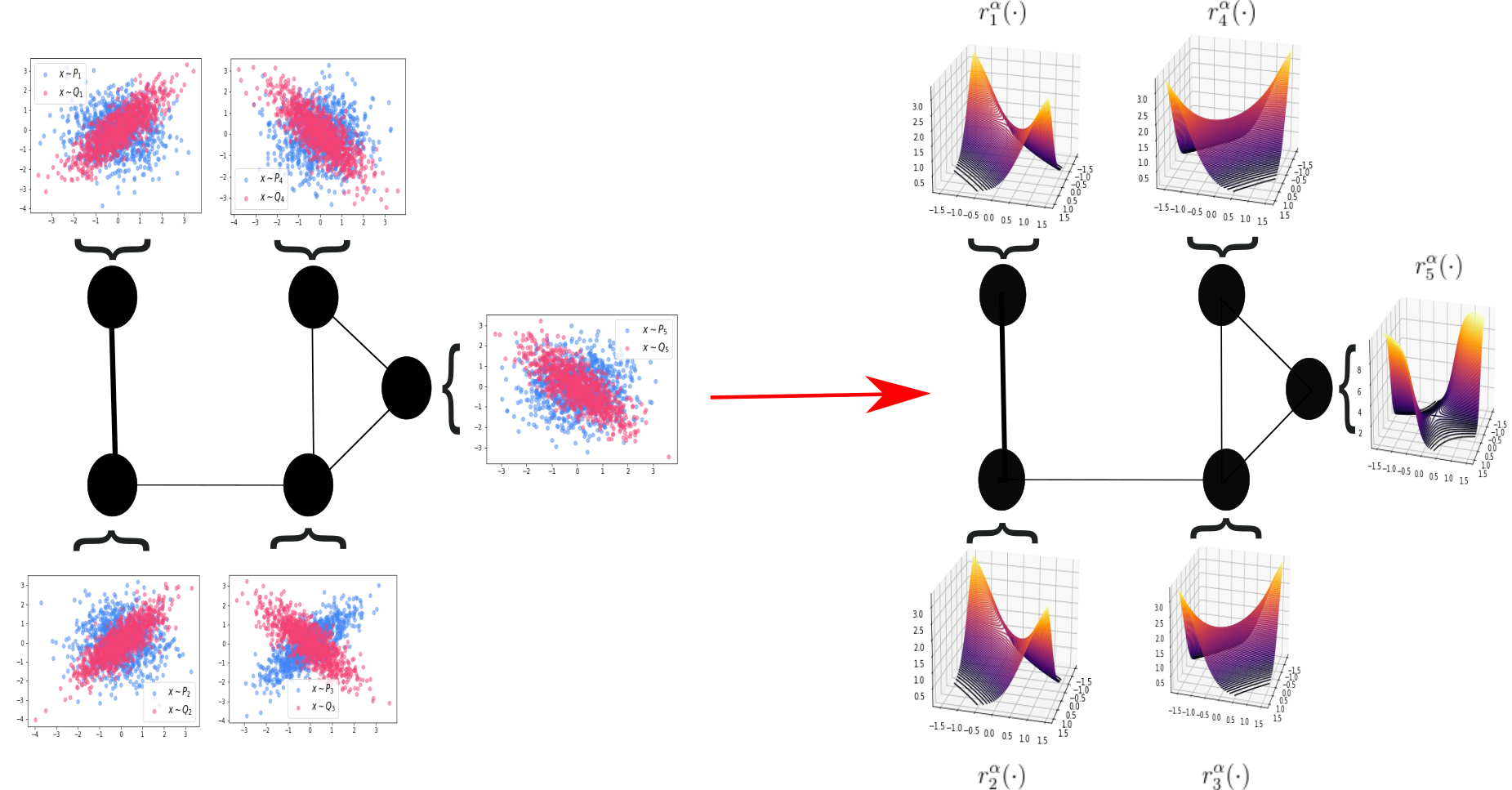}%
\vspace{-0.5mm}
    \caption{\small \textbf{Likelihood-ratio estimation over a graph $G$.} Simple example of the problem addressed by GRULSIF. Given different data points from two probabilistic models $p_v$ (blue) and $q_v$ (pink) at each node of $G$ (left-side figure), we aim to estimate the associated relative-likelihood ratio $r_v^{\alpha}$ (right-side figure) in a collaborative and distributed manner. In this example, the input domain of the data is $\mathcal{X}=\R^2$, and it is easy to see how any given $x\in\X$ gets essentially mapped to the graph signal $\mathbf{r}^{\alpha}(x) = (r^{\alpha}_1(x),...,r^{\alpha}_\Gdims(x))^{\top}$. 
    }
\label{fig:GRULSIF}
\end{figure}

Our goal is to quantify how different $p_v$ and $\q_v$ are for each node by taking into account the structural information provided by the graph. The idea is to formalize this problem as a likelihood-ratio estimation (\LRE) problem where each node $v$ learns a node-specific model $f_v$ approximating the likelihood-ratio between $p_v$ and $\q_v$, while capitalizing over the similarity of $v$ to its adjacent nodes. \Fig{fig:GRULSIF} presents an insightful visualization of the problem.

This problem setting can be relevant to many complex real-world applications. For instance, each graph node may correspond to an agent or a sensor collecting local data at the different locations where they lie (\eg meteorological data, air pollution sensor networks, medical surveys across geographic regions). Then, one's interest may be to estimate a likelihood-ratio at each location, which can be subsequently used in tasks such as those we mentioned earlier, \ie Hypothesis Testing, Transfer Learning, Change-point Detection, \etc For example, under an event as a pollution peak, a two-sample test could be used to understand the global air quality state measured by a sensor network, as well as the situation at each individual site based on what the associated sensor measures, and relate the effect of the event at those locations. Similarly, in the context of Transfer Learning, a network of hospitals may desire to update their diagnostic algorithms to an unexpected change in the behavior of a disease. For such applications, collaborative \LRE could play a crucial role by ensuring that \emph{the heterogeneity among what each agent or sensor observes locally will not get diluted by any sort of global data aggregation}. Under certain conditions, this approach can enhance the performance in both the estimation of the $r_v$'s and the related application tasks, as compared to approaches that ignore the interdependence between the nodes.  

\subsection{Important notions for LRE}

\inlinetitle{\fdiv
}{.} %
It is a similarity measure between two probability models that are described by the \pdfuncs $p$ and $\q$, over the input space $\X \subset \R^d$. Formally this is expressed as: 
\begin{equation}{\label{def:f_divergence}}
 \mathcal{D}_\phi(P \Vert Q) 
		= \int \,\phi\!\left(\frac{q(x)}{p(x)}\right)\!p(x) dx 
   = \int \!\phi\!\left(r^*\right)\!(x) p(x) dx ,
\end{equation}
where $P$ and $Q$ are the measures with which the pdfs $p$ and $q$ are associated; recall, that for any arbitrary function $\phi$ with input domain $\X$, it holds $\int\!\phi(x) p(x)dx = \int\! \phi(x) dP$, which is a measure-theoretic definition of expectation. For \Eq{def:f_divergence}, the interesting cases are those where the likelihood-ratio $r^*(x)=\frac{\q(x)}{p(x)}$ can be defined, and $\fdivsymb: \real \rightarrow \real$ is a convex and semi-continuous real function such that $\fdivsymb(1) = 0$ \citep{Csiszar1967}. Notably, for $z\in \real$, when $\fdivsymb(z)=\log(z)$ we recover the well-known \KLdiv \citep{Kullback59}, and when $\fdivsymb(z)=\frac{1}{2}(z-1)^2$ we get Pearson's \PEdiv \citep{Pearson1900}.

\inlinetitle{Relative likelihood-ratio}{.} %
One issue of the usual likelihood-ratio, $r(x)=\frac{\q(x)}{p(x)}$, is that it may be an unbounded function, hence its non-parametric estimation may be an ill-posed problem. %
In this work, we employ an elegant and flexible alternative, the $\alpha$-\emph{relative likelihood-ratio function} \citep{Yamada2011}, $r^{\alpha} : \mathcal{X} \rightarrow \R$ (note: $\alpha$ is only an index in $r^\alpha$):
\begin{equation}\label{eq:rel-density-ratio}
\,r^{\alpha}(x)=\frac{\q(x)}{(1-\alpha)p(x)+\alpha \q(x)},\quad \text{for any } 0 \leq \alpha < 1.
\end{equation}
Here, $\q$ is compared against $p^{\alpha}(x) = \alpha \q(x)+(1-\alpha)p(x)$, which is 
the convex combination of $p$ and $\q$. Notably, when $\alpha > 0$, the ratio $r^{\alpha}$ is always bounded above by $1/\alpha$. %

To address the problem we introduced in \Sec{sec:problem_statement} we will need to jointly estimate all $\{r_v^{\alpha}\}_{v \in V}$, one (relative) likelihood-ratio $r_v^{\alpha}(x)=\frac{\q_v(x)}{(1-\alpha)p_v(x)+\alpha \q_v(x)} \in \real$ for each node. If we represent a graph-level observation by a vector $X=(x_1,...,x_{\Gdims})^{\top} \in \mathcal{X}^{\Gdims}$ whose entries correspond to node-level observations, then -with little abuse of notation- we can also denote vectors computed component-wise, \eg $\mathbf{r}^{\alpha}(X) \mydef (r^{\alpha}_1(x_1),...,r^{\alpha}_{\Gdims}(x_\Gdims))^{\top} \in \real^\Gdims$\!.

\inlinetitle{Connection between \fdiv and \LRE}{.}
In \cite{Nguyen2007}, it is shown that in some cases \fdiv can be rewritten as the solution of a convex optimization problem defined over a space of functions (\emph{variational formulation}). 
\begin{lemma}{\label{lemma:Nguyen2008}} (Lemma\,1 in \cite{Nguyen2007}). 
For any class of functions 
$\mathcal{F} : \mathcal{X} \rightarrow \R$, the lower-bound for the similarity between two probability measures, $P$ and $Q$, admiting \pdfs $p$ and $q$ with respect to the Lebuesgue measure
is:
\begin{subequations}{\label{eq:variational_formulation}}
\begin{align}
    \mathcal{D}_{\phi}(P \Vert Q) &= \int\!\! \phi\left(\frac{q}{p}\right)\!(x)p(x)dx \label{eq:variational_formulation_1}\\ &\geq\, \sup_{ g \in \mathcal{F}} \int\!\! g(x') q(x')dx' - \!\int\!\! \phi^{\conj}(g)(x) \,p(x)dx\!\!\!\label{eq:variational_formulation_2}%
\end{align}%
\end{subequations}%
where $\phi^{\conj}$ denotes the convex conjugate of $\phi: \R \rightarrow \R$. %
The equality \Eq{eq:variational_formulation_1} 
holds 
iff the subdifferential $\nabla \phi (\frac{q}{p})$ contains an element of $\mathcal{F}$.
\end{lemma}
The characterization of the solutions of the optimization problem, and then the estimation of the \fdiv, depend on the functional space $\mathcal{F}$ and
the set defined by the subdifferential of $\phi$, which is evaluated on the likelihood-ratio $r(x)$ for $x \in \X$. This connection has been exploited to define convex functional optimization problems that aim to estimate first the likelihood-ratio, and then the associated \fdiv \citep{Sugiyama2007,Nguyen2007,Nguyen2010}. This approach has the advantage of not requiring any parametric hypothesis about the form of $\q$ or $p$, and leads to non-parametric algorithms that only require data observations coming from $p$ and $\q$. 

\subsection{Important notions for non-parametric estimation}\label{sec:background-non-parametric}

We can solve the optimization problem appearing in \Lemma{lemma:Nguyen2008} using different functional spaces depending on prior information we may have about the likelihood-ratio, the number of available data, and the available computation resources. For example, we can use families of parametric functions, Neural Networks, or Reproducing Kernel Hilbert Spaces (\RKHS). In this work, we will focus on \RKHS as they offer numerous advantages when compared with other functional spaces: they provide geometrical operations defined in Hilbert spaces that facilitate the estimation and theoretical analysis, they allow us to learn in rich infinite-dimensional spaces, and the complexity of the function to be approximated can be elegantly encoded by the norm in the \RKHS. 

\inlinetitle{Scalar RKHS}{.}  Let  $\mathcal{X}$ be a set and $\Hilbert$ a class of functions forming a real Hilbert space with inner-product $\dott{\cdot}{\cdot}_{\Hilbert}$. The function $\KernelFunc:\mathcal{X} \times \mathcal{X} \rightarrow \R$ is called a reproducing kernel of $\Hilbert$ if: 
\begin{enumerate}[topsep=0.4em, itemsep=0.em]
\item $\Hilbert$ contains all functions of the form: $\forall x \in \mathcal{X}, \KernelFunc(x,\cdot): t \rightarrow \KernelFunc(x,t)$. 
\item For every $x \in \mathcal{X}$ and $f \in \Hilbert$ the reproducing property holds: $f(x)=\dott{f}{\KernelFunc(x,\cdot)}_{\Hilbert}$
\end{enumerate}
If a reproducing kernel exists, then $\Hilbert$ is called a \RKHS. It can be shown that a \RKHS has a unique reproducing kernel, and conversely, that a function $\KernelFunc$ describes at most one \RKHS. 

Traditional Kernel Methods focus on modeling scalar functions in the \RKHS space associated with a positive definite kernel $\KernelFunc$. In our case, we want to approximate the vector-valued function $\mathbf{r}^{\alpha}(X) \mydef (r^{\alpha}_1(x_1),...,r^{\alpha}_{\Gdims}(x_\Gdims))^{\top} \in \real^\Gdims$\! where each dimension is associated with a node of the graph $G$. Moreover, we would like the functional space to be rich enough to approximate each of the likelihood-ratios and incorporate the structure of the graph. This is possible via Vector-Valued Reproducing Kernel Hilbert Spaces (\VVRKHS), which are a generalization of scalar \RKHS and have been studied previously in the literature \citep{Micchelli2005,Carmeli2006,Alvarez2012}. %
In this multivariate formulation, the reproducing kernel function is a matrix in $\R^{\Gdims \times \Gdims}$ instead of a scalar, and the elements of the associated Hilbert space will be vector-valued functions $\mathbf{f}:\mathcal{X} \rightarrow \R^{\Gdims}$.

\inlinetitle{Vector-valued Kernels and associated RKHS}{.} 
A positive vector-valued kernel in $\R^{\Gdims}$  on $\mathcal{X}\times \mathcal{X}$ is a map $\mathbf{\Gamma}:\mathcal{X}\times \mathcal{X} \rightarrow \R^{\Gdims \times \Gdims}$ such that, for all $n \in \N$, $x_1,...,x_n \in \mathcal{X}$ and $a_1,...,a_n \in \C$:%
\begin{equation}
    \sum_{i,j=1}^{n} a_i \bar{a}_j \dott{\mathbf{\Gamma}(x_i,x_j)c}{c} \geq 0, \ \ \forall c \in \R^{\Gdims}.
\end{equation}%

As in the scalar case, the positive vector-valued kernel will be associated with a unique vector-valued RKHS (\VVRKHS) 
denoted by $\mathbb{G}$ (see Theorem\,1 in \cite{Micchelli2005}). However, the conditions to characterize  the associated functional spaces differs in the multivariate case. More precisely:
\begin{enumerate}[topsep=0.4em, itemsep=0.em]
\item For every $c \in \R^{\Gdims}$  and $x \in \mathcal{X}$,
$\mathbf{\Gamma}(x,\cdot)c \in \mathbb{G}$.
\item For every $\mathbf{f} \in \mathbb{G}$:
$\dott{\mathbf{f}}{\mathbf{\Gamma}(x,\cdot)c}_{\mathbb{G}}=\mathbf{f}(x)^{\top}c$,
\end{enumerate}
where $\dott{\cdot}{\cdot}_{\mathbb{G}}$ 
is the dot product of the Hilbert space $\mathbb{G}$. %
A more complete discussion over these concepts can be found in \cite{Micchelli2005, Carmeli2006}.

\subsection{Important notions of Graph Signal Processing}

The combinatorial graph Laplacian operator associated with $\G=(V,E,W)$ is defined by: 
\begin{equation}
\Lpc=\diag((d_v)_{v \in V}) - W,
\end{equation}
where $\diag(\cdot)$ is a diagonal matrix with the elements of the input vector in its diagonal. %

Given a metric space $\mathcal{Y}$, let us define the notion of \emph{graph function} $\vartheta:V \rightarrow \mathcal{Y}$, which assigns to each node of a graph an element of $\mathcal{Y}$. When $\mathcal{Y}=\R$, $\vartheta$ is also known as \emph{graph signal} \citep{Shuman2013}.  The smoothness of $\vartheta$ \wrt a graph is defined as: 
\begin{equation}{\label{def:smoothness1}}
 S(\vartheta)=\sum_{v \in V} \sum_{u \in \nghood(v)}  W_{uv} (\vartheta(u)-\vartheta(v))^2.
\end{equation}
In this paper, we use a generalization of this notion to elements of $\mathcal{Y}$;
formally, for $\vartheta(u)\in \Hilbert$, 
$S(\vartheta)$ takes the form: 
\begin{equation*}{\label{def:smoothness2}}
 S(\vartheta)= \sum_{v \in V} \sum_{u \in \nghood(v)} W_{uv} \norm{\vartheta(u)-\vartheta(v)}^2_{\Hilbert}.
\end{equation*}
The lower $S(\vartheta)$ is, the smoother we 
say the function $\vartheta$ is \wrt the graph $G$. 
The concept of smoothness formalizes the idea that two connected nodes $u$ and $v$ have similar behavior in terms of $\vartheta(u)$, $\vartheta(v)$.  


\section{Graph-based Relative Unconstrained Least-squares Importance Fitting (GRULSIF)}\label{sec:GRULSIF}

\subsection{\LRE and \fdiv estimation via Laplacian-penalized least-squares}{\label{sec:estimation}}

In this section, we present our graph-based \LRE framework for the joint estimation of the $\Gdims$ likelihood-ratios at the nodes of a graph (see \Fig{fig:GRULSIF}), in a collaborative and distributed manner. We desire to approximate each node's $r_v^{\alpha}$ with a function $f_v \in \Hilbert$. Note that, for any input $X \in \mathcal{X}^{\Gdims}$\!, $\mathbf{r}^{\alpha}(X)$ essentially induces a graph signal, which we assume to be smooth \wrt the graph. This essentially suggests that two adjacent nodes, $u$ and $v$, should exhibit similar likelihood-ratios, $r_u^{\alpha}$ and $r_v^{\alpha}$; and for that to happen the learned models, $f_u$ and $f_v$, should give similar estimations. Notice that, by the definition of the likelihood-ratio, this hypothesis is true when $p_v=\q_v$, $\forall v$, even if there is heterogeneity among the nodes (\ie generally, $p_v \neq p_u$). This latter is the basis of our approach, which takes inspiration from the RULSIF method of \cite{Yamada2011}, and hence we call Graph-based Relative Unconstrained Least-Squares Importance Fitting (GRULSIF). %
In general terms, our estimation strategy capitalizes over the toolbox described in \Sec{sec:background} as follows: 
\begin{enumerate}[topsep=0.4em, itemsep=0.em]
    \item The variational formulation of \Lemma{lemma:Nguyen2008} will allow us to define a functional optimization problem at the node level, aiming to approximate the relative likelihood-ratio $r_v^{\alpha}$, while requiring minimal hypotheses for $\{p_v\}_{v \in V}$ and $\{\q_v\}_{v \in V}$.
    \item The concept of graph smoothness and \VVRKHS will encode the geometry of the problem, and will formalize a \emph{collaborative estimation procedure}. 
    \item The properties of \VVRKHS, more precisely the Representer theorem, will provide the required elements to translate the optimization problem from a potentially infinite-dimensional space into a simple optimization problem in $\R^L$, where $L$ is the total number of available observation from all the nodes. Moreover, this approach will lead to efficient likelihood-ratio estimators, $\hat{f}_v$, that can be evaluated at any point $x \in \X$ by just computing a dot product in $\R^L$.
\end{enumerate}

\subsection*{A) Node-level relative likelihood-ratio estimation}

We introduce the scalar RKHS $\Hilbert$ with the scalar reproducing kernel $K$ 
and the feature map $\varphi(\cdot)$ (see \Sec{sec:background-non-parametric}). 
Our goal is to estimate the vector-valued function $\mathbf{r}^{\alpha} = (r^{\alpha}_1,...,r^{\alpha}_\Gdims)^\top$ via the $\mathbf{f} = (f_1,...,f_\Gdims)^\top \in \Hilbert^{\Gdims}$, under the hypothesis that $f_u$ and $f_v$ are expected to be similar if nodes $u$ and $v$ are connected in $\G$. When \Lemma{lemma:Nguyen2008} is applied to the \PEdiv between $p^{\alpha}_v$ and $\q_v$ we conclude that it can be alternately be estimated as: 
\begin{equation}{\label{eq:PE_div}}
\begin{aligned}
    \PE(p^{\alpha}_v \Vert \q_v)= \frac{1}{2}\ExpecAlphav{(r_v^{\alpha}(y)-1)^2} 
     = \ExpecA{r_v^{\alpha}(x')} - \frac{1}{2} \ExpecAlphav{r_v^{\alpha}(y)^2}  - \frac{1}{2}.
\end{aligned}
\end{equation}
The details are left for the Appendix\,\ref{appendix:methodological}. Furthermore, the variational formulation $\PE(p^{\alpha}_v,\q_v)$ motivates the following functional optimization problem:
\begin{equation}{\label{eq:least_squares_node_formulation}}
  \hat{f}_v = \argmin_{f \in \Hilbert} \frac{1}{2}\ExpecAlphav{f^2(y)} - \ExpecAm{f(x')}{v} = \argmin_{f \in \Hilbert} \frac{1}{2}\ExpecAlphav{(r^\alpha_v(y)-f(y))^2}.
\end{equation}%
To arrive to the second equality, it is sufficient to use the relationship: $\ExpecAlphav{r^\alpha_v(y)f(y)}= \ExpecAm{f(x')}{v}$,
which is easy to verify using the definition of $r^\alpha_v(x)$ (\Eq{eq:rel-density-ratio}) and by ignoring the constants that do not depend on $f$. Notice that although the above least-squares problem is not directly solvable in practice, as the function $r^\alpha_v$ is unknown, yet it provides an appealing viewpoint to the problem we aim to solve: the approximation of $r^\alpha_v$ using a quadratic risk with respect to the node-level \pdfs $p_v^{\alpha}$. 

\subsection*{B) Multitasking formulation of the \LRE over graphs}

The \LRE problem based on \PEdiv, motivates a multi-task formulation of the relative likelihood-ratios estimation over a graph, through the following objective function: 
\begin{equation}
{\label{eq:multitasking}}
\begin{aligned}
\argmin_{\{f_v\}_{v \in V} \in \Hilbert^{\Gdims}} & \frac{1}{N} \sum_{v \in V} \left( \frac{1}{2}\ExpecAlphav{(r^\alpha_v(x)-f_v(x))^2}  \right)
    + \frac{\lambda}{4} \sum_{u,v \in V} \! W_{uv} \norm{f_u-f_v}^2_{\Hilbert} 
	+ \frac{\lambda \gamma}{2} \sum_{v \in V} \norm{f_v}^2_{\Hilbert} \\
=  \argmin_{\{f_v\}_{v \in V} \in \Hilbert^{\Gdims}} & \frac{1}{N} \sum_{v \in V} \left( \frac{1}{2} \ExpecAlphav{f_v^2(x)} - \ExpecAm{f_v(x)}{v} \right) + \frac{\lambda}{4} \sum_{u,v \in V} \! W_{uv} \norm{f_u-f_v}^2_{\Hilbert} 
+ \frac{\lambda \gamma}{2} \sum_{v \in V} \norm{f_v}^2_{\Hilbert}.
\end{aligned}
\end{equation}%
The first term of the objective function is a loss asking for a good approximation at each node; the second term  introduces our 
hypothesis that adjacent nodes are expected to have similar likelihood ratios; the third term is a penalization term aiming to reduce the risk of overfitting \citep{Sheldon2008}, where $\lambda$,\,$\gamma>0$ are penalization coefficients. 

Let us define the vector-valued kernel: 
\begin{equation}{\label{eq:def_gamma}}
    \mathbf{\Gamma}(x,x')=\KernelFunc(x,x')(\Lpc+ \gamma\id_{\Gdims})^{-1} \in \R^{ \Gdims \times \Gdims}.
\end{equation}
Given the properties of the graph Laplacian $\Lpc$, it can be shown that $\mathbf{\Gamma}(\cdot,\cdot)$  is a positive vector-valued kernel inducing a \VVRKHS $\mathbb{G}$, in which the norm of any $\mathbf{f} \in \mathbb{G} $ is defined as: 
\begin{equation}{\label{eq:penalized_term_G}}
\norm{\mathbf{f}}^2_{\mathbb{G}}=   \frac{1}{2} \sum_{u,v \in V} \! W_{uv} \norm{f_u-f_v}^2_{\Hilbert}+ \gamma \sum_{v \in V} \norm{f_v}^2_{\Hilbert}.
\end{equation}
Notice that the norm in $\mathbb{G}$ incorporates \emph{both} the geometry induced by the structure of the graph Laplacian $\Lpc$ and the geometry of the scalar RKHS $\Hilbert$. 

As explained in \Sec{sec:problem_statement}, we assume that there is access to samples $\mathbf{X}$, $\mathbf{X}'$. Then, the optimization problem \ref{eq:multitasking} can be written as a penalized empirical risk minimization problem  in terms of the elements of  $\mathbb{G}$, the vector-valued functions $\mathbf{f}=(f_1,...,f_{\Gdims})^{\top}$:%
\begin{equation}{\label{eq:problemG}}
    \min_{\mathbf{f} \in \mathbb{G}} \frac{1}{N} \sum_{v \in V} \left( \frac{1-\alpha}{2 n_v} \sum_{i=1}^{n_v} f^2_v(x_{v,i}) +  \frac{\alpha}{2 n'_v} \sum_{i=1}^{n'_v} f^2_v(x'_{v,i}) -\frac{1}{n'_v}\sum_{i=1}^{n'_v} f_v(x'_{v,i})  \right) + \frac{\lambda}{2} \norm{\mathbf{f}}^2_\mathbb{G}.
\end{equation}
The \VVRKHS formulation allows us to apply the Representer theorem (Theorem\,5 in \cite{Micchelli2005}), meaning that the solution to \Problem{eq:problemG} takes the form: 
\begin{equation}
    \hat{\mathbf{f}}(\cdot)= \sum_{i=1}^{L} \mathbf{\Gamma}(x_i,\cdot) c_i = \sum_{i=1}^{L} \KernelFunc(x_i,\cdot) (\Lpc+\gamma \id_{\Gdims})^{-1} c_i ,
\end{equation}
where $L=\sum_{v \in V} (n_v+n'_v)$ is the total number of observations in all nodes, and  $c_{i} \in \R^{\Gdims}$, $i=1,...,L$. The second equality comes from \Eq{eq:def_gamma}.

More specifically, the node-level approximation takes now the form: 
\begin{equation}{\label{eq:nodelevel_likelihoodratio}}
\hat{f}_v(\cdot)=\sum_{i=1}^{L} \KernelFunc(\cdot,x_i) [(\Lpc+\id_{\Gdims})^{-1}]_{v,:} c_i= \sum_{i=1}^{L} \KernelFunc(\cdot,x_i) \theta_{v,i}  = 
\varphi(\cdot)^{\top} \theta_{v},
\end{equation}
where, for the second equality we have defined $\theta_{v,i}=[(\Lpc+\id_{\Gdims})^{-1}]_{v,:} c_i$, we define the feature map \wrt all observations as the function $\varphi:\mathcal{X} \rightarrow \R^{L}$,  $\varphi(x) 
= (\KernelFunc(x,x_1),...,\KernelFunc(x,x_L))^{\top} \in \R^{L}$. 
The last equality uses $\theta_v=(\theta_{v,1},...,\theta_{v,L})^{\top} \in \R^{L}$, which is an abuse of notation that is helpful for the presentation. 
By the definition of $\hat{f}_v(\cdot)$, its norm in $\Hilbert$ can be elegantly written as: %
\begin{equation}{\label{eq:norm_fv}}
\norm{\hat{f}_v}^2_{\Hilbert}= \theta_v^{\top} \Gramm \theta_v,
\end{equation}
where $\Gramm \in \R^{L \times L}$ is the Gram matrix
$\Gramm_{ij}$ associated with the scalar kernel function $\KernelFunc(\cdot,\cdot)$. %
We can conclude from \Expr{eq:nodelevel_likelihoodratio} that approximating 
$(r_1^{\alpha},..,r^{\alpha}_{\Gdims})^{\top}$ amounts to estimating the node parameters $\theta_{v}$, $v \in V$.

\subsection*{C) \LRE as a quadratic problem in $\R^{NL}$}

Let all node parameter vectors get concatenated in a single vector $\ThetaG=\Vect(\theta_1^{\top},...,\theta^{\top}_{\Gdims})^{\top} \in \real^{\Gdims L}$. Moreover, let us introduce the following terms associated with a specific feature map (here this is $\phi(\cdot)$), which need to be computed only once at the beginning of the process: %
\begin{equation}{\label{eq:updatehs}}
\begin{aligned}
\!\!\!\!H_{v} &= \frac{1}{\Npre_v}\sum_{x \in \mathbf{X}_v} \! \varphi(x) \varphi(x)^\top \!\in \R^{L \times L}\!, \ \ 
H_{v}' = \frac{1}{\Npost_v}\sum_{x \in \mathbf{X}'_v} \! \varphi(x)\varphi(x) ^\top \!\in \R^{L \times L}\!, 
\ \ h_{v}' = \frac{1}{\Npost_v}\sum_{x \in \mathbf{X}'_v} \! \varphi(x) \in \R^{L}, \\
\mathbf{H} &= \Block(H_{1},...,H_{\Gdims}) \in \R^{L \Gdims \times L \Gdims}\!, \ \ \mathbf{H'}= \Block(H'_{1},...,H'_{\Gdims}) \in \R^{L \Gdims \times L \Gdims}\!, \ \
\\
\!\!\!\! \mathbf{h}' &= \Vect(h'_1,h'_2,...,h'_{\Gdims})^{\top
} \in \R^{L \Gdims}.
\end{aligned}
\end{equation}
%
%
By putting everything together and by using \Eq{eq:norm_fv}, 
\ref{eq:updatehs}, we conclude that the functional optimization \Problem{eq:multitasking} can be restated as a quadratic problem \wrt the vector $\ThetaG$: 
\begin{equation}\label{eq:thetas_cost_function1}
\begin{aligned}
\min_{\ThetaG \in \real^{\Gdims L }} \Phi(\ThetaG) & = \min_{\ThetaG  \in \real^{\Gdims L }}    \frac{1}{\Gdims} \sum_{v \in V }  \left( \frac{1-\alpha}{2} \theta_v ^\top  H_v \theta_v + \frac{\alpha}{2} \theta_v ^\top  H'_v \theta_v  -  {h'}_v^{\top} \theta_v \right) \\
&\hspace{3.2em} + \frac{\lambda}{4} \sum_{u,v \in V}  W_{uv} (\theta_v-\theta_u)^{\top} \Gramm (\theta_v-\theta_u) +
\frac{\lambda \gamma}{2} \sum_{v \in V } \theta_v^{\top} \Gramm \theta_v
\\ 
& =\min_{\ThetaG \in \real^{\Gdims L }} \frac{1}{\Gdims}  \left( \frac{1-\alpha}{2} \ThetaG ^\top  \mathbf{H} \ThetaG + \frac{\alpha}{2} \ThetaG^\top \mathbf{H'} \ThetaG  -  \mathbf{h'}^{\top} \ThetaG \right) \\
&\hspace{3.5em} + \frac{\lambda}{2}  \ThetaG^{\top} ( \id_{N} \otimes \Gramm^{\frac{1}{2}} )^{\top}
 \left[\Lpc  \otimes \id_{L} \right](\id_{N} \otimes \Gramm^{\frac{1}{2}}) \ThetaG 
\\
&\hspace{3.5em} + \frac{\lambda\gamma}{2}  \ThetaG^{\top} ( \id_{N} \otimes \Gramm) \ThetaG 
\\
\Longleftrightarrow\ \  \min_{\ThetaG \in \real^{\Gdims L}} \Phi(\ThetaG) & = \min_{\ThetaG \in \real^{\Gdims L }}  \ThetaG^{\top} \mathbf{A} \ThetaG - \frac{1}{N} \mathbf{h'}^{\top} \ThetaG,
\end{aligned}
\end{equation}%
where:%
\begin{equation}
\mathbf{A} = \frac{1}{N} \left(\frac{1-\alpha}{2} \ThetaG ^\top  \mathbf{H} \ThetaG + \frac{\alpha}{2} \ThetaG^\top \mathbf{H'} \ThetaG \right)
+  \frac{\lambda}{2}  \ThetaG^{\top} ( \id_{N} \otimes \Gramm^{\frac{1}{2}} )^{\top} 
 \left[\Lpc  \otimes \id_{L} \right](\id_{N} \otimes \Gramm^{\frac{1}{2}}) \ThetaG  + \frac{\lambda\gamma}{2}  \ThetaG^{\top} ( \id_{N} \otimes \Gramm) \ThetaG. 
\end{equation}
Notice that $\mathbf{A}$ is a semi-positive definite matrix given that $\Lpc$ and $\Gramm$ are semi-positive definite as well, which implies that \Problem{eq:thetas_cost_function1} is a quadratic optimization problem in $\ThetaG$. We will exploit this fact in \Sec{sec:practical_implementation} to propose an efficient optimization procedure that scales nicely \wrt $\Gdims$ and $L$. 

\subsection*{D) Pearson's \PEdiv estimation}{\label{sec:Pearson_approximation}}

We can use the estimated likelihood-ratio $\hat{f}_{v}$,
\Eq{eq:nodelevel_likelihoodratio}, and \Eq{eq:updatehs} to approximate the following expectation that corresponds to the loss $\ell_v(\theta_v)$ at node $v$: 
\begin{equation}{\label{eq:definition_Lv}}
\begin{aligned}
\!\!\!\!\!\!\frac{1}{2}\ExpecAlphav{f_v^2(x)} - \ExpecAm{f_v(x)}{v} &\approx    \frac{1-\alpha}{2} \left( \sum_{x \in \setprev} \!\!\frac{\hat{f}_v(x)^2}{\Npre_v} \right) + \frac{\alpha}{2} \left(  \sum_{x' \in \setpostv} \!\!\frac{\hat{f}_v(x')^2}{\Npost_v}\right)  - \sum_{x' \in \setpostv} \!\!\frac{\hat{f}_v(x')}{\Npost_v}\!\!  \\
& = \frac{1-\alpha}{2} \hat{\theta}_v ^\top  H_v \hat{\theta}_v + \frac{\alpha}{2}  \hat{\theta}_v ^\top  H'_v \hat{\theta}_v  -  h'_v \hat{\theta}_v \ =:\  \hat{L}_v(\hat{\theta}_v),
\end{aligned}
\end{equation}%
We can use this expression to arrive at the more compact and convenient formulation of \Problem{eq:thetas_cost_function1}:
\begin{equation}\label{eq:thetas_cost_function_2}
\begin{aligned}
\!\!\!\!\!\min_{\ThetaG \in {\R}^{ \Gdims L}} \frac{1}{\Gdims} \bigg(\sum_{v \in V} \hat{L}_v(\theta_v)\bigg) +  \frac{\lambda}{2}  \ThetaG^{\top} ( \id_{N} \otimes \Gramm^{\frac{1}{2}} )^{\top} 
 \left[\Lpc  \otimes \id_{L} \right](\id_{N} \otimes \Gramm^{\frac{1}{2}}) \ThetaG  + \frac{\lambda\gamma}{2}  \ThetaG^{\top} ( \id_{N} \otimes \Gramm) \ThetaG.
\end{aligned}
\end{equation}
Moreover, we use \Eq{eq:PE_div} to propose an approximation of $\PE(p^{\alpha}_v \Vert \q_v)$ based on the estimated parameters $\hat{\ThetaG}$ and the available samples $\setprev$ and $\setpostv$:%
\begin{equation}\label{eq:PEapprox}
     \hat{\PE}{}^{\alpha}_v(\setprev \Vert \setpostv)  =  -\hat{L}_v(\hat{\theta}_v) - \frac{1}{2}.  
\end{equation}
\Eq{eq:thetas_cost_function_2} and \Eq{eq:PEapprox} highlight how
minimizing \Eq{eq:thetas_cost_function_2} amounts to maximizing the estimated \PEdiv while at the same time accounting for the structure of the graph and the geometry of the RKHS $\Hilbert$. 

Finally, let us define the following expression for $\mathbf{f} \in \Ghilbert$: 
\begin{equation}
\PE{}^{\alpha}_v(f_v):= \ExpecAm{f_v(x)}{v} - \frac{1}{2}\ExpecAlphav{f_v^2(x)} - \frac{1}{2},  \ \ \forall v \in V.
\end{equation}
Notice that, as a consequence of \Lemma{lemma:Nguyen2008}:
\begin{equation}{\label{eq:PE_div_f_v}}
\PE{}^{\alpha}_v(r^{\alpha}_v)=\PE(p^{\alpha}_v \Vert \q_v) \ \ \text{and} \ \ \PE{}^{\alpha}_v(r^{\alpha}_v) \geq \PE{}^{\alpha}_v(f_v).
\end{equation}

\subsection{Comments regarding other \fdivs}

The line of reasoning presented in \Sec{sec:estimation} is general enough to be applicable using other \fdivs to produce 
likelihood-ratio estimates that account for a graph structure. To summarize: First, we infer a node-level loss function in terms of the RKHS $\Hilbert$, thanks to the variational formulation of \Lemma{lemma:Nguyen2008}; then, we restate the graph-based \LRE problem as a sum of the node-level loss functions plus a Laplacian penalty term that enforces adjacent nodes to estimate similar likelihood-ratios. The resulting functional optimization problem satisfies the hypotheses of the Representer Theorem for \VVRKHS, meaning that it leads to an optimization problem in $\R^{LN}$, where $N$ is the number of graph nodes and $L$ is the total number of data observations available in all nodes.

The main reason for using Pearson's \PEdiv is that the \LRE takes the form of an unconstrained penalized least-squares problem. Moreover, the likelihood-ratio estimates are the solution to a linear system. Leveraging these features, we can seamlessly adapt existing and efficient optimization techniques tailored for penalized least-squares, and hence integrate a mature theoretical framework, to gain insight into the properties of the estimators. Such advantages may not be offered or be readily available for other \fdivs.

\section{Convergence guarantees}{\label{sec:theory}}

In this section, we discuss the generalization properties of GRULSIF, more precisely the gains brought by the collaborative \LRE when Pearson's \PEdiv is used as surrogate cost function. The main result of this section is summarized in \Theorem{thm:convergence_results}. 

For the rest of the section, we will assume $n_v=n'_v:=n$, \ie that we have the same number of observations from $p_v$ and $\q_v$ at each node $v$, and all the nodes have the same sample size. Moreover, we will assume observations come in pairs $z_v=(x_v,x'_v)$ as realizations of a probabilistic model described by the joint \pdf $p_{z,v}$ with marginal \pdfs  $p_v$ and $q_v$. 

Let us start by defining the functional space: 
\begin{equation}{\label{eq:functional_space_graph_smoothness}}
\mathcal{F}_G=\{ \mathbf{f}=(f_1,...,f_{\Gdims}) \in \Ghilbert \,:\, \frac{1}{2} \norm{\mathbf{f}}_{\Ghilbert} \leq \Lambda^2 \}, 
\end{equation}
where $\Lambda \geq 0$ is a positive constant controlling the smoothness of the vector-valued function to be learned \wrt the graph $G$ and the Hilbert space $\Hilbert$. The first thing to notice is that the estimation problem 
\ref{eq:problemG} can alternatively be written in terms of the functional space $\mathcal{F}_G$: 
\begin{equation}{\label{eq:problem_collaborative_FG}}
    \min_{\mathbf{f} \in \mathcal{F}_G } \frac{1}{N} \sum_{v \in V} \left(  \frac{(1-\alpha)}{2n} \sum_{i=1}^{n} f^2_v(x_{v,i}) +  \frac{\alpha}{2n} \sum_{i=1}^{n} f^2_v(x'_{v,i}) - \frac{1}{n}\sum_{i=1}^{n} f_v(x'_{v,i}) \right).
\end{equation}
\begin{assumption1}\label{ass: independence}
$\{(x_{v,i},x'_{v,i})\}_{v\in V,i=1,...n}$ represent $n\Gdims$ pairs of independent observations, where for each node $v \in V$, the pairs $\{(x_{v,1},x'_{v,1}),...,(x_{v,n},x'_{v,n})\}$ are identically distributed under the joint law $p_{z,v}$, where  
$x_i \sim p_v$ and  $x'_i \sim q_v$.
\end{assumption1}
The node independence assumption is present in previous works on \LRE \citep{Nguyen2007, Nguyen2010, Sugiyama2012} when a single data source is studied ($\Gdims=1$). This can be considered to be a strong hypothesis for many applications, but it is standard in theoretical analyses based on Multitasking on Vector-Valued Kernels \citep{Maurer2006,Yousefi2018}. 

\begin{assumption1}\label{ass:kernel_map_upperbound} 
The reproducing kernel map $\KernelFunc(\cdot,\cdot)$ can be upper-bounded by a constant $C > 0$:\!\!\!\!\\ 
\begin{equation}
    \sup_{x \in \mathcal{X}} \sqrt{\KernelFunc(x,x)} \leq C < \infty.
\end{equation} 
\end{assumption1}
This assumption is satisfied by commonly used kernels, such as the Gaussian and the Laplacian kernels, and in general, for continuous kernel maps $\KernelFunc(\cdot,\cdot)$ defined in a compact input space $\mathcal{X}$. This hypothesis is commonly used in the Kernel Methods literature. 
\begin{assumption1}{\label{ass:model_definition}}
  There exists $\Lambda>0$ such that $\mathbf{r}^{\alpha} = (r^{\alpha}_1,...,r^{\alpha}_\Gdims) \in \mathcal{F}_{G}$.
\end{assumption1}
\Assumption{ass:model_definition} states that the proposed statistical model is well-defined. In particular, it implies: i) $r_{v} \in \Hilbert$ , for all $v \in V$, a common hypothesis in the \LRE literature \citep{Nguyen2007,Nguyen2010,Sugiyama2012}; ii) it introduces the parameter $\Lambda$, which relates to the regularization constant $\lambda$ (\Problem{eq:thetas_cost_function1}), and formalizes the a priori information encoded in the graph and is required to estimate the vector $\mathbf{r}^{\alpha}$. 

Let $\varphi(y)$ the feature map associated with the \RKHS $\Hilbert$, and let us consider $g,h \in \Hilbert$ and define the operator $g\otimes h:\Hilbert \rightarrow \Hilbert$ as $g\otimes h(f)=\dotH{f}{h}g$. Then, we can define the covariance operator associated to the node $v \in V$ as:
\begin{equation} \CO_v=\ExpecAlphav{\varphi(y)\otimes\varphi(y)},
\end{equation}
Assuming the feature space $\mathcal{X}$ is compact and the $\KernelFunc$ is continuous, the Mercer's theorem implies \citep{Aronszajn1950,Dieuleveut2017}: 
\begin{equation}{\label{eq:eigendecomposition}}
    \CO_v= \sum_{i=1}^{\infty} \mu_{v,i} \tilde{\varphi}_{v,i} \otimes \tilde{\varphi}_{v,i},
\end{equation}
 where $\{\tilde{\varphi}_{v,i}\}_{i\in\N}$  forms a Hilbertian basis of $\Hilbert$, with assoaciated eigenvalues $\{\mu_{v,i}\}_{i \in \N}$. Nevertheless, there exists more general settings where \Expr{eq:eigendecomposition} is satisfied (see \cite{Dieuleveut2017}). 
\begin{assumption1}{\label{ass:capacitycondition}}
For each $v \in V$, assume $\CO_v$ satisfies \ref{eq:eigendecomposition}. Denote
by $I$ the set of indexes of non-zero eigenvalues
$\left\{\mu_{v,i}\right\}_{i \in I}$ of the operator $\CO_v$ arranged in decreasing order. We assume that $\mu_{v,i} \leq s^2_v i^{-\zeta_v}$ $i \in I$ for some $\zeta_v>1$ and some $s_v>0$.   
\end{assumption1} 
This assumption is known in the literature as the \emph{capacity condition}. It quantifies the size of the \RKHS $\Hilbert$ \wrt the eigenbasis $\left\{\mu_{v,i}\right\}_{i \in I}$. Larger $\zeta_v$ values lead to faster eigenvalue decay, which means the number of basis functions required to approximate $\Hilbert$ reduces. In particular, this means that $r_{v}^{\alpha}$ can be approximated by a smaller space. When $\zeta_v$ approaches $1$, a bigger space will be needed to approximate the elements of $\Hilbert$, including $r_{v}^{\alpha}$. This assumption has been discussed and analyzed in previous works to obtain optimal convergence rates in the context of Kernel Ridge regression \citep{Caponnetto2006,Ying2007,Steinwart2009}. The assumption of exponential eigenvalues decay is satisfied for finite rank kernels and other well-known kernels, such as the Gaussian kernel.

\begin{theorem}{\label{thm:convergence_results}}
If Assumptions\,\ref{ass: independence}-\ref{ass:model_definition} are satisfied, and $\mathcal{F}_G$ is a class of functions with ranges in $[-b,b]$. Then, for any $C \geq 1$ and $\delta \in (1,0)$, with probability at least $1-\delta$, the solution to \Problem{eq:problem_collaborative_FG}, $\hat{\mathbf{f}}=(\hat{f}_1,...,\hat{f}_{\Gdims})$, satisfies: 
\begin{equation}
\begin{aligned}
\frac{1}{\Gdims} \sum_{v \in V} \left[ \PE(p^{\alpha}_v \Vert \q_v)-\PE{}^{\alpha}_v(\hat{f}_v) \right]
& \leq  2C(20^2) B_1  \rho^* + \frac{16 B_0^2 C}{n \Gdims} \log{\frac{1}{\delta}} + \frac{24B_0B_1}{n \Gdims} \log{\frac{1}{\delta}}\\
\frac{1}{\Gdims} \sum_{v \in V} \Expecalphav\left[ [\hat{f}_v-r^{\alpha}_{v}]^2  \right] & \leq   4C(20^2) B_1  \rho^* + \frac{32 B_0^2 C}{n \Gdims} \log{\frac{1}{\delta}} + \frac{48B_0B_1}{n \Gdims} \log{\frac{1}{\delta}},
\end{aligned}
\end{equation}
where: 
\begin{equation*}\textstyle
B_0= \frac{1}{2}\left[\left(b+\frac{1}{\alpha}\right)^2 +\frac{4}{\alpha} \right], \ \ B_1= \frac{1}{2}\left(b+\frac{1}{\alpha}\right)^{2}+\left(b+\frac{1}{\alpha}\right),
\end{equation*}
\begin{equation}\textstyle
\rho^* \leq 8B_0\sqrt{ \frac{\zeta^{*}+1}{\zeta^{*}-1}} \left[\left(b+\frac{1}{\alpha}\right)^{2 \zeta^{*}} \!\!\Lambda^2 \Lpc^{-1}_{\max}\right]^{\frac{1}{1+\zeta^{*}}}   n^{\frac{-\zeta^*}{1+\zeta^{*}}} N^{\frac{-1}{1+\zeta^{*}}} s_{\max}^{\frac{1}{1+\zeta^{*}}},
\end{equation}
\begin{equation*}
\zeta^*=\min_{v \in V} \zeta_{v} \ (\text{recall} \ \ \zeta_v>1, \,\forall v\in V), \ \ s_{\max}=\max_{v \in V} s_v,  \ \ \Lpc^{-1}_{max}= \max_{v \in V} \abs{(\Lpc+\gamma)^{-1}_{vv}}.
\end{equation*}
\end{theorem}
The proof of \Theorem{thm:convergence_results} is provided in Appendix\,\ref{app:convergence_guarantees}, and it relies mainly on the framework of Local Rademacher Complexities for Multitask Learning introduced in \cite{Yousefi2018}. Notice that convergence rates are given in terms of the excess risk and the $L_2$ distance between $\hat{f}_v$ and $r_v^{\alpha}$
with respect to the measure $p^{\alpha}_v$. The excess risk 
takes the form of the difference between the expected divergence $\PE{}^{\alpha}_v(\hat{f}_v)$ and the real \PEdiv $\PE(p^{\alpha}_v \Vert \q_v)$ it aims to approximate.

The convergence rates depend mainy on the number of observations per nodes $n^{\frac{-\zeta^*}{1+\zeta^{*}}}$, the number of nodes in the graph $N^{\frac{-1}{1+\zeta^{*}}}$, the smoothness of the function to be approximated ${\Lambda}^{\frac{1}{1+\zeta^{*}}}$ and the effective dimension of the space to approximate each $r_{v}^{\alpha}$, this feature is encoded in the variables $\zeta^*$ and $s_{\max}$. When $\zeta^*$ is small, \ie close to $1$, the convergence rate can be as slow as $\bigO\left(\frac{\sqrt{s_{\max}} \Lambda}{\sqrt{n \Gdims}}\right)$, and as fast as $\bigO\left(\frac{1}{n}\right)$ when $\zeta_v \rightarrow \infty$ for all $v \in V$.  This means that the gains of the collaborative estimation in excess risk will be more relevant as $\zeta^*$ is smaller, since both the number of nodes and smoothness play a role in the convergence rate. This situation occurs when a larger number of basis functions $\{\tilde{\varphi}_{v,i}\}_{i\in\N}$ are required to approximate the space $\Hilbert$, which could mean $r_v^{\alpha}$ is harder to estimate \wrt the kernel function $\KernelFunc$ and the data distribution. In this situation, a larger number of nodes and a smoother $\mathbf{r}^{\alpha}$ over the graph would improve performance. However, the collaborative estimation will offer little advantage when the \RKHS is low-dimensional (large values of $\zeta^*$), since convergence is governed by the number of observations per node. This suggests that GRULSIF can be used in the regime in which Multitasking is also recommended: when there are many interrelated tasks with little data per task, and each of the tasks is complex to be solved using only the available local data \citep{Yousefi2018,Zhang2021}.

\section{Practical implementation}
\label{sec:practical_implementation}

A straightforward optimization of \Problem{eq:thetas_cost_function1} would set the derivative of the objective function to zero, \ie $\nabla_\ThetaG\Phi(\ThetaG) = 0$, and solve to get the estimated parameters $\hat{\ThetaG}$: 
\begin{equation}{\label{eq:close_solution}}
   \!\hat{\ThetaG} =  \frac{1}{\Gdims}\mathbf{A}^{\dag}\mathbf{h}',
\end{equation}
where $A^{\dag}$ denotes the pseudoinverse of $A^{NL \times NL}$. Nevertheless,  the size of matrix $A$ scales  with the number of nodes in the graph ($\Gdims$) and the number of available observations ($L$). The total complexity of this optimization approach would be of scale $\bigO((LN)^3)$, which makes it prohibitive to compute in most practical situations. %
For deploying GRULSIF in practice, we propose in this section an optimization procedure that can handle efficiently large graphs and a substantial number of \VOID{available} observations. Additionally, we detail the strategy to identify the regularization constants $\lambda$, $\gamma>0$, and the hyperparameters related to the kernel that we will denote by $\sigma$, when $\KernelFunc(\cdot,\cdot)$ is the Gaussian kernel $\sigma$ is the width parameter.

\subsection{Computing the node parameter updates via CBCGD}

Instead of computing $A^{\dag}$ to solve \Problem{eq:thetas_cost_function1}, we propose to use the Cyclic Block Coordinate Gradient Descent (CBCGD) method \citep{Beck2013,Li2018}. The CBCGD-based optimization schema operates in cycles, and each cycle involves multiple iterations of a block coordinate gradient descent (GD) one for each node $v$; therefore, the high-level complexity is $\bigO(\#\text{Cycles} \cdot \#\text{Nodes} \cdot \text{Cost\_of\_GD\_at\_one\_node})$. Starting from  the last term, CBCGD's $i$-th cycle has to estimate the node parameter $\hthetanode{v}^{(i)}$ at each node $v$:
\begin{equation}\label{eq:update}
\begin{aligned} 
 &  \hthetanode{v}^{(i)} \mydef  
 (\lambda \gamma \Gramm + \eta_v \id_{L})^{-1}
\Bigg[ 
\overbrace{\eta_v \hthetanode{v}^{(i-1)} - \left[ \left(
 \frac{1-\alpha}{\Gdims} H_{v} + \frac{\alpha}{\Gdims} H'_{v} \right) \hthetanode{v}^{(i-1)} - \frac{h'_v}{\Gdims} \right]}^{\mathclap{\text{component depending on node } v}} \\
 & \hspace{8mm} \hspace{8mm} \hspace{8mm} \hspace{8mm} \hspace{9mm}- \overbrace{\lambda \Gramm \bigg(\! d_v \hthetanode{v}^{(i-1)}- \!\!\!\sum_{u \in \nghood(v) \!\!\!\!\!\!\!\!} W_{uv}  \big( \Ind{u<v}\hthetanode{u}^{(i)} + \Ind{u \geq v} \hthetanode{u}^{(i-1)} \!\big) \!\bigg)\!\!}^{\mathclap{\text{component depending on the graph}}} \Bigg]\!,
\end{aligned}
\end{equation}
where $\eta_v$ is the node learning rate, and recall that $d_v$ is the node degree. Notice the elegance of the decomposition of the update into two components: one depending on the node $v$ itself, and the other depending on the graph, \ie only on $v$'s neighbors. Important to note that, the node parameters are estimated asynchronously in each cycle in an arbitrary but fixed cyclic order; this is clear in the summation inside the graph-related component. Tweaking this order to adapt it to specific communication restrictions between nodes is possible, but it is left to the reader to specify the most convenient setting for her needs (see \cite{Wright2015} for a review of the topic). CBCGD is easy to implement, and when applied to quadratic problems leads to a manageable complexity in terms of the number of cycles required to achieve convergence \citep{Li2018}. This kind of result is made explicit for the optimization schema described in \Expr{eq:update} in the following theorem.
\begin{theorem}\label{Th:convergence} 
 Suppose that for a dictionary $D$ of size $L \geq 2$ we desire to solve the optimization \Problem{eq:thetas_cost_function1} via the CBCGD strategy, where the update \wrt the node parameter $\theta_v$ at the $i$-th cycle is computed as detailed in \Eq{eq:update}. Then, if we fix the learning rate for node $v$ at $\eta_v= \eigmax{\!\frac{(1-\alpha)}{\Gdims} H_{v} + \frac{\alpha}{\Gdims} H'_{v} + \lambda d_v \Gramm \!}$, we will need at most the following number of cycles for achieving a pre-specified accuracy level $\epsilon>0$:
%
\begin{equation}{\label{eq:complexity}}
\begin{aligned}
i_{\textup{max}}  \mydef&  
\Bigg\lceil \frac{\lambda \gamma c (C_{\textup{min}}+\lambda \gamma c) + 16 C^2 \log^2(3 \Gdims L)}{\lambda \gamma c (C_{\textup{min}}+\lambda \gamma c)} 
\cdot\,\log\bigg( \frac{1}{\epsilon} \left(\Phi(\ThetaG^{(0)})-\Phi(\ThetaG^*)\right)\!\!\bigg) \Bigg\rceil, 
\end{aligned}
\end{equation} 
%
where $\Phi(\ThetaG)$ is the cost function of \Expr{eq:thetas_cost_function1}, $c>0$ is a positive constant, $C_{\textup{min}} \mydef \min_{v \in V} C_v$: 
\begin{equation}
\begin{aligned}
C &\mydef \eigmax{ \frac{1-\alpha}{\Gdims} \mathbf{H}  + \frac{\alpha}{\Gdims} \mathbf{H'} +  \lambda ( \id_{N} \otimes \Gramm^{\frac{1}{2}} )
 \left[\Lpc  \otimes \id_{L} \right](\id_{N} \otimes \Gramm^{\frac{1}{2}})}, \\
C_{v} &\mydef \eigmax{\frac{1-\alpha}{\Gdims} H_{v} + \frac{\alpha}{\Gdims} H'_{v}  + \lambda d_v \Gramm }\! .
\end{aligned}
\end{equation}
\end{theorem}
The proof of \Theorem{Th:convergence} is provided in Appendix\,\ref{appendix:optimization}. The computational complexity of the full optimization schema depends on two components: 1) estimating  the optimal learning rates $\eta_v$ and  the  inversion of the matrix $\lambda \gamma \Gramm + \eta_v \id_{L}$, operations to be done just once for each of the nodes, this step amounts to a computational cost of $\bigO(NL^3)$. 2) The cost of  \Eq{eq:update} across all nodes and cycles. The cost for a node $v$ at a given cycle $i$ is dominated by matrix-vector multiplications of dimension $L$, leading to a cycle cost of $\bigO(N L^2)$. As indicated by Eq. 25, the required number of CBCGD cycles for achieving a given accuracy level $\epsilon$ scales in $\bigO(\log^2(N L))$. The total cost of the second step is then  $\bigO(N L^2\bigO\log^2(N L))$. The total cost of the whole optimization schema is then $\bigO(NL^3+ N L^2\bigO(\log^2(N L))$.

\subsection{Nyström dimensionality reduction strategy}{\label{sec:Nyström_dictionary}}

The main computation burden of the CBCGD method is related to the dataset size, \ie the number of observations $L$. This is a common problem in Kernel Methods and has motivated extensive and diverse research on how to reduce their complexity. For instance, the \emph{random features} approach \citep{Rahimi2007} uses a randomized feature map to approximate the input space by a low dimensional Euclidean space. Low-rank approximations, such as Nyström approximations \citep{Williams2000,Smola2000}, use a subsample of observations as a dictionary, to define a finite dimensional space that preserves the approximation properties of the original space. In the context of time-series analysis, \cite{Cedric2009} proposed to grow the dictionary by adding new elements one-by-one, according to a \emph{coherence threshold} whose aim is to keep the linear dependency of the dictionary elements as low as possible (\ie the basis functions to be as diverse as possible) while still being able to approximate any of the functions in $\Hilbert$.  The usual approach followed in non-parametric \LRE is simply to create a dictionary out of subsample of the observations chosen uniformly at random \citep{Sugiyama2012}.

In general, the choice of the dictionary learning method depends on the task, the time complexity requirements, and the nature of the chosen kernel. Nyström approximations replace the feature map $\varphi(x)$ by its orthogonal projection into a finite-dimensional space $\mathbb{F}={\spankern(\{\varphi(x) : x \in D_{\hat{L}}\})}$, where $D_{\hat{L}} = \{x_i \in \X\}_{i=1}^{\hat{L}}$ is a set of carefully chosen points in the original input space (not restricted to data observations), and $\spankern(\cdot)$ refers to the set of lineal combinations of the input elements, for some chosen $\hat{L} \ll L$. The points $\varphi(x_1),..., \varphi(x_{\hat{L}})$ are known as \emph{anchor points} in $\Hilbert$, and, via the associated kernel matrix 
$\Gramm_{\hat{L}} \in \R^{\hat{L}\times\hat{L}}$, $[\Gramm_{\hat{L}}]_{ij}=\KernelFunc(x_i,x_j)$,
they allow the definition of a new feature map: 
\begin{equation}{\label{eq:psi_definition}}
    \psi(\cdot)=\Gramm_{\hat{L}}^{-\frac{1}{2}} \left(\KernelFunc(\cdot,x_1),..,\KernelFunc(\cdot,x_{\hat{L}})\right)^{\top},%
\end{equation}%
The idea is to choose the anchor points such that the geometry of $\Hilbert$ is preserved, in the sense that the dot product in the infinite dimensional space $\Hilbert$ gets translated into a dot product in $\R^{\hat{L}}$: %
$$\KernelFunc(x,y)=\dott{\varphi(x)}{\varphi(y)}_{\Hilbert} \approx \dott{\psi(x)}{\psi(y)}, \ \ \forall x,y \in \X.$$ 
%
According to the empirical risk minimization and the Representer Theorem (\Expr{eq:nodelevel_likelihoodratio}), the node-level approximation in this new space takes the form: 
\begin{equation}
f_v(x) = \sum_{i=1}^{L} \KernelFunc(x,x_i) \theta_{v,i} = \Big\langle \sum_{i=1}^{L} \varphi(x_i) \theta_{v,i} , \,\varphi(x)\Big\rangle_{\Hilbert} \approx \dott{w_v}{\psi(x)},
\end{equation}
where $w_v \in \R^{\hat{L}}$. This approximation  can rephrase \Problem{eq:thetas_cost_function1} in terms of vectors in $\R^{\hat{L}}$ and the new feature map $\psi(\cdot)$: 
\begin{equation}{\label{eq:Nyström_problem}}
\begin{aligned}
\min_{\ThetaG \in \real^{\Gdims \hat{L} }}  & \frac{1}{\Gdims} \sum_{v \in V }  \left( \frac{1-\alpha}{2} \theta_v ^\top  H_{\psi,v} \theta_v + \frac{\alpha}{2} \theta_v ^\top  H'_{\psi,v} \theta_v  -  h'_{\psi,v} \theta_v \right)  + \frac{\lambda}{4} \sum_{u,v \in V} \! W_{uv} \norm{\theta_v-\theta_u}^2
+ \frac{\lambda \gamma}{2} \sum_{v \in V } \norm{\theta_v}^2\!,
\end{aligned}
\end{equation}
where $\norm{\cdot}$ refers to the Euclidean norm, and the terms $H_{\psi,v},\, H_{\psi,v}' \in \R^{\hat{L} \times \hat{L}}$, and $h_{\psi,v}' \in \R^{\hat{L}}$ are those of \Eq{eq:updatehs}, but now computed using their new associated feature map $\psi(\cdot)$. Recall that these terms need to be computed only once at the beginning. %
Moreover, Nyström approximation does not affect the structure of the problem, which remains quadratic and can be solved via CBCGD, with each iteration taking the form: 
\begin{equation}\label{eq:update_Nyström}
\begin{aligned} 
 &  \hthetanode{v}^{(i)} \mydef  \frac{1}{\lambda \gamma  + \eta_v}
\Bigg[ \overbrace{
\eta_v \hthetanode{v}^{(i-1)} - \left[\left( 
 \frac{1-\alpha}{\Gdims} H_{\psi,v} + \frac{\alpha}{\Gdims} H'_{\psi,v}\right) \hthetanode{v}^{(i-1)} - \frac{1}{\Gdims}h'_{\psi,v} \right]}^{\mathclap{\text{component depending on node } v}} \\
 & \hspace{12mm}\hspace{8mm}\hspace{6mm}- \overbrace{\lambda  \bigg(\! d_v \hthetanode{v}^{(i-1)}- \!\!\!\sum_{u \in V
} \! W_{uv}  \big( \Ind{u<v}\hthetanode{u}^{(i)} + \Ind{u \geq v} \hthetanode{u}^{(i-1)} \!\big) \!\bigg) \!\bigg)\!\!}^{\mathclap{\text{component depending on the graph}}} \Bigg].
\end{aligned}
\end{equation}
The final computational cost is the sum of the cost of encoding the data points via \Expr{eq:psi_definition} and the optimization procedure. The encoding requires a matrix inversion $\bigO(\hat{L}^3)$ and $L$ matrix-vector multiplications of dimension $\hat{L}$ (this is overall $\bigO(L \hat{L}^2)$, while CBCGD requires the estimation of the optimal learning rates $\eta_v$ and has total cost $\bigO(N \hat{L}^3)$, and the cost of all node iterates across cycles that amounts to $\bigO(\Gdims \hat{L}^2  \log^2(\Gdims \hat{L}))$.  In conclusion, Nyström approximation enables the reduction of the computation complexity from $\bigO(\Gdims L^3+\Gdims L^2 \log^2(\Gdims \hat{L}))$ to $\bigO(\Gdims \hat{L}^3+L \hat{L}^2+
\Gdims \hat{L}^2 \log^2(\Gdims \hat{L}))$, where $\hat{L} \ll L$. 

Using Nyström approximation not only offers computational gains, but it also brings interesting features from a data accessibility perspective: notice that computing the node-level quantities $H_v$, $H'_v$, $h'_v$ requires access to the full dataset (\Expr{eq:updatehs}), while $H_{\psi,v}$, $H'_{\psi,v}$, $h'_{\psi,v}$ requires only the anchor points and the available samples at that node ($\setprev$, $\setpostv$) (\Expr{eq:psi_definition}). In this problem formulation, the update of the vector parameter $\theta_v$ of node $v$ only requires the computation of $\psi(\cdot)$ using node's own local observations, and the use of the parameters of its neighbors $\{\theta_{u}\}_{u  \in \nghood(v)}$. In conclusion, Nyström approximation combined with the optimization schema enables a distributed estimation  of the algorithm at each node and, therefore, limits to only indirect node access to foreign data of other nodes through the parameters $\theta_u$ for $u \in \nghood(v)$. This is appealing for applications with data access restrictions or privacy-preserving requirements.

The remaining important question is how to select the set of anchor points. There are many strategies to address this problem; for example, Kernel PCA \citep{Scholkopf1998}, random sampling \citep{Williams2000,Talwalkar2008}, greedy approaches \citep{Bach2002}, and k-means clustering \citep{Zhang2008}. In this work we use the approach proposed by \cite{Cedric2009} that is based on the \emph{coherence} measure, which is a measure of linear dependency between the dictionary elements. That algorithm builds a dictionary of manageable size and low redundancy, has a low computational cost, and, under mild conditions, it produces good approximations of the whole space. The adaptation of this strategy in our context can be found in \Appendix{appendix:optimization}.

\subsection{POOL: a \emph{no graph} variant} \label{sec:POOL}

One important by-product of the GRULSIF framework and our supporting analysis, is that we can easily derive a reduced \LRE method that disregards the graph, while enjoying all the other major advantages of our non-parametric optimization formulation. This reduction can be easily obtained by setting $W = \zeros{\Gdims\times\Gdims}$, which neutralizes the graph component, and hence the associated terms disappear from \Eq{eq:update_Nyström}. We call this variant as POOL, and its optimization problem is concretely:
\begin{equation}{\label{eq:Nyström_problem_pool}}
\begin{aligned}
\min_{\ThetaG \in \real^{\Gdims \hat{L} }}  & \frac{1}{\Gdims} \sum_{v \in V }  \left( \frac{1-\alpha}{2} \theta_v ^\top  H_{\psi,v} \theta_v + \frac{\alpha}{2} \theta_v ^\top  H'_{\psi,v} \theta_v  -  h'_{\psi,v} \theta_v \right) 
+ \frac{\lambda \gamma}{2} \sum_{v \in V } \norm{\theta_v}^2\!.
\end{aligned}
\end{equation}
This leads to $\Gdims$ independent quadratic problems, which admit a closed form solution: 
\begin{equation}{\label{eq:pool_thetas_learning}}
    \hat{\theta}_v= \frac{1}{\Gdims} \left[ \frac{1}{\Gdims} \left( (1-\alpha)  H_{\psi,v} + \alpha H'_{\psi,v} \right) + \lambda \gamma \id_{\hat{L}} \right]^{-1} h'_{\psi,v}.
\end{equation}

POOL leads to a total computational complexity of $\bigO(\Gdims \hat{L}^3+L \hat{L}^2)$ (the term related with the cost of the CBCGD schema disappears). POOL can be relevant when it is believed that there is no graph behind the observed phenomena at the different locations, or in situations like those detailed in \Sec{sec:theory} where the collaborative estimation may offer little advantage. Moreover, POOL can be seen as a RULSIF variant \citep{Yamada2011}, where POOL's main differences are: i) its hyperparameters are selected jointly 
\wrt to the mean score $\frac{1}{\Gdims} \sum_{v \in V} \ell_v(\theta_v)$, 
while RULSIF selects independently the hyperparameters for each task; 
ii) POOL uses the Nyström dimensionality reduction technique over the full set of observations, while RULSIF uses a simple uniform random sampling at each node \citep{Sugiyama2012}.

\subsection{Hyperparameters selection}
The performance of GRULSIF depend on the penalization constants $\gamma$, $\lambda$, and the hyperparameters of the kernel $\KernelFunc$ (\eg for a Gaussian kernel, that would be only the width $\sigma$).
As in previous works in non-parametric \fdiv estimation, we use a cross-validation strategy \citep{Sugiyama2007,Sugiyama2011,Yamada2011}.  The main difference is that in GRULSIF we aim to minimize the average of the cost function over all the nodes of the graph. Thus, the score used to identify the optimal hyperparameters is: 
\begin{equation}{\label{eq:score_cross_validation}}
    L(\theta,\sigma):= \frac{1}{\Gdims} \sum_{v \in V} \hat{L}_v(\theta,\sigma):=  \frac{1}{\Gdims} \sum_{v \in V}  \left( \frac{1-\alpha}{2} \theta_v ^\top  H_{\psi,v}(\sigma) \theta_v + \frac{\alpha}{2} \theta_v ^\top  H'_{\psi,v}(\sigma) \theta_v  -  h'_{\psi,v}(\sigma) \theta_v \right)
\end{equation}

where $H_{\psi,v}(\sigma)$,$H_{\psi,v}(\sigma)(\sigma)$,$h'_{\psi,v}(\sigma)$ explicit the relationship between these operators and the hyperparameters of the Kernel function $\KernelFunc$.  

At each iteration of cross-validation, we 
select two training sets: $\mathbf{X}'_{\text{train}},\mathbf{X}'_{\text{train}}$ to update $H_{\psi,v}(\sigma)$,$H_{\psi,v}(\sigma)(\sigma)$,$h'_{\psi,v}(\sigma)$. We fix the two hyperparameters $\lambda$ and $\gamma$ to estimate the parameter $\hat{\Theta}(\sigma,\lambda,\gamma)$, which is solution to the  optimization problem: 
\begin{equation}{\label{eq:optimization_model_selection}}
 \hat{\Theta}(\sigma,\gamma,\lambda) = \argmin_{\Theta} \frac{1}{\Gdims} \sum_{v \in V} \hat{L}_v(\theta,\sigma) + \frac{\lambda}{4} \sum_{u,v \in V} \! W_{uv} \norm{\theta_v-\theta_u}^2
+ \frac{\lambda \gamma}{2} \sum_{v \in V } \norm{\theta_v}^2\!
\end{equation}

The solution is found via \Alg{alg:GRULSIF}.

Finally, the parameter  $\hat{\Theta}(\sigma,\gamma,\lambda)$, is used to identify which parameters $\sigma,\lambda,\gamma$ are optimal. We look for the combination of $\sigma^*,\lambda^*,\gamma^*$ which minimizes the expected value of the score \ref{eq:score_cross_validation}. The full implementation details of the model selection procedure are provided in \Alg{alg:model_selection}.

\begin{algorithm}[t]
\small
   \caption{\textbf{--} Model selection procedure for finding GRULSIF hyperparameters\!\!\!\!}\label{alg:model_selection}
\begin{algorithmic}[1]
   \STATE {\bfseries Input:}  $\mathbf{X}, \mathbf{X}'$: the two sets of observations to be used for estimating the \likelihood-ratios; a graph $\G=(V,E,W)$;\\
	\STATE \hspace*{\algorithmicindent}\hspace*{\algorithmicindent}\quad $D$: a precomputed dictionary associated with the chosen kernel, containing $L$ elements; \\
	\STATE \hspace*{\algorithmicindent}\hspace*{\algorithmicindent}\quad
$\#_\sigma, \#_\lambda, \#_\gamma$: parameter grid to explore for values of $\sigma,\lambda,\gamma$; \\
\STATE \hspace*{\algorithmicindent}\hspace*{\algorithmicindent}\quad 
$R$: the number of random splits.
\STATE {\bfseries Output:} $\sigma^*$ the optimal scale parameter for the Gaussian kernel, and the two penalization constants $\lambda^*$ and $\gamma^*$.\\
\vspace{1.3mm}
\hrule
\vspace{1.3mm}
  \STATE Randomly split  $\mathbf{X}$ and $\mathbf{X}'$ into $R$ disjoint subsets  $\{\upright{X}_r\}_{r=1}^{R}$ and $\{\upright{X}'_r\}_{r=1}^{R}$ \hfill 
  \FOR{ each $\sigma \in \#_\sigma$}
  \FOR{each $(\lambda,\gamma) \in \#_\lambda \times \#_\gamma$}
  \FOR{each data subset $r=1,...,R$} 
	\STATE Let $\mathbf{X}'_{\text{train}} = \mathbf{X}' \backslash \upright{X}'_r$, \,$\mathbf{X}'_{\text{test}} = \upright{X}'_r$, and $\mathbf{X}_{\text{train}} = \mathbf{X} \backslash \upright{X}_r$, \,$\mathbf{X}_{\text{test}} = \upright{X}_r$
  \STATE Compute $h'_{\text{train}}(\sigma)$ and $H'_{\text{train}}(\sigma)$ using the observations in $\mathbf{X}'_{\text{train}}$ \hfill(see \Eq{eq:updatehs})
	 \STATE Compute $H_{\text{train}}(\sigma)$ using the observations in $\mathbf{X}_{\text{train}}$  \hfill(see \Eq{eq:updatehs})
   \STATE Find $\hat{\ThetaG}(\sigma,\gamma,\lambda)$, the solution of Problem \ref{eq:optimization_model_selection}. 
   \hfill(see \Alg{alg:GRULSIF}) 
    \STATE Compute $h'_{\text{test}}(\sigma)$ and $H'_{\text{test}}(\sigma)$ using the observations in $\mathbf{X}'_{\text{test}}$ \hfill(see \Eq{eq:updatehs})
	 \STATE Compute $H_{\text{test}}(\sigma)$ using the observations in $\mathbf{X}_{\text{test}}$  \hfill(see \Eq{eq:updatehs})
  \STATE Compute $\hat{L}^{(r)}(\hat{\ThetaG}(\sigma,\gamma,\lambda)) \mydef  \frac{1}{\Gdims} \sum_{v \in V} \hat{L}_v(\hat{\theta}_v(\sigma,\gamma,\lambda))$ using $h'_{\text{test}}(\sigma),H'_{\text{test}}(\sigma),H_{\text{test}}(\sigma)$ \hfill(see \Eq{eq:score_cross_validation})
  \ENDFOR
	\STATE Compute $\hat{L}(\sigma,\lambda,\gamma) \mydef \frac{1}{R} \sum_{r=1}^{R} \hat{L}^{(r)}(\ThetaG(\sigma,\gamma,\lambda))$ \hfill 
  \ENDFOR
  \ENDFOR
\STATE $\gamma^* \mydef  \argmin_{\sigma,\lambda,\gamma}\hat{\ell}(\sigma,\lambda,\gamma)$
\STATE 
\textbf{return} $\sigma^*$, $\lambda^*$, $\gamma^*$
\end{algorithmic}
\end{algorithm}

We can apply a similar approach to find the hyperparameters of POOL.
As POOL ignores the graph structure, we fix $\lambda=1$, and the penalization term related to the norm of each functional $f_v$ will depend just on the parameter $\gamma$ (\Eq{eq:Nyström_problem_pool}). Then, we use cross-validation to identify the optimal values of the hyperparameters of interest $\sigma$ and $\gamma$. The score to rank the different options is as well the one described in \Eq{eq:score_cross_validation}.

Once the anchor points of Nyström approximation and the hyperparameters have been fixed, we can learn the likelihood ratios. For GRULSIF this amounts to estimate the parameter $\hat{\Theta}(\sigma^*,\lambda^*,\gamma^*)$ via \Alg{alg:GRULSIF}, while POOL learns $\hat{\Theta}(\sigma^*,\lambda^*=1,\gamma^*)$, the solution of \Eq{eq:pool_thetas_learning}.

The hyperparameter $\alpha$ requires a more complex discussion. On one hand, it depends on the application of the likelihood-ratio and \PEdiv estimates. It is clear that when $\alpha=1$, the relative likelihood-ratio $r^{\alpha}(x)=\frac{q(x)}{(1-\alpha)p(x)+\alpha q(x)}$ equals one and the \PEdiv $P^{\alpha}(p \Vert q)$ equals zero, independently of $p$ or $\q$. This made it meaningless quantities to quantify the dissimilarity between $p$ and $q$. On the other extremity, if $\alpha=0$, the classical likelihood-ratio $r(x)=\frac{\q(x)}{p(x)}$ becomes the statistic of interest. In this case, $r(\cdot)$ may be an unbounded function, which is a feature that is related to convergence problems in terms of the sample size and numerical instability \citep{Yamada2011}, such phenomena is visible as well in the rates provided in \Theorem{thm:convergence_results}, where the constants depending on $\alpha$ become undefined. The role of $\alpha$ is to prevent this from happening, since it upper-bounds $r^{\alpha}$:
\begin{equation*}
   r^{\alpha}(x)=\left[\frac{\alpha \q(x) + (1-\alpha) p(x)}{\q(x)}\right]^{-1}= \frac{1}{(1-\alpha) r(x) + \alpha} \,\leq\, \frac{1}{\alpha}.
\end{equation*}
The best way to see $\alpha$ is as a way to smooth $p$ in the denominator of $r^{\alpha}$, which is the reference quantity for comparing $\q$ with. In that sense, to build a meaningful and sensitive estimator, we need values of $\alpha>0$, which are still far from $1$. 

This means the optimal value of $\alpha$ will depend on the interplay between the convergence rates of the estimation method being used and the 
performance of the intended application. We detail the first point in the experiments, when we study the sensitivity of GRULSIF and POOL with respect to this parameter.

\begin{algorithm}[t]
\small
   \caption{\textbf{--} GRULSIF: Collaborative and distributed LRE 
	over a graph}\label{alg:GRULSIF}
\begin{algorithmic}[1]
   \STATE {\bfseries Input:} $\mathbf{X}, \mathbf{X}'$: two samples with observations over the nodes of a graph graph $\G=(V,E,W)$;\\
\STATE \hspace*{\algorithmicindent}  $\alpha \in [0,1)$: parameter of the relative likelihood-ratio (\Eq{eq:rel-density-ratio});\\
\STATE \hspace*{\algorithmicindent} $\sigma$, $D$: kernel hyperparameter, and a dictionary containing precomputed set of $\hat{L}$ anchor points associated with a kernel $\KernelFunc:\mathcal{X} \times \mathcal{X} \rightarrow \R$;\\
\STATE \hspace*{\algorithmicindent}  $\lambda$, $\gamma$: constants multiplying the penalization terms;\\
\STATE \hspace*{\algorithmicindent}  $\hat{\ThetaG}^{(0)}$, $tol:$ initialization of node parameters, and tolerated relative error before termination. %
\STATE {\bfseries Output:}  estimated parameters $\hat{\ThetaG} = \Vect(\hat{\theta}_1,...,\hat{\theta}_{\Gdims})$.
\vspace{1.3mm}
\hrule
\vspace{1.3mm}
\FOR{each node $v \in \{1,...,\Gdims\}$}
\STATE Compute $H_{\psi,v}, H'_{\psi,v}$, $h'_{\psi,v}$ \hfill(see \Eq{eq:updatehs}, using $\psi$ instead of $\phi$)
\STATE Compute the learning rate by 
$\eta_{v} = \eigmax{\frac{1-\alpha}{\Gdims} H_{\psi,v} + \frac{\alpha}{\Gdims} H'_{\psi,v} + \lambda d_v \id_{\hat{L}}}$ \hfill (see \Theorem{Th:convergence})
\ENDFOR
\STATE $i=0$
\REPEAT 
\STATE $i=i+1$
\FOR{each node $v \in \{1,...,\Gdims\}$} 
\STATE Update the node parameter $\hthetanode{v}^{(i)}$ \hfill(see \Eq{eq:update_Nyström}  and \Sec{sec:Nyström_dictionary})
\ENDFOR
\UNTIL{$\frac{\norm{\hat{\ThetaG}^{(i)}-\hat{\ThetaG}^{(i-1)}}}{\norm{\hat{\ThetaG}^{(i-1)}}}>tol$} 
\STATE \textbf{return} $\hat{\ThetaG}^{(i)}$
\end{algorithmic}
\end{algorithm}

\section{Experiments}{\label{sec:experiments}}

\subsection{Design and setup}{\label{sec:exp_setup}}

The empirical evaluation of the GRULSIF framework is conducted for the objective of estimating of the likelihood-ratio $r^\alpha_v$ for each node of a given fixed graph. Since $r^\alpha_v$ is an unknown quantity when dealing with real-data, the comparison of the models for this task is feasible only by designing insightful synthetic experiments. 

\begin{table*}[t]
\caption{\textbf{Synthetic scenarios.}{\label{tab:descr_synthetic_scenarios}}
The scenarios are defined by the graph structure they employ and the node-level distributions ($p_v$ and $q_v$) generating the data observations at each node. When the distributions or their parameters remain unchanged between $p_v$ and $q_v$, this is indicated by \scalebox{.85}{$\bullet$}.
}
\newcommand{\algnopar}{}
\footnotesize
\centering
\vspace{-1mm}
\makebox[\linewidth][c]{%
\scalebox{.85}{
\begin{tabular}{c c  c l r c l}
		\cmidrule[0.8pt]{5-7}
      & & &  \multicolumn{4}{c}{\textbf{Node-level hypotheses}} \\
		\cmidrule[0.8pt](r{1mm}){1-2}\cmidrule[0.8pt](r{1mm}){3-4}\cmidrule[0.8pt]{5-7}
     \textbf{Experiment}  & $\X$ & \textbf{Graph} &\textbf{Location} & $p_v$ & \textbf{\vs} & $q_v$ \\
		\cmidrule[0.5pt](r{1mm}){1-2}\cmidrule[0.5pt](r{1mm}){3-4} \cmidrule[0.5pt]{5-7}
		\multirow{3}{*}{\textbf{Synth.Ia}} & \multirow{3}{*}{$\R^1$}	& \multicolumn{1}{c}{\multirow{3}{*}{\raisebox{-2pt}{$\stackrel{\mbox{SBM}}{\stackrel{\mbox{\tiny $4$ clusters,}}{\mbox{\tiny $25$ nodes each}}}$}}}  & $ v \in C_1$ & N$(\mu=0, \, \sigma=1)$  & \vs & $\text{Uniform}(-\sqrt{3}, \, \sqrt{3})$ 
		\\
		&&& $v \in C_2 \cup C_3$ & N$(\mu=0, \, \sigma=1)$  &\vs&$\bullet$ \\
    &&& $v \in C_4$ & N$(\mu=0, \, \sigma=1)$  &\vs  & N$(\mu=1, \, \sigma=\bullet)$\\
		\cmidrule[0.5pt](r{1mm}){1-2}\cmidrule[0.5pt](r{1mm}){3-4} \cmidrule[0.5pt]{5-7}
		\multirow{3}{*}{\textbf{Synth.Ib}} & \multirow{3}{*}{$\R^2$}    & \multicolumn{1}{c}{\multirow{3}{*}{\raisebox{-2pt}{$\stackrel{\mbox{SBM}}{\stackrel{\mbox{\tiny $4$ clusters,}}{\mbox{\tiny $25$ nodes each}}}$}}}  & $v \in C_1\cup C_2$& N$(\mu=(0,0)^{\top}, \, \Sigma_{1,2}=-\frac{4}{5})$ &\vs&$\bullet$ \\
    &&& $v \in C_3$ & N$(\mu=(0,0)^{\top}, \, \Sigma_{1,2}=\phantom{-} \frac{4}{5})$  &\vs&  N$(\mu=\bullet, \, \Sigma_{1,2}=\,0)$\\
    &&& $v \in C_4$ & N$(\mu=(0,0)^{\top}, \, \Sigma_{1,2}= \,\phantom{-}0)$   &\vs& N$(\mu=(1,1)^{\top}, \, \Sigma_{1,2}= \, \bullet)$ \\
		\cmidrule[0.5pt](r{1mm}){1-2}\cmidrule[0.5pt](r{1mm}){3-4} \cmidrule[0.5pt]{5-7}
		\multirow{2}{*}{\textbf{Synth.IIa}}  &	\multirow{2}{*}{$\R^3$}	& \multirow{2}{*}{\raisebox{-2pt}{$\stackrel{\mbox{BA}}{\mbox{\tiny $100$ nodes}}$}}  & $v \in C(u)$ &  N$(\mu=(0,0,0)^{\top}, \, \Sigma_{i,i}=1, \, \Sigma_{1,2}={\textstyle\frac{4}{5}}, \,\Sigma_{3,1}=0)$ 
		&\vs&	N$(\mu=\bullet, \, \Sigma_{i,i}=\bullet, \, \Sigma_{1,2}=\bullet, \,\Sigma_{3,1}=0)$  \\
		&&& $v \notin C(u)$ & N$(\mu=(0,0,0)^{\top}, \, \Sigma_{i,i}=1, \, \Sigma_{1,2}={\textstyle\frac{4}{5}}, \,\Sigma_{3,1}=0)$ &\vs&$\bullet$ \\
		\cmidrule[0.5pt](r{1mm}){1-2}\cmidrule[0.5pt](r{1mm}){3-4} \cmidrule[0.5pt]{5-7}
		\multirow{7}{*}{\textbf{Synth.IIb}} &	\multirow{7}{*}{$\R^3$}	& \multirow{7}{*}{\raisebox{-2pt}{$\stackrel{\mbox{BA}}{\mbox{\tiny $100$ nodes}}$}} & $v \in C(u)$ & N$(\mu=(0,0)^{\top}, \, \Sigma = 10 \id_{2})$ & \vs & Gaussian Mixture $\big(\raisebox{-2pt}{$\stackrel{\mbox{\tiny with equal}}{\mbox{\tiny proportion}}$}\big)$	\\
		&&&&&& \ \ N$(\mu_1=(\phantom{-}0,\phantom{-}0)^{\top}, \Sigma=5 \id_{2})$\\
		&&&&&& \ \ N$(\mu_2=(\phantom{-}0,\phantom{-}5)^{\top}, \Sigma=5 \id_{2})$\\
		&&&&&& \ \ N$(\mu_3=(\phantom{-}0,-5)^{\top}, \Sigma=5 \id_{2})$\\
		&&&&&& \ \ N$(\mu_4=(\phantom{-}5,\phantom{-}0)^{\top}, \Sigma=5 \id_{2})$\\
		&&&&&& \ \ N$(\mu_5=(-5,\phantom{-}0)^{\top}, \Sigma=5 \id_{2})$\\
		&&& $v \notin C(u)$ & \ \ N$(\mu=(0,0)^{\top}, \, \Sigma = 10 \id_{2})$ & \vs & $\bullet$\\
		\cmidrule[0.8pt](r{1mm}){1-2}\cmidrule[0.8pt](r{1mm}){3-4} \cmidrule[0.8pt]{5-7}
\end{tabular}
}
}
\end{table*}

\inlinetitle{Designed scenarios}{.}~Each experiment instance of the four designed fully synthetic scenarios is generated following the three stages below.
\begin{itemize}[topsep=0.4em, itemsep=0.em]
\item[1.] \textbf{Graph structure.} A random graph is generated according to a standard model:\\~{\mySqBullet}~A \emph{Stochastic Block Model} (SBM) with $4$ clusters, each containing $20$ nodes (intra-cluster edge probability: $0.5$; inter-cluster edge probability: $0.01$).\\~{$\mySqBullet$}~A \emph{Barabási-Albert} (BA) model with $100$ nodes (starts with $5$ nodes, and each new node gets connected preferentially with $2$ nodes, \ie with a probability proportional to the current node degree distribution.

\item[2.] \textbf{Structure of nodes' behavior.} A scheme is considered that first specifies if a node $v$ shall experience a change of measure or not ($p_v \neq q_v$ \vs $p_v = q_v$), and then associates the specific \pdfs to it. This is 
a critical feature because, for the collaborative \LRE to be meaningful, nodes' behavior (expressed as in likelihood-ratios) should be explainable by the graph. In each scenario, one of the following two schemes is used:\\~{$\mySqBullet$}~\emph{Cluster-based scheme}: sets the same behavior for all nodes in a cluster. It is used for SBM graphs that exhibit a cluster structure. Clusters are denoted by $C_1,...,C_4$.\\~{$\mySqBullet$}~\emph{Ego-network-based scheme}: picks node $u$ at random, with a probability proportional to its node degree, and then considers that only the nodes in $u$'s $2$-hop ego network, denoted as $C(u)$, shall experience a change of measure. This scheme is used for BA that do not exhibit a particular cluster structure. $C(u)$ is a connected set of nodes, and its complement is denoted by $C(u)^\mycomplement$.
\item[3.] \textbf{Data observations.}~Finally, for each node $v$, an equal number of $n_v=n'_v=n$ (\ie same for all nodes) data observations are generated from each associated $p_v$ and $q_v$.
\end{itemize}%
The four scenarios are summarized in \Tab{tab:descr_synthetic_scenarios}. The generated data observations are $1$-, $2$-, or $3$-dimensional, and this dimensionality is required to be the same for all nodes in each scenario, since the feature space $\mathcal{X}$ and the associated \RKHS are assumed to be the same for all the nodes. The scenarios are designed such that they pose various challenges. As it can be seen in the part of the table related to the node-level hypotheses, $p_v$ and $q_v$ may be different probability models, or the same model with different parametrization. There can be more than one type of change in a scenario; \eg in Synth.Ia, all $p_v$'s are the same Normal distribution, while two clusters ($C_1$ and $C_4$) experience a different change of measure (to Uniform  with same first two moments of a standard normal distribution) or a change in the mean, respectively. In Synth.IIb, the change of measure is a Normal distribution \vs a Gaussian mixture with $5$ equally mixed Normal components, where the overall mean value remains unchanged.

\inlinetitle{Compared \LRE methods}{.}~ We include our POOL variant (\Sec{sec:POOL}) that uses the proposed optimization scheme, but neutralizes the effect of the graph. We also test existing kernel-based \LRE methods built upon \fdiv minimization, namely ULSIF \citep{Sugiyama2011}, RULSIF \citep{Yamada2011}, and KLIEP \citep{Sugiyama2007}. POOL, RULSIF, ULSIF rely on Pearson's \PEdiv; the first two use the relative likelihood-ratio (\Eq{eq:rel-density-ratio}), and ULSIF uses the classical definition (eqv. to setting $\alpha=0$). KLIEP relies on the \KLdiv. \Tab{tab:LRE-competitors} summarizes the compared methods. The way the hyperparameters were fixed is detailed in \Appendix{appendix:hyperparemeters}.


\begin{table*}[t]
\caption{\textbf{\LRE competitors.} All the methods included in our experimental evaluation study.}{\label{tab:LRE-competitors}}
\footnotesize
\centering
\vspace{-1mm}
\makebox[\linewidth][c]{%
\scalebox{.85}{
\begin{tabular}{l c r c l c}
     \toprule
\textbf{Method} & \textbf{Reference} & \textbf{Estimate} & \textbf{\fdiv}  & \textbf{Graph}\\
\midrule
KLIEP   & \cite{Sugiyama2007} & l.-r. & \KLdiv  & No\\
ULSIF   & \cite{Sugiyama2011} & l.-r. & \PEdiv  & No\\
RULSIF  & \cite{Yamada2011}   & relative l.-r. & \PEdiv  & No\\
\cmidruledashed{1-6}\\
POOL    & this work (\Sec{sec:POOL}) & relative l.-r. & \PEdiv  & No\\
GRULSIF & this work & relative l.-r. & \PEdiv  & Yes\\
     \bottomrule
\end{tabular}
}
}
\end{table*}

\inlinetitle{Evaluation measures}{.}
We quantify the performance of each method by an average \emph{Mean Squared Error} (MSE), which in our context we define as the average of node-level MSE between the likelihood-ratio estimates $f_v$ and the real likelihood-ratio $r_v^{\alpha}$.
Formally, this writes:%
\begin{equation}\label{eq:eval_measure}
     P^{\alpha}\left[ [\mathbf{f}-\mathbf{r}^{\alpha}]^2 \right]:=\frac{1}{N} \sum_{v \in V} 
		\ExpecAlphav{[r^\alpha_v-\hat{f}_v]^2(y)}.
\end{equation}
%
The above expected value, $\ExpecAlpha{[f_v-r^{\alpha}_v]^2(y)}$, is computed by averaging $10,000$ independent samples, which were not used during the training phase.

As \Eq{eq:eval_measure} is written, MSE speaks about the whole graph. However, knowing the design of a synthetic experiment
, we can have a more detailed view over the performance on different parts of the problem. More specifically, we can also measure the MSE for specific groups of nodes that should by design experience the same change of measure and hence to lead to the same \fdiv estimate. 

\inlinetitle{Results and Findings}{.}~The experimental results for each of the four designed scenarios are given in \Fig{fig:results_1A}-\ref{fig:results_2B}. At the top of each figure there is a first line plot showing the convergence behavior of the methods in terms of MSE, for the case where the graph size is $N=100$ nodes; like in \Theorem{thm:convergence_results}. The second line plot zooms in to view more clearly the difference between GRULSIF and POOL, hence the effect of using or not the graph in the \LRE. Right below, there is a grid of several box plots, indexed by the graph scale ($N=\{50,100,250,500\}$ nodes) and the used method.
Each box plot details the precision of the estimates in different groups of differently behaving nodes. Depending on the case, the groups may correspond to node clusters $C_1$ to $C_4$, or the subsets $C(u)$ and $C(u)^\mycomplement$. The red and green lines account for the \fdiv each method aims to approximate.


The first thing to notice is that, in most of the experiments, POOL and GRULSIF show a superior performance when compared with all the other methods. Even if POOL disregard the graph structure of the problem as well as ULSIF, RULSIF, and KLIEP, it still shows better convergence behavior than those methods (although slower in Synth.IIb). This carries a clear message: the introduction of a global non-redundant dictionary, Nyström approximation, and the joint hyperparameters selection that we propose, are enough to boost the performance of \LRE methods when multiple sources of information are available and all of them can be approximated by the same \RKHS.  

The second clear message of the results is that, when the smoothness hypothesis is satisfied, exploiting the graph structure leads to an improved convergence compared to the other techniques. The consistent difference between GRULSIF and POOL provide evidence that the geometry of the problem can indeed lead to performance gains.
In fact, the difference of GRULSIF to all other methods is in general more evident when the sample size ($n$) at each node is smaller. When we compare GRULSIF against the other techniques with respect to the quality of the estimates of the respective \fdiv, we can see that it finds estimators with lower bias, especially for nodes where $p_v=q_v$. This bias gets smaller as the sample size increases. 


\newcommand{\scalefactor}{1.3}
\newcommand{\lineplottrim}{10 5.5 5 10}
\newcommand{\boxplottrim}{0 5.5 35 25}
\newcommand{\boxplotbottomtrim}{12 0 20 320}
\newcommand{\boxplotsize}{0.9\linewidth}
\newcommand{\boxplotlinespace}{\fpeval{-0.75 * \scalefactor}em}
\newcommand{\labelmargin}{0em}
\newcommand{\figlabelgap}{1.187em}
\newcommand{\figlabel}[2]{\colorbox{gray!15}{\scriptsize\hspace{#1}#2\hspace{#1}}}

\newcommand{\bracketHeight}{2.9}
\newcommand{\firstFile}{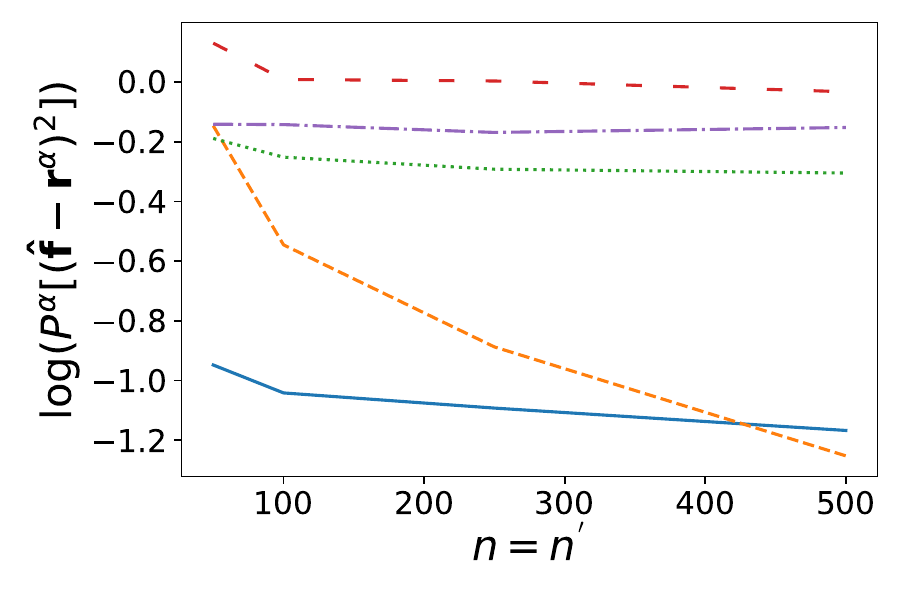}
\newcommand{\secondFile}{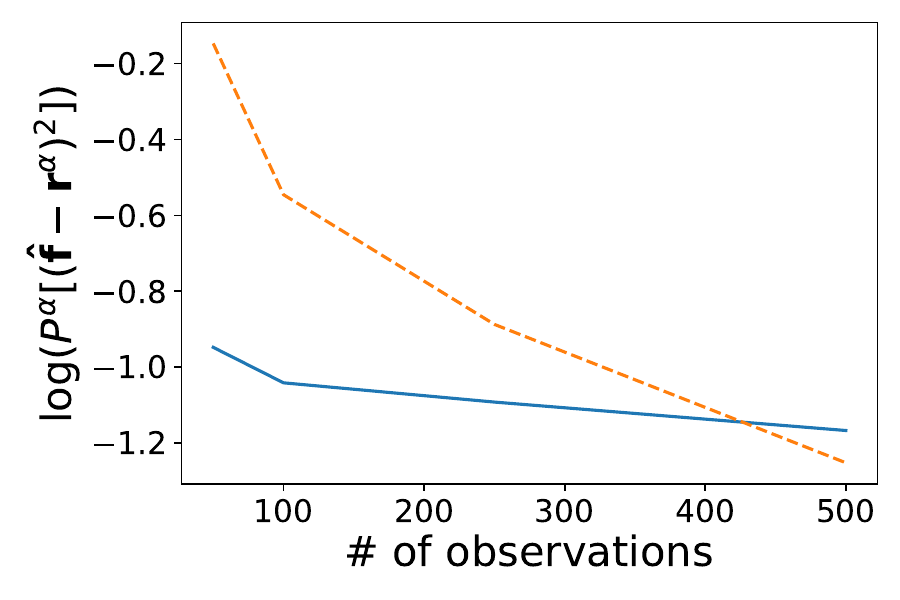}
\newcommand{\thirdFile}{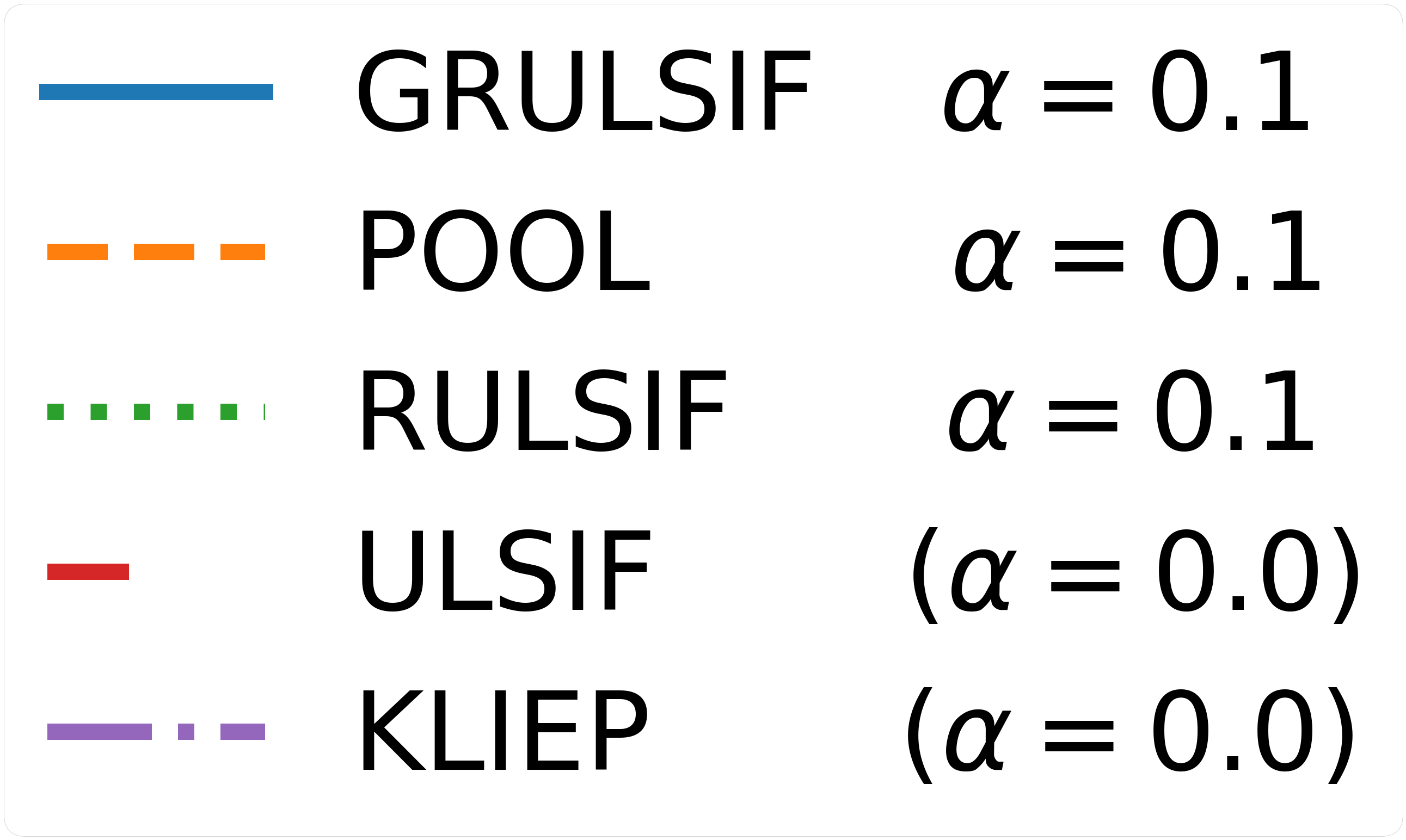}
\newcommand{\fourthFile}{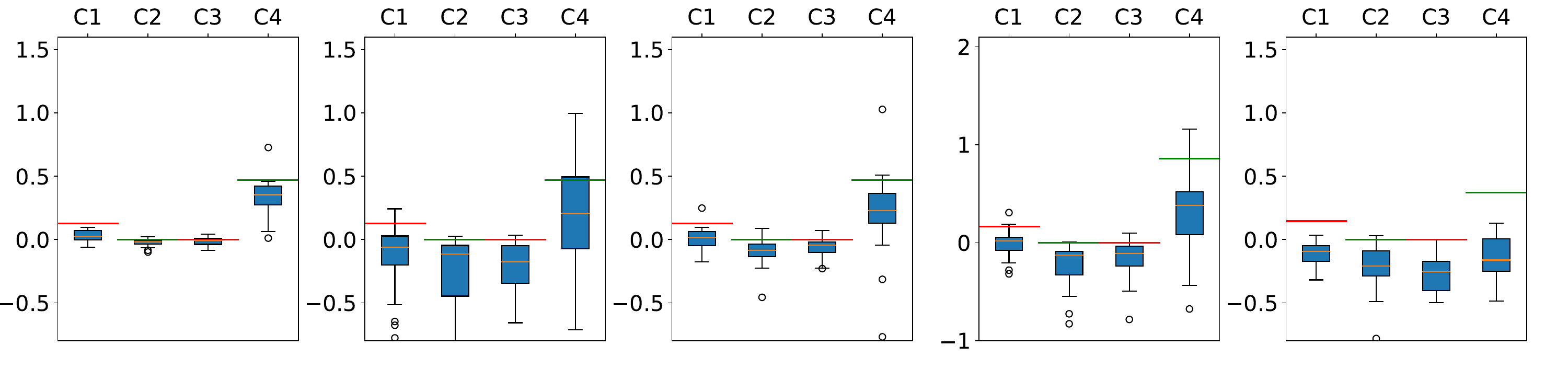}
\newcommand{\fifthFile}{figures/1A_fdiv_boxplot_n_nodes_100_Nref_50.pdf}
\newcommand{\sixthFile}{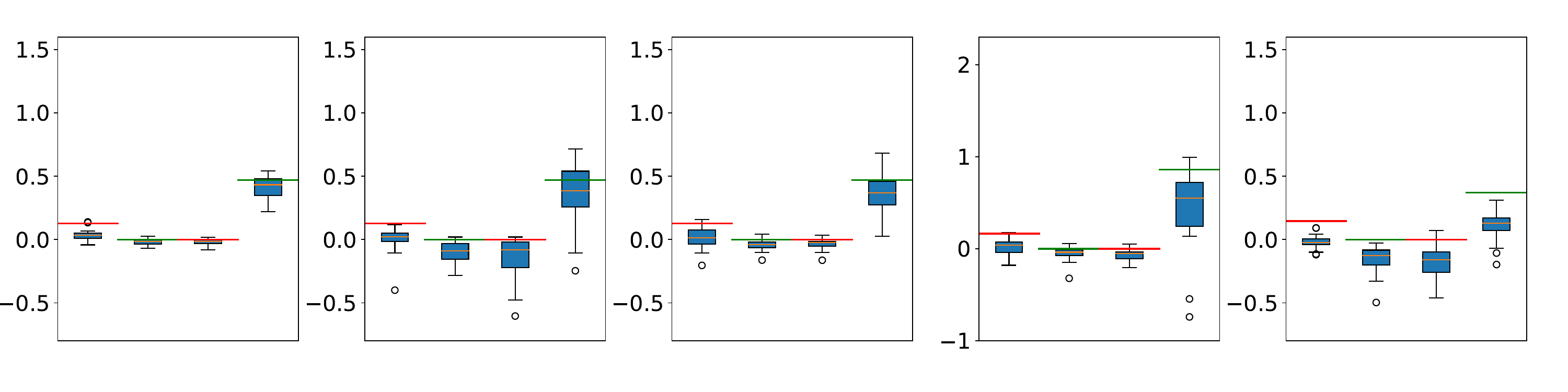}
\newcommand{\seventhFile}{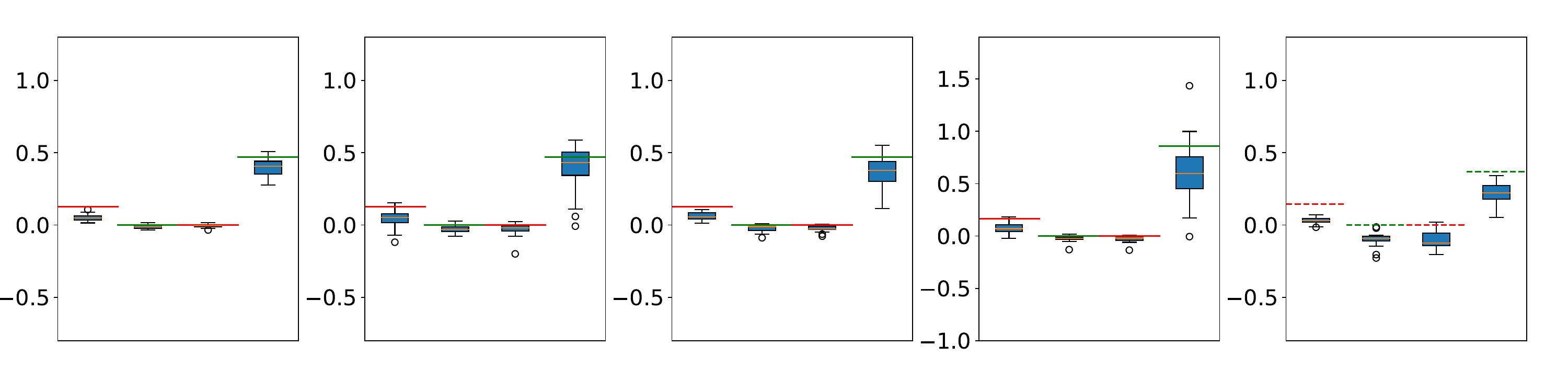}
\newcommand{\eighthFile}{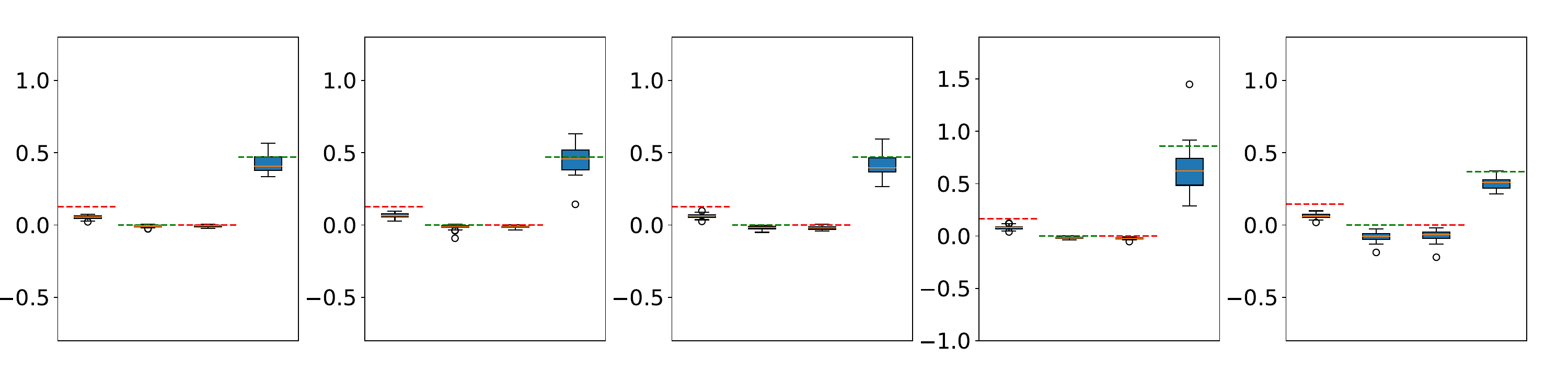}

\begin{figure}
\input{draw_figure.tex}
\caption{\textbf{Experiment Synth.Ia}}\label{fig:results_1A}
\end{figure}

\renewcommand{\bracketHeight}{1.5}
\renewcommand{\firstFile}{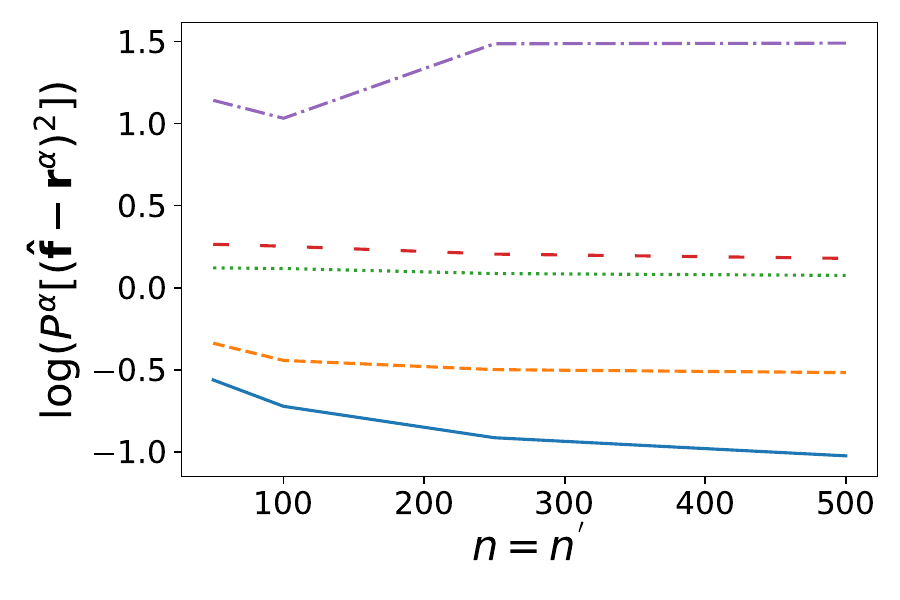}
\renewcommand{\secondFile}{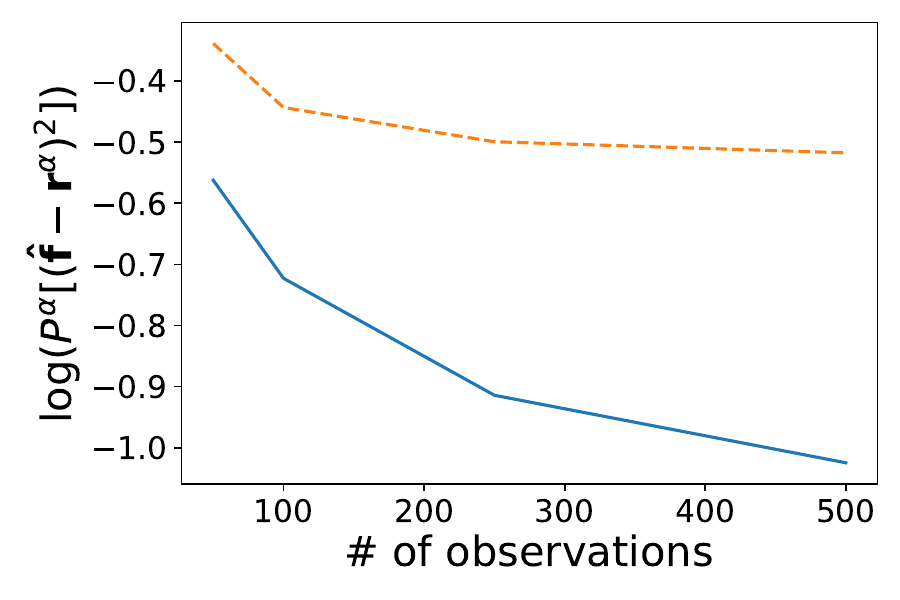}
\renewcommand{\thirdFile}{figures/legend_box.png}
\renewcommand{\fourthFile}{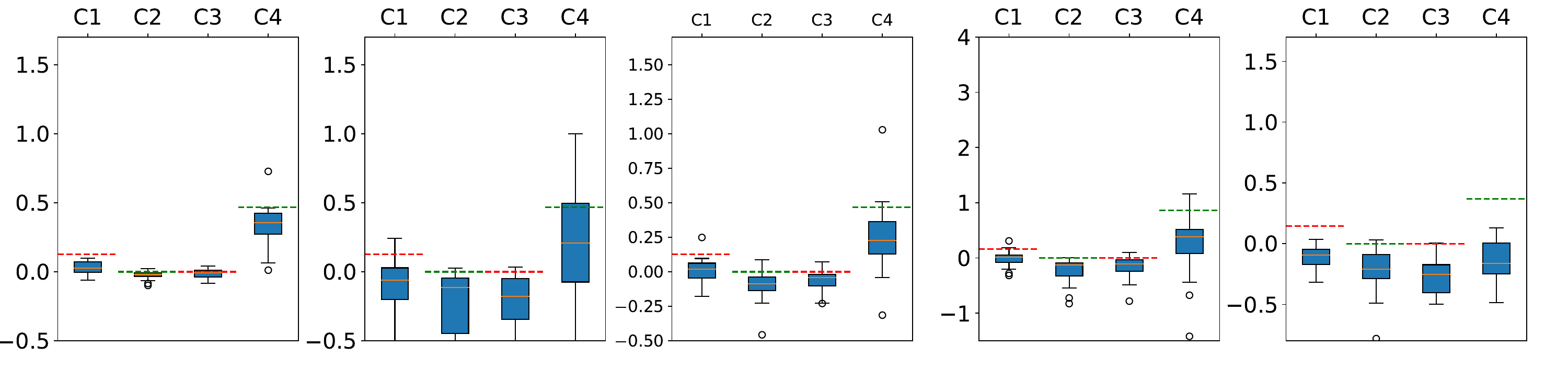}
\renewcommand{\fifthFile}{figures/1B_fdiv_boxplot_n_nodes_100_Nref_50.pdf}
\renewcommand{\sixthFile}{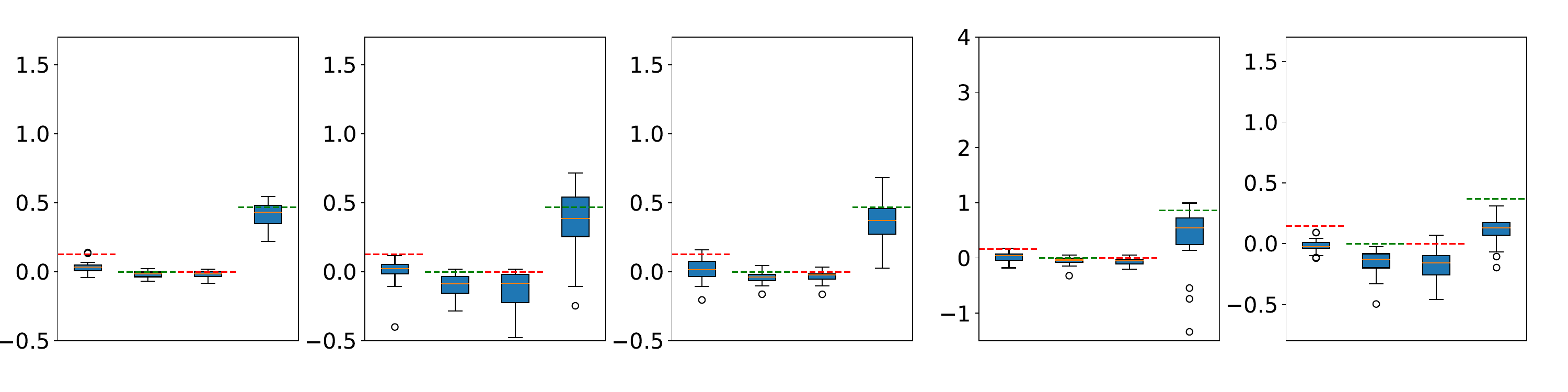}
\renewcommand{\seventhFile}{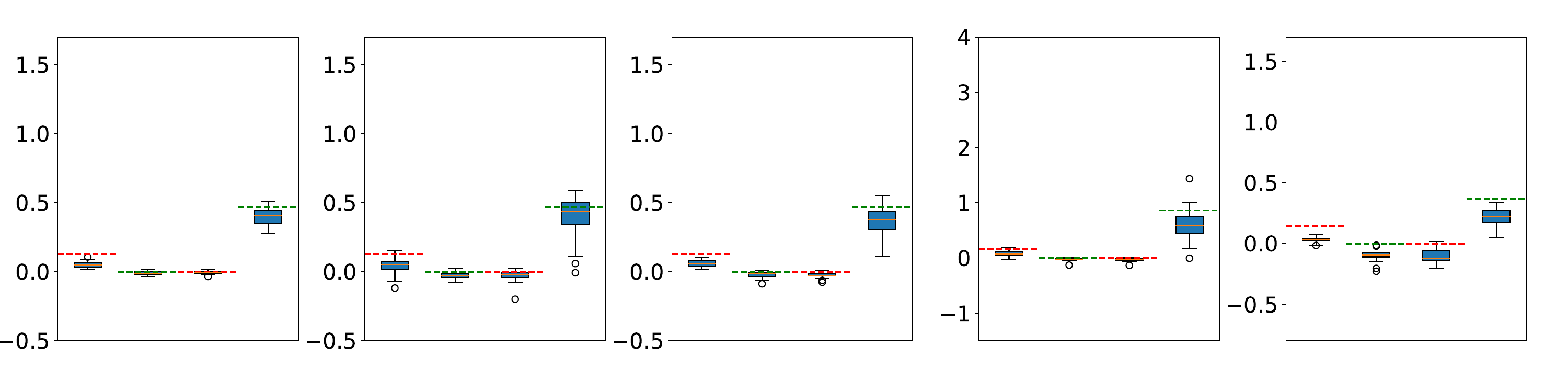}
\renewcommand{\eighthFile}{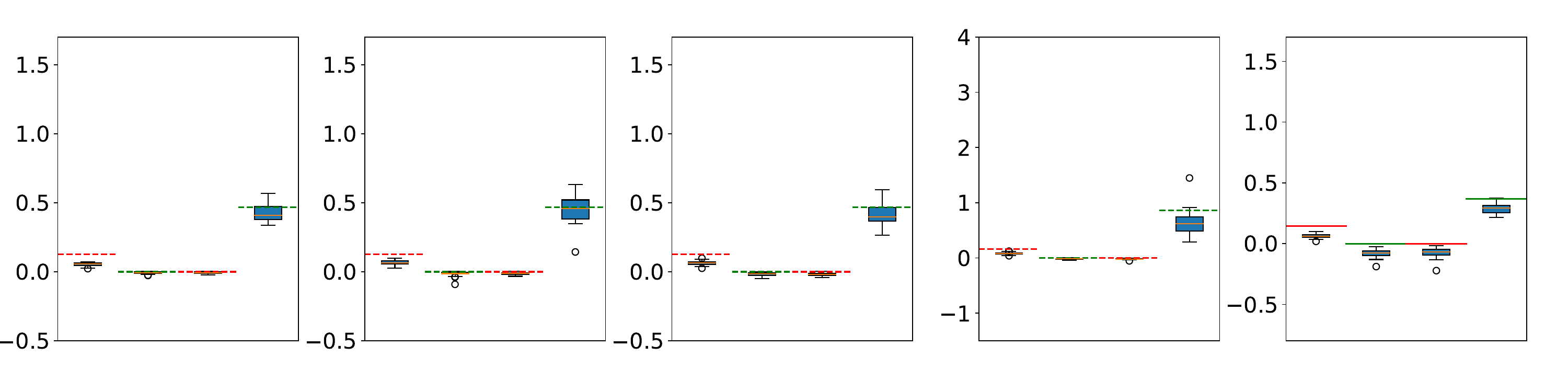}

\begin{figure}
\input{draw_figure.tex}
\caption{\textbf{Experiment Synth.Ib}}\label{fig:results_1B}
\end{figure}

\renewcommand{\bracketHeight}{2.9}
\renewcommand{\firstFile}{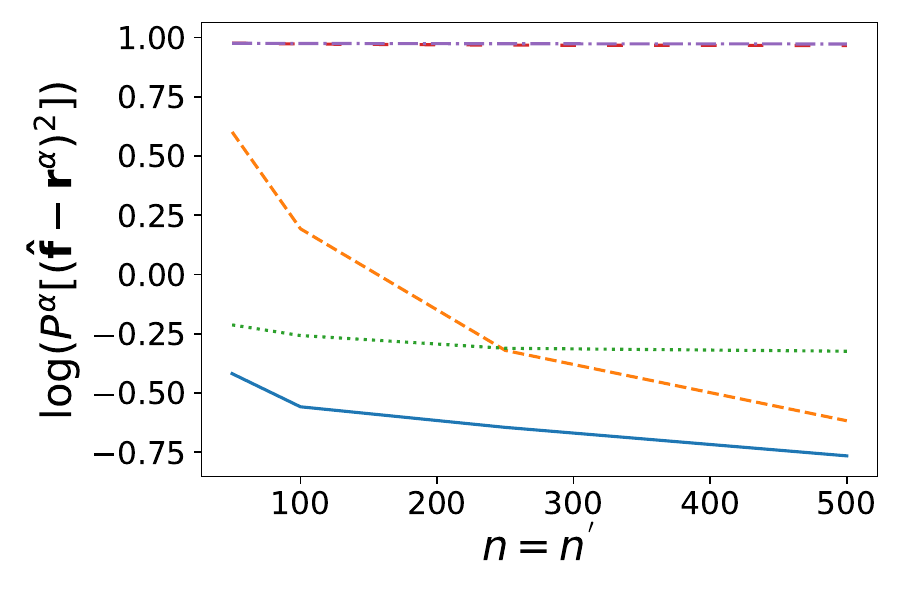}
\renewcommand{\secondFile}{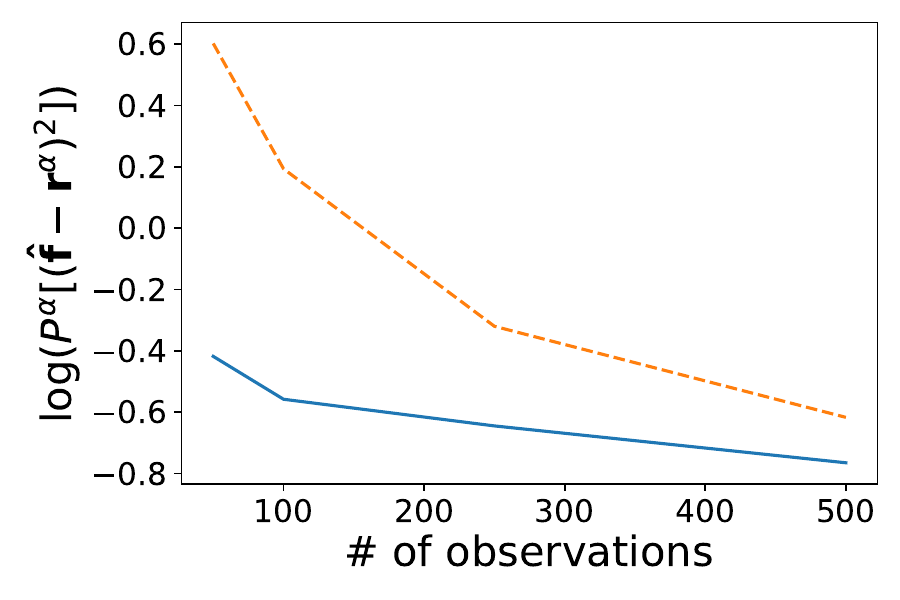}
\renewcommand{\thirdFile}{figures/legend_box.png}
\renewcommand{\fourthFile}{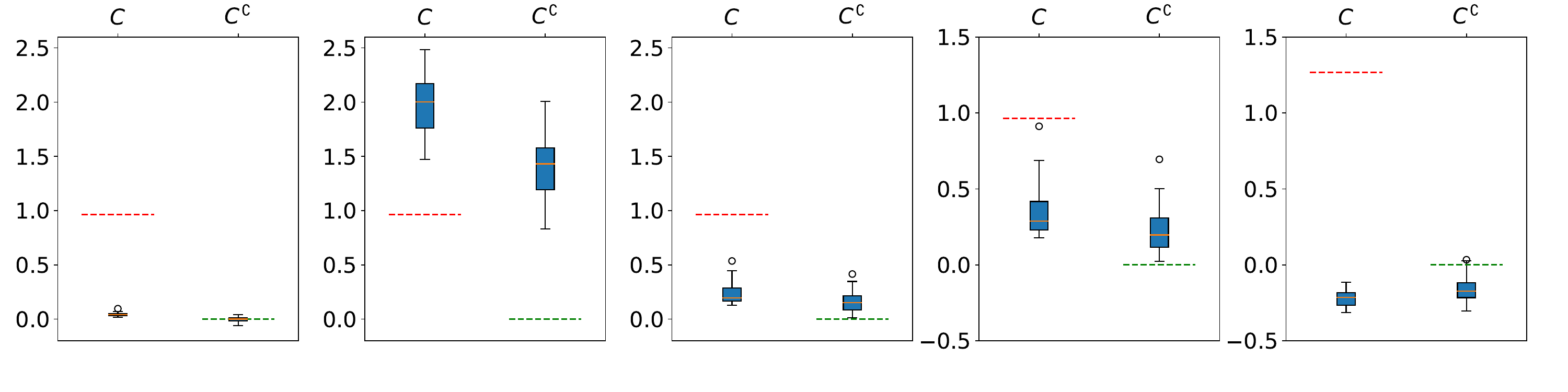}
\renewcommand{\fifthFile}{figures/2A_fdiv_boxplot_n_nodes_100_Nref_50.pdf}
\renewcommand{\sixthFile}{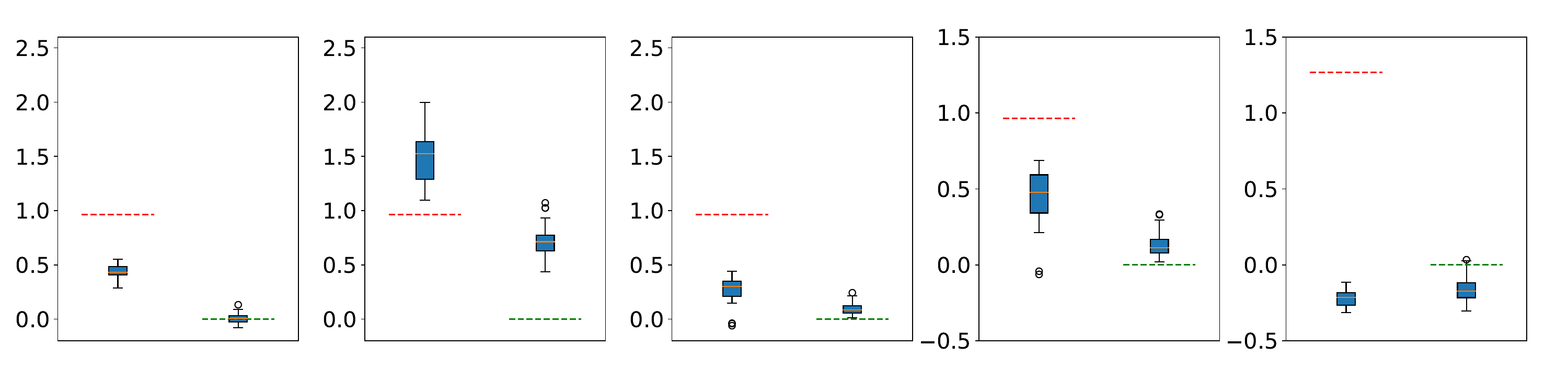}
\renewcommand{\seventhFile}{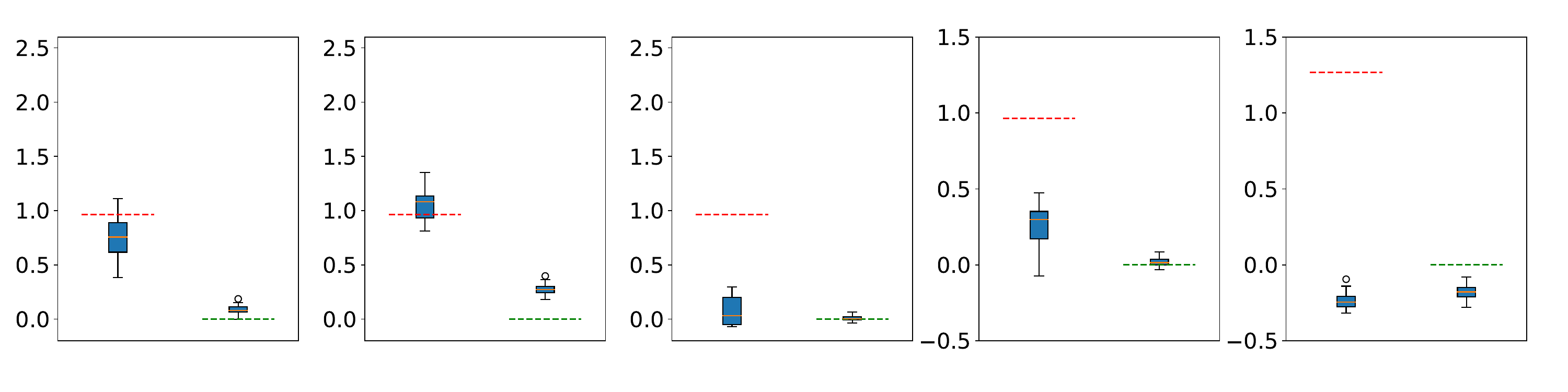}
\renewcommand{\eighthFile}{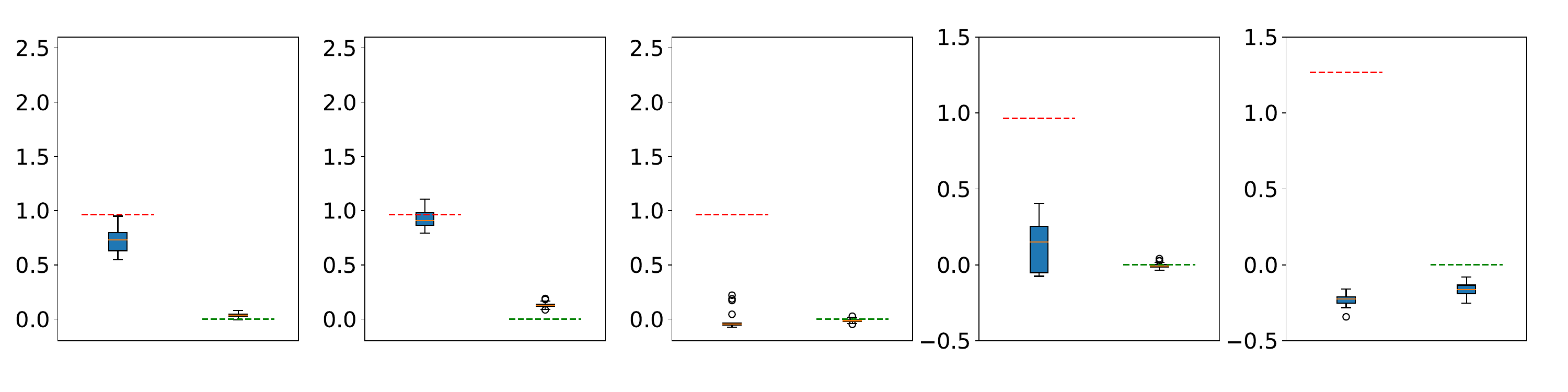}

\begin{figure}
\input{draw_figure.tex}
\caption{\textbf{Experiment Synth.IIa}}\label{fig:results_2A}
\end{figure}

\renewcommand{\bracketHeight}{3.35}
\renewcommand{\firstFile}{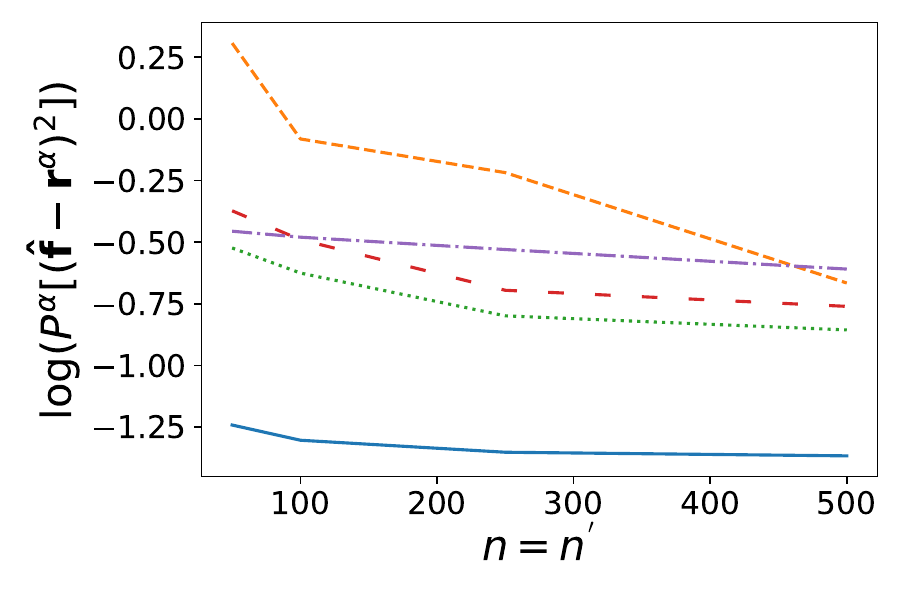}
\renewcommand{\secondFile}{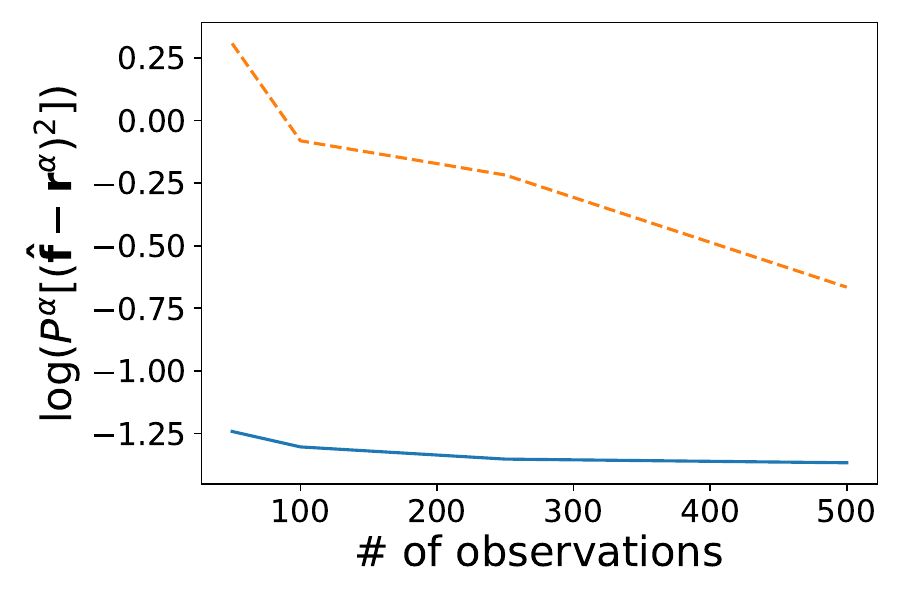}
\renewcommand{\thirdFile}{figures/legend_box.png}
\renewcommand{\fourthFile}{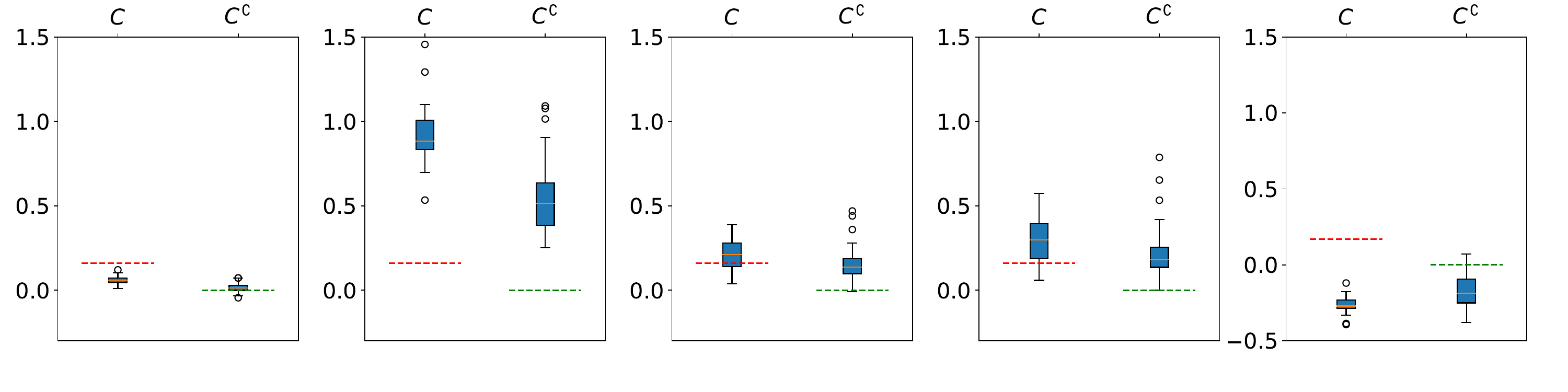}
\renewcommand{\fifthFile}{figures/2B_fdiv_boxplot_n_nodes_100_Nref_50.pdf}
\renewcommand{\sixthFile}{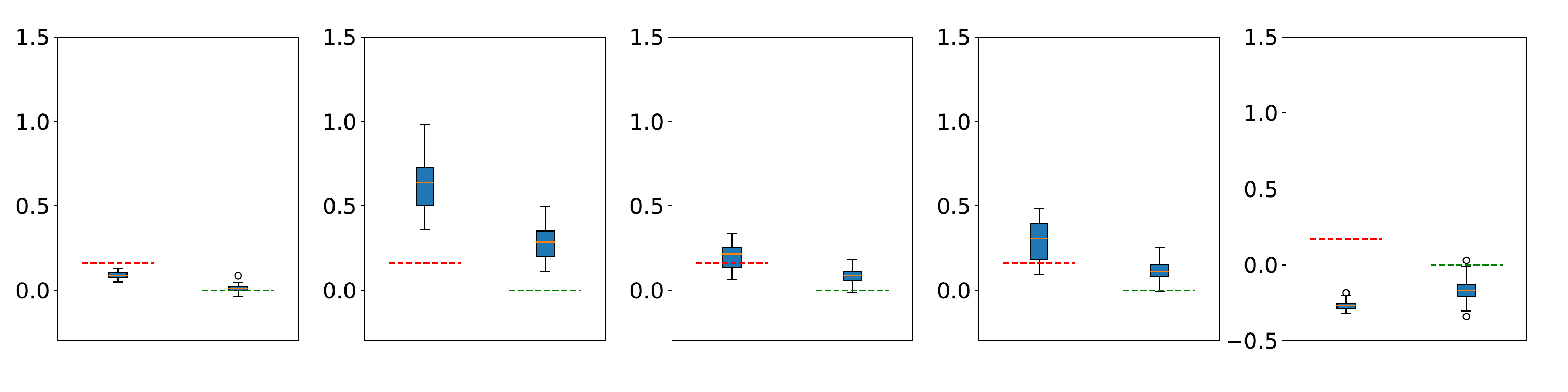}
\renewcommand{\seventhFile}{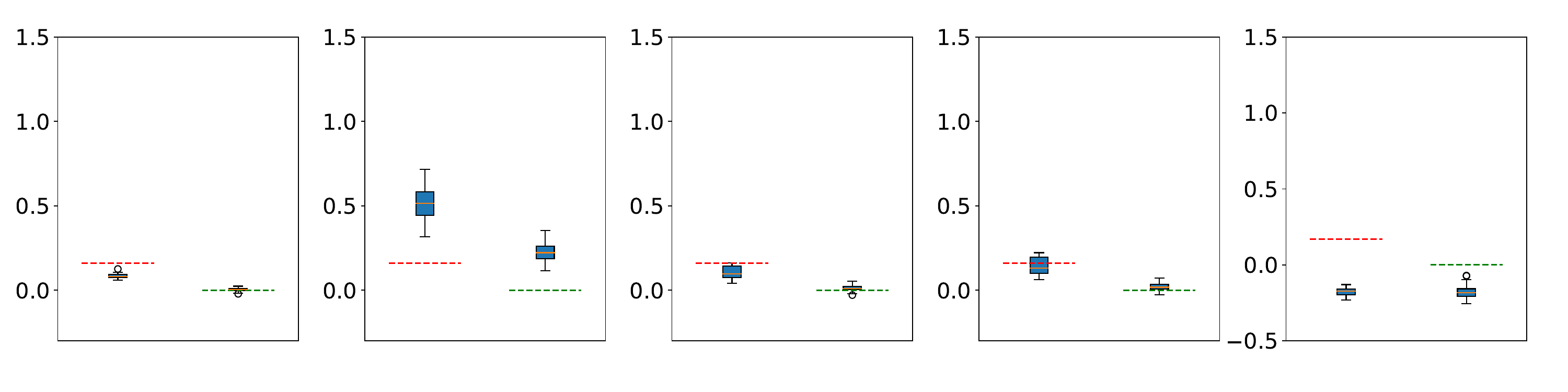}
\renewcommand{\eighthFile}{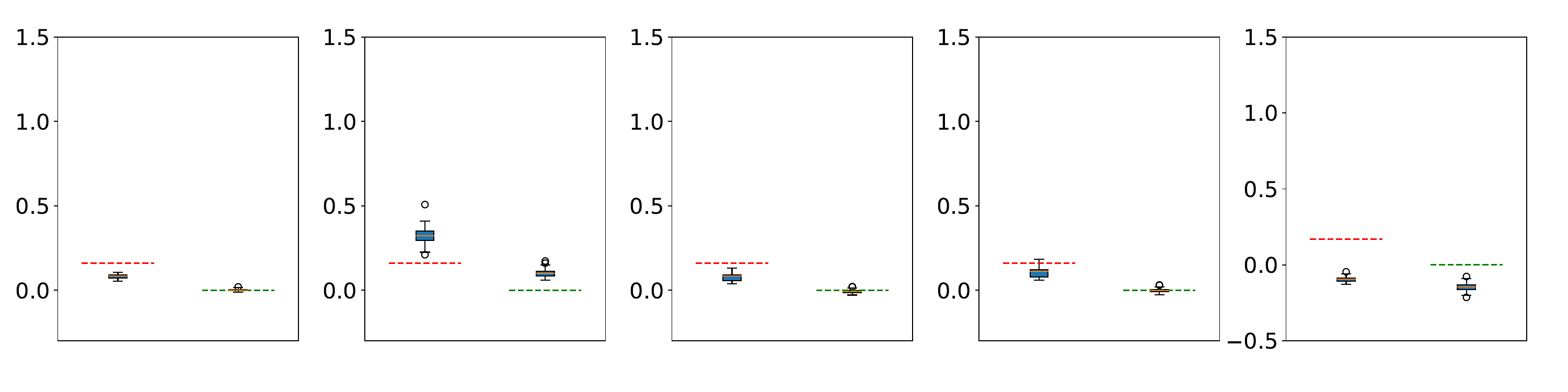}

\begin{figure}
\input{draw_figure.tex}
\caption{\textbf{Experiment Synth.IIb}}\label{fig:results_2B}
\end{figure}

\subsection{The role of $\alpha$ in practice}{\label{appendix:experiments_alpha}}

In this section, we discuss the 
sensibility of GRULSIF and POOL with respect to the parameter 
$\alpha$. We comment over the same experiments presented in \Sec{sec:experiments}. We compare the values $\alpha=\{0.01,0.1,0.5\}$ in the results summarized in \Fig{fig:results_1A_alphas}-\ref{fig:results_2B_alphas}.

In the presented experiments summarized in \Fig{fig:results_1A_alphas}-\ref{fig:results_2B_alphas}, we can point to three interesting findings. First, tuning $\alpha$ leads to different convergence behavior, as suggested by \Theorem{thm:convergence_results}. A lower $\alpha$ value leads to higher bias and variance in the associated \fdiv estimates and a slower convergence rate to the real likelihood ratio. Second, the graph regularization leads to more robust node-level estimates, \ie a lower variance within clusters and faster convergence are observed, especially for nodes where $p=\q$. Third, when $\alpha$ is closer to $1$, meaning the likelihood ratio becomes easier to estimate, the difference between GRULSIF and POOL becomes more evident, confirming the idea that collaborative \LRE leads to improved performance as the node-level tasks become more challenging (see \Sec{sec:theory}).

In conclusion, the choice of the optimal $\alpha$ value is an interplay between the convergence rates of both POOL and GRULSIF as well as the intended application. In particular, for cases where it is required to approximate the unregularized likelihood-ratio, collaborative \LRE would be more relevant as it leads to more robust estimators when $\alpha$ approaches $0$.

\renewcommand{\scalefactor}{1.3}
\renewcommand{\lineplottrim}{10 5.5 5 10}
\renewcommand{\boxplottrim}{0 5.5 35 25}
\renewcommand{\boxplotbottomtrim}{12 0 20 320}
\renewcommand{\boxplotsize}{0.9\linewidth}
\renewcommand{\boxplotlinespace}{\fpeval{-0.725 * \scalefactor}em}
\renewcommand{\labelmargin}{0em}
\renewcommand{\figlabelgap}{1.18em}
\renewcommand{\figlabel}[2]{\colorbox{gray!15}{\scriptsize\hspace{#1}#2\hspace{#1}}}

\renewcommand{\bracketHeight}{1.5}
\renewcommand{\firstFile}{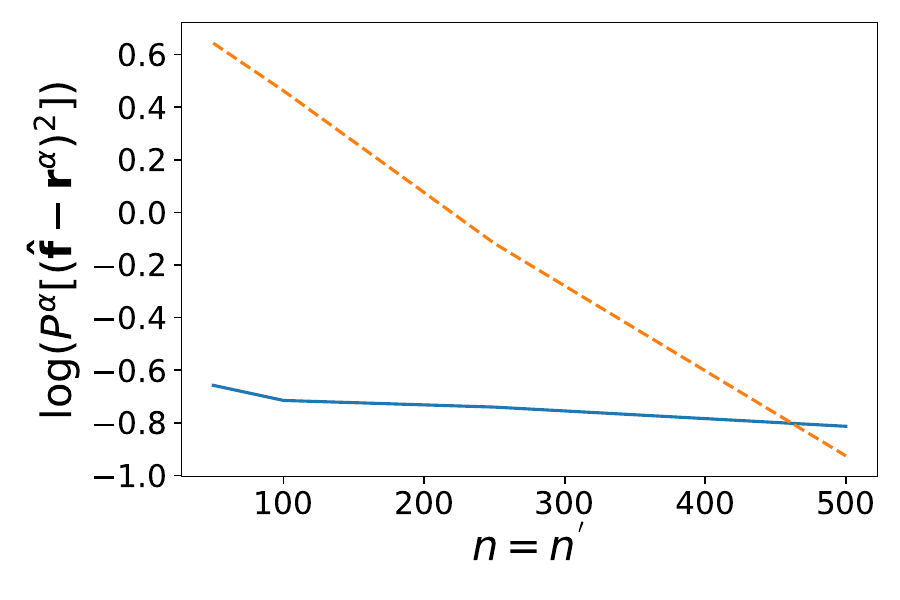}
\renewcommand{\secondFile}{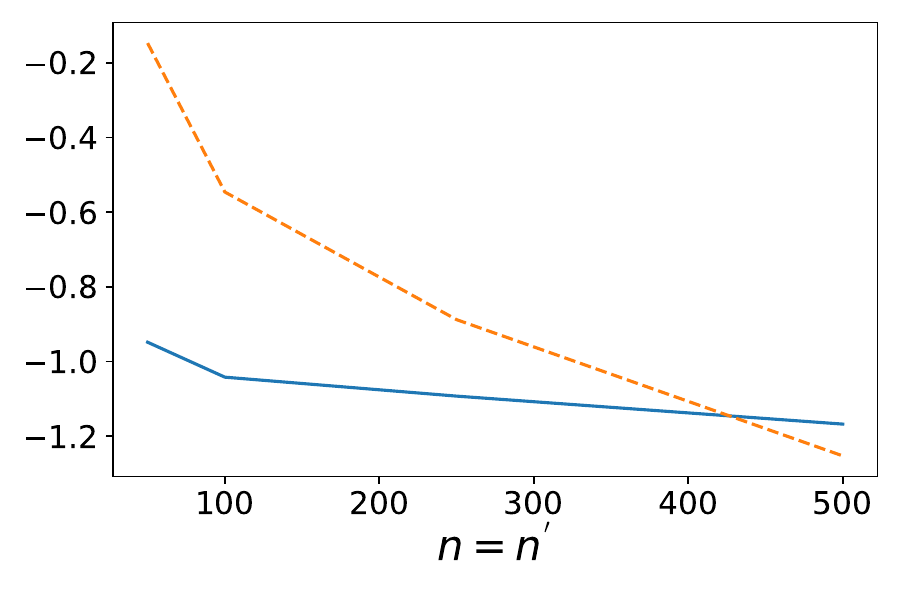}
\renewcommand{\thirdFile}{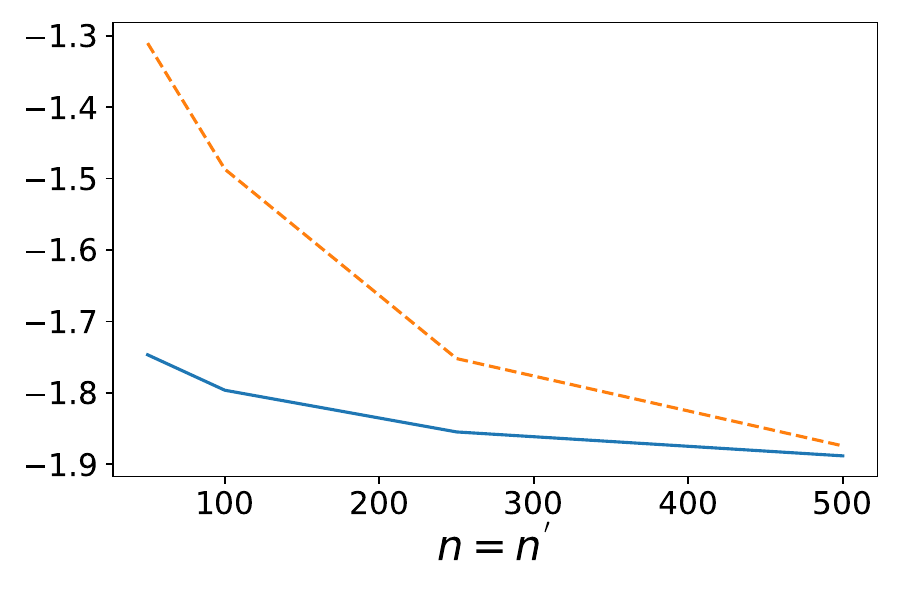}
\renewcommand{\fourthFile}{figures/legend_box_GRULSIF_POOL.png}
\renewcommand{\fifthFile}{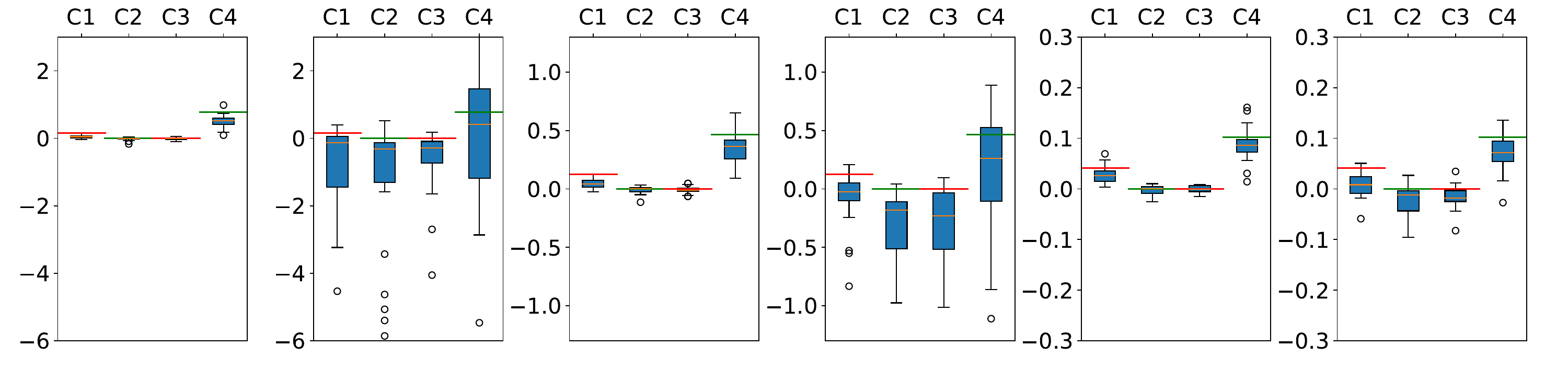}
\renewcommand{\sixthFile}{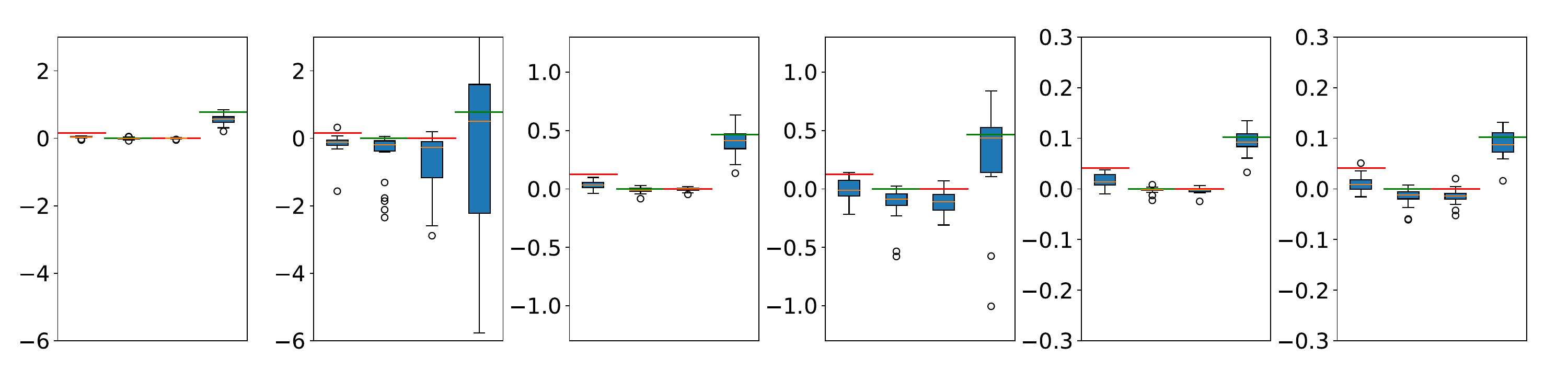}
\renewcommand{\seventhFile}{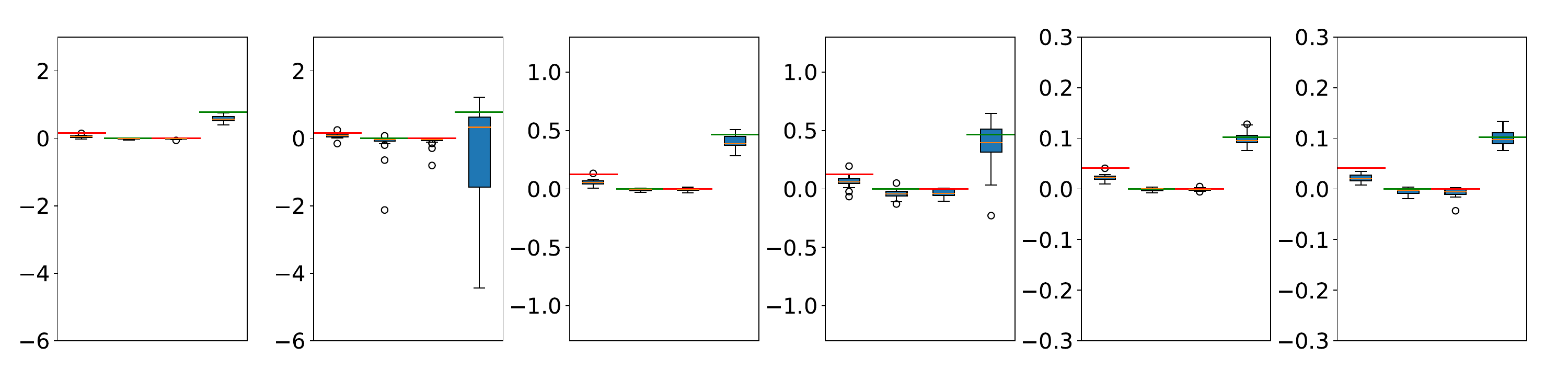}
\renewcommand{\eighthFile}{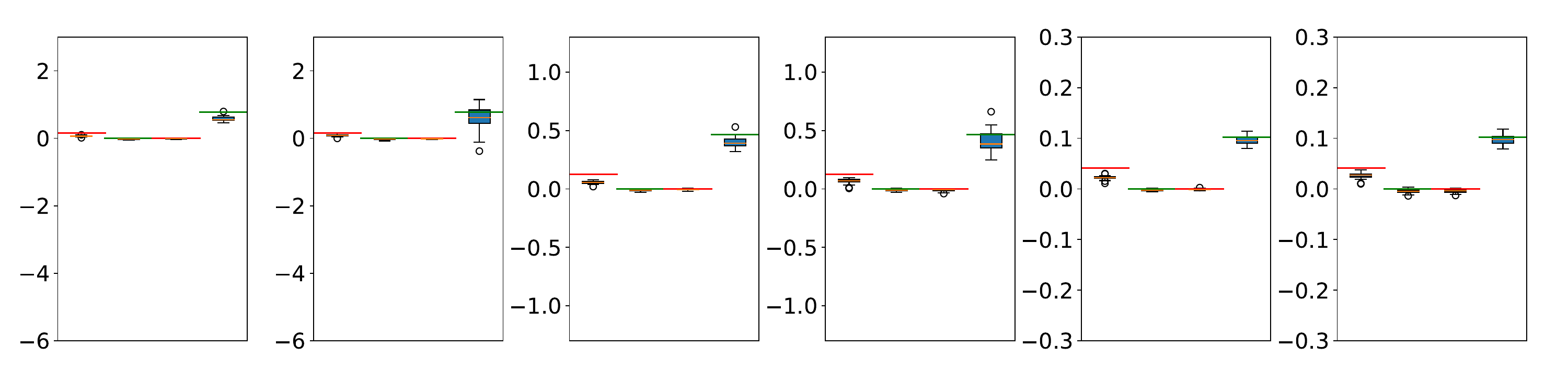}

\begin{figure}
\input{draw_figure2.tex}
\caption{\textbf{Experiment Synth.Ia}}\label{fig:results_1A_alphas}
\end{figure}

\renewcommand{\bracketHeight}{1.5}
\renewcommand{\firstFile}{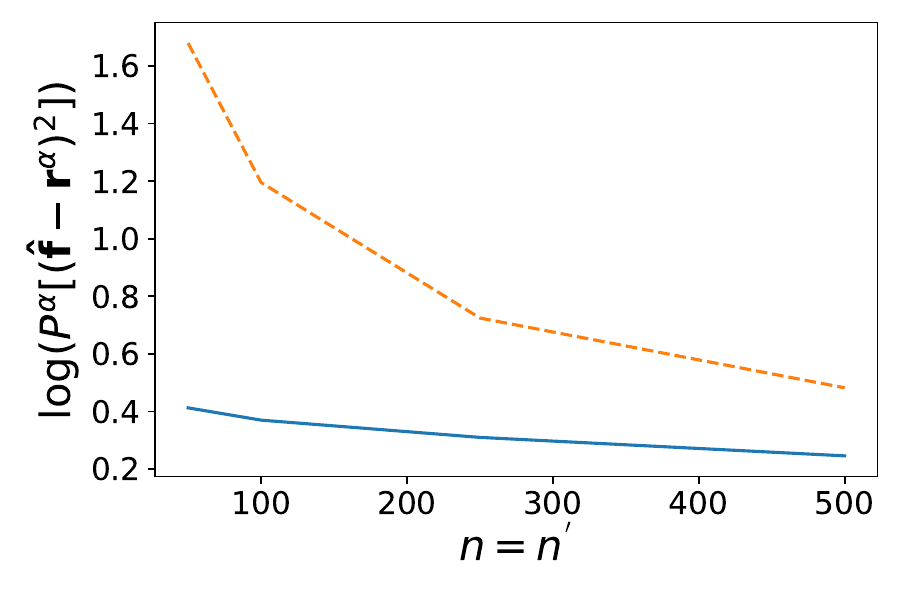}
\renewcommand{\secondFile}{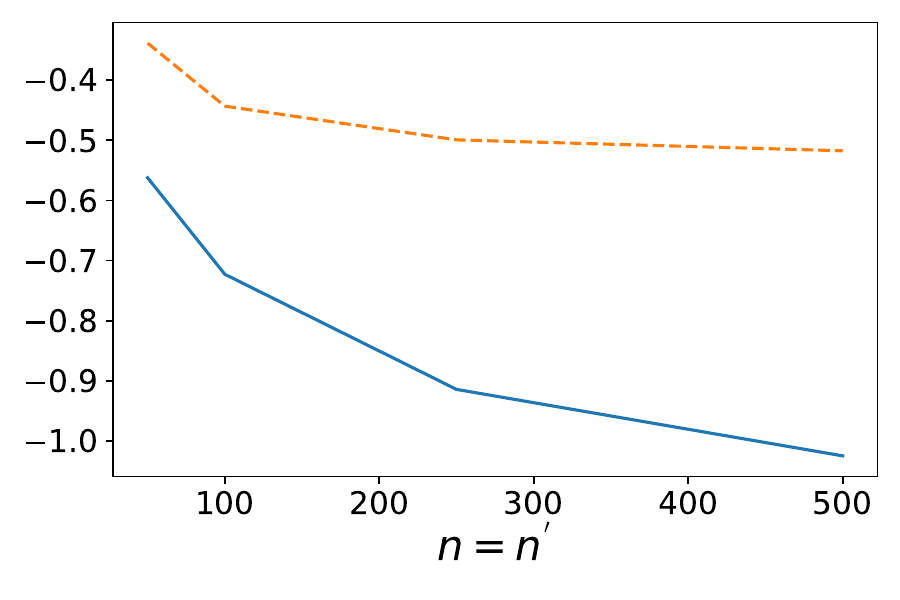}
\renewcommand{\thirdFile}{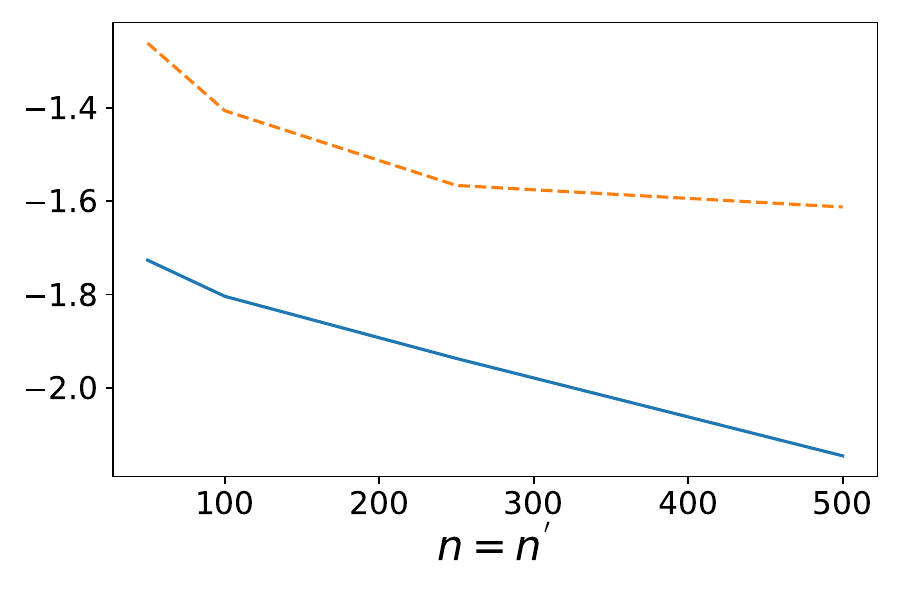}
\renewcommand{\fourthFile}{figures/legend_box_GRULSIF_POOL.png}
\renewcommand{\fifthFile}{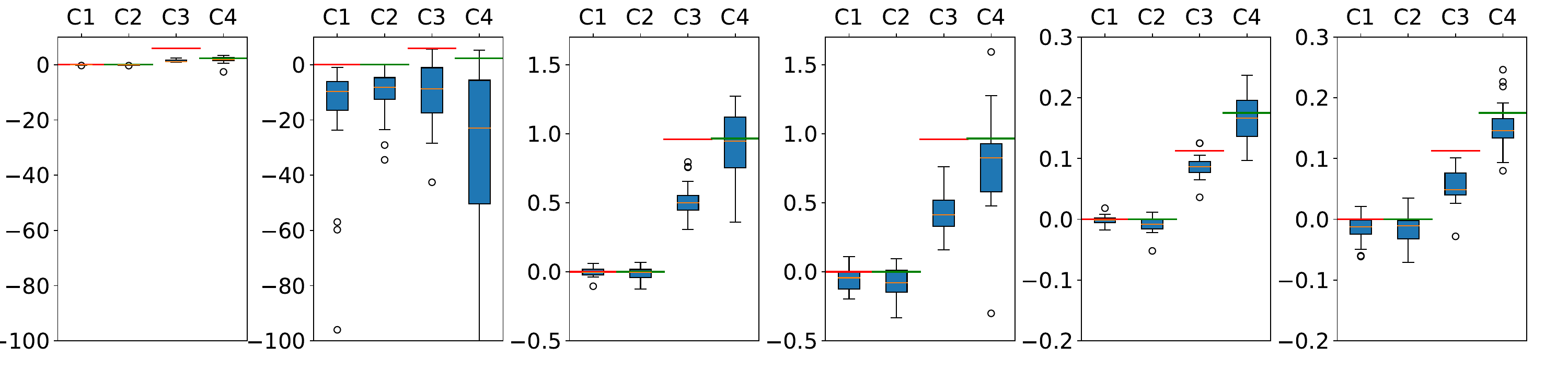}
\renewcommand{\sixthFile}{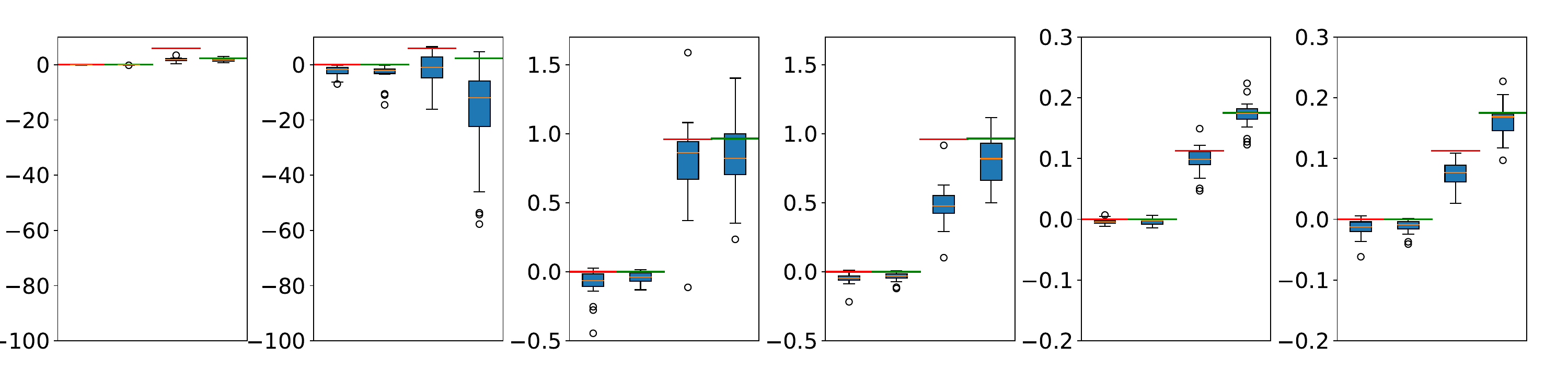}
\renewcommand{\seventhFile}{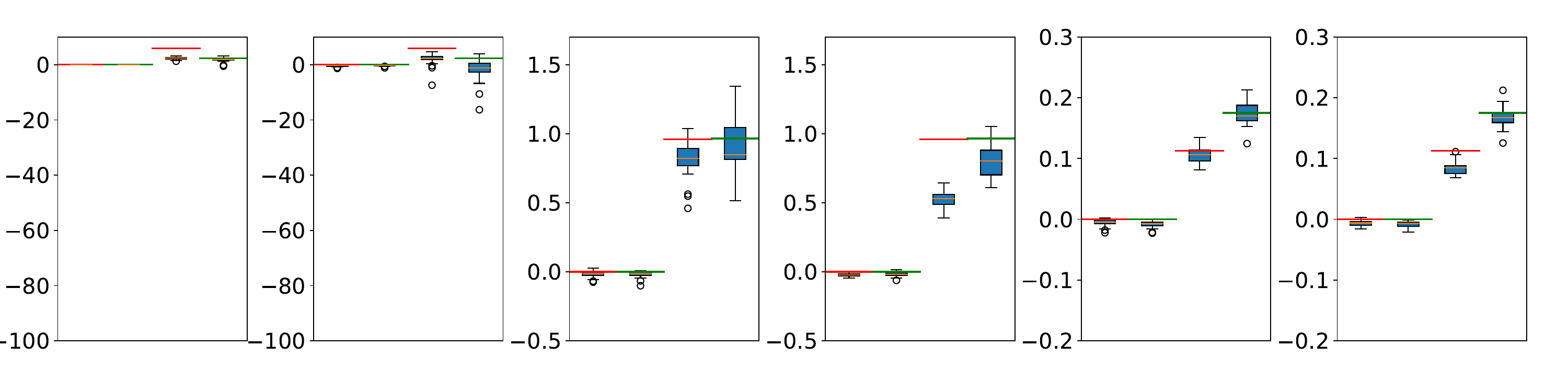}
\renewcommand{\eighthFile}{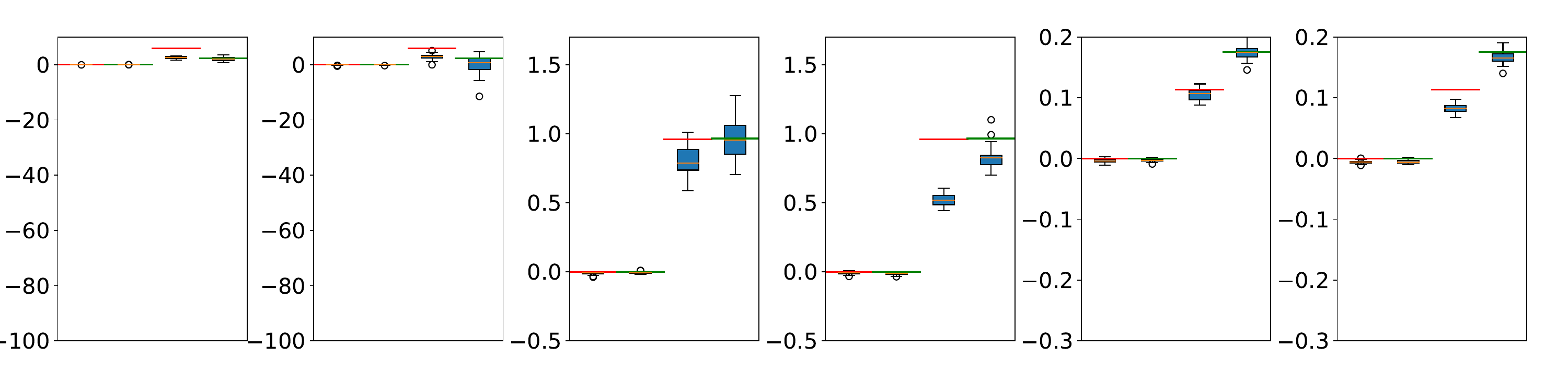}

\begin{figure}
\input{draw_figure2.tex}
\caption{\textbf{Experiment Synth.Ib}}\label{fig:results_1B_alphas}
\end{figure}

\renewcommand{\bracketHeight}{1.5}
\renewcommand{\firstFile}{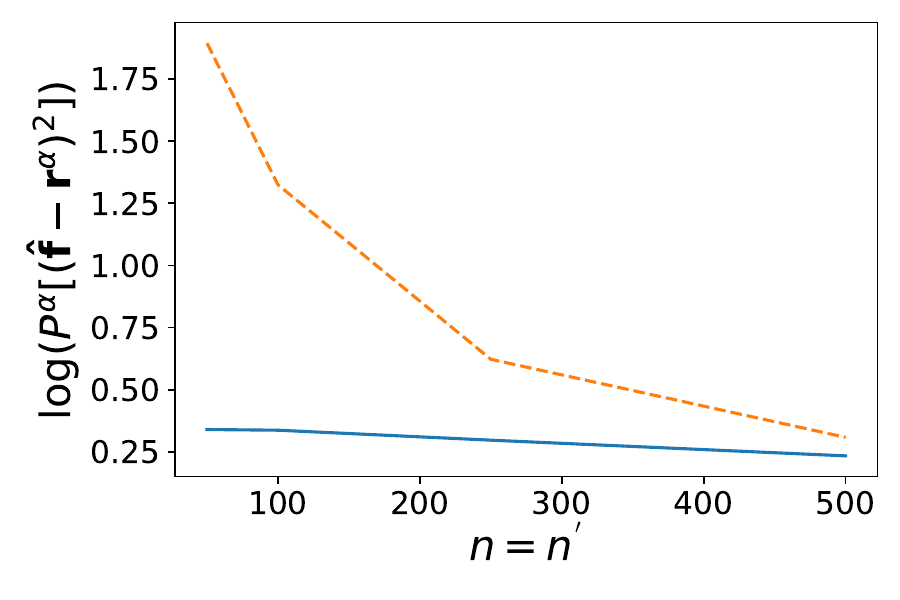}
\renewcommand{\secondFile}{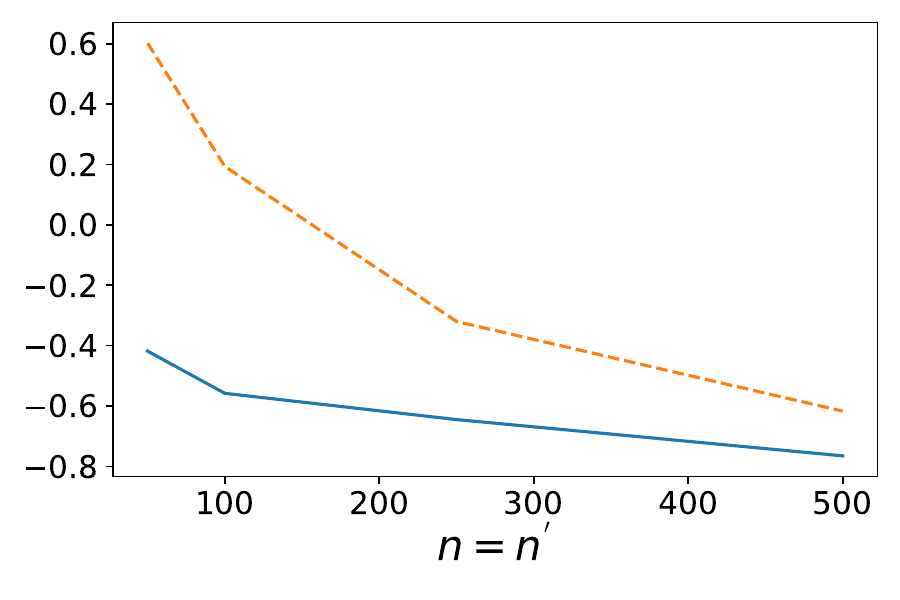}
\renewcommand{\thirdFile}{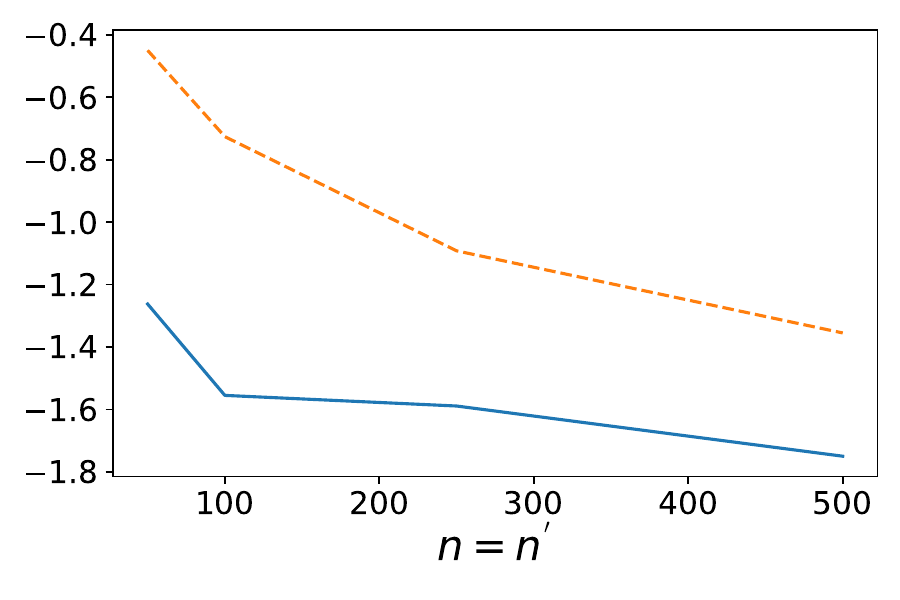}
\renewcommand{\fourthFile}{figures/legend_box_GRULSIF_POOL.png}
\renewcommand{\fifthFile}{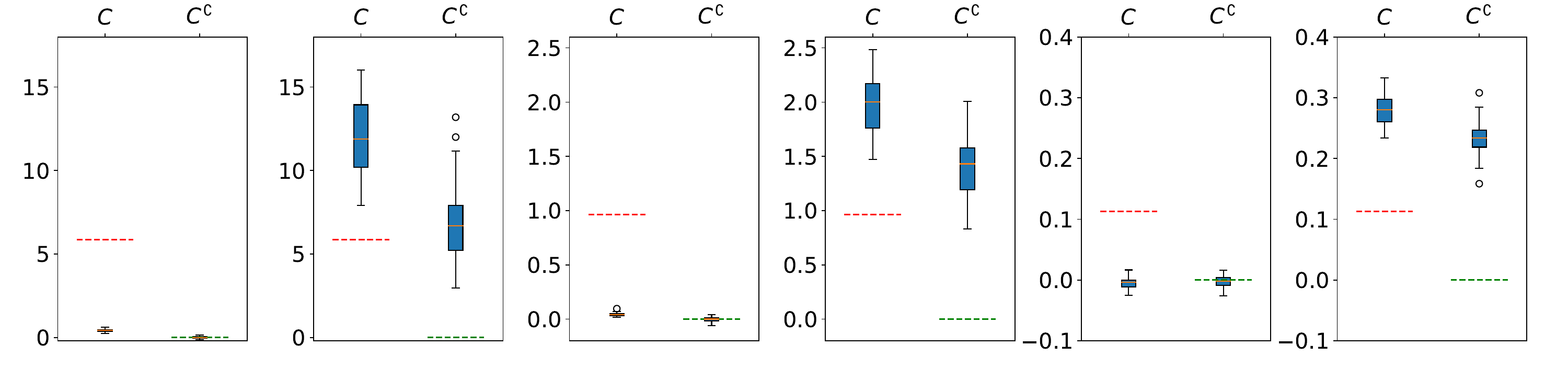}
\renewcommand{\sixthFile}{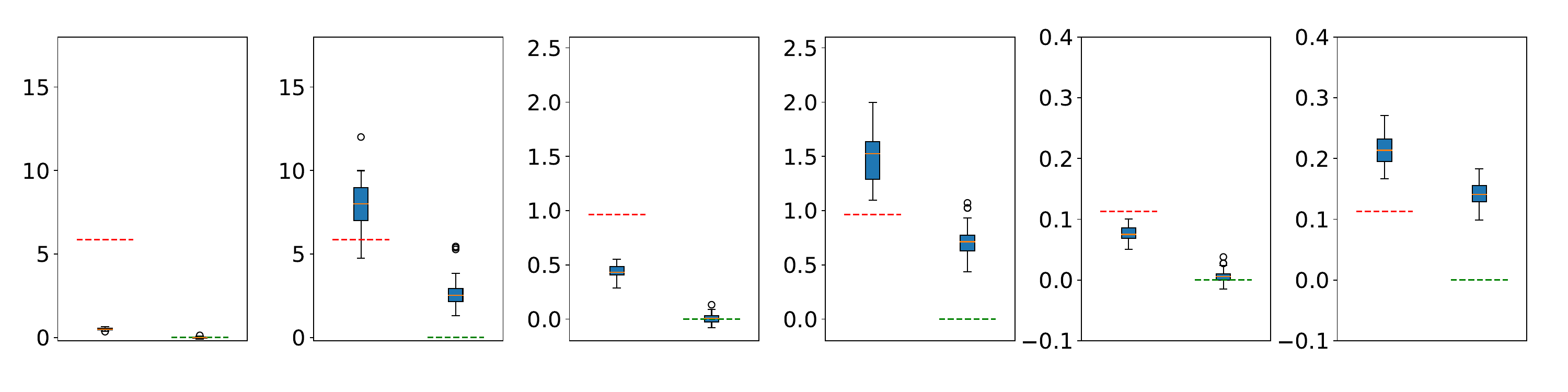}
\renewcommand{\seventhFile}{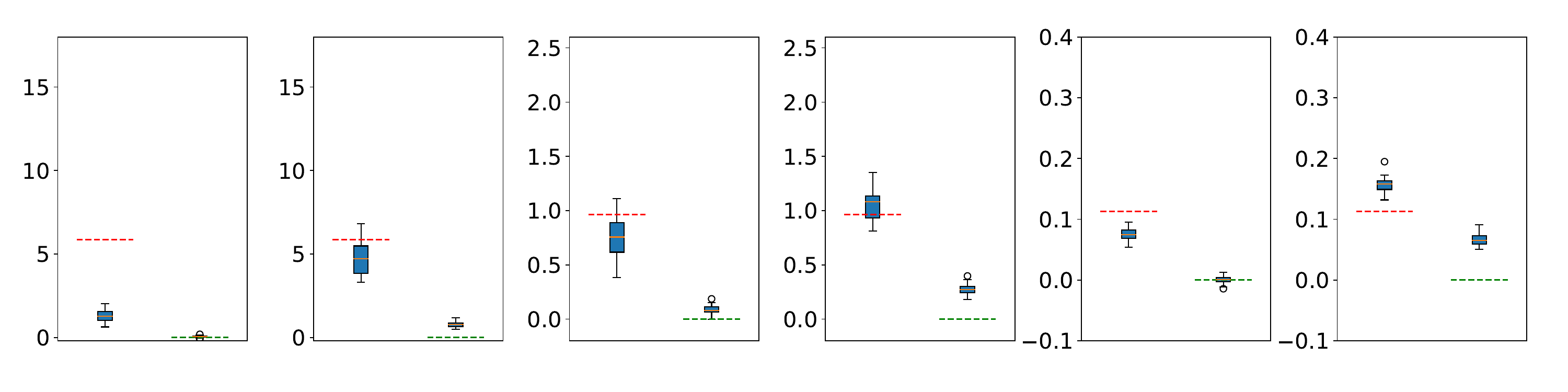}
\renewcommand{\eighthFile}{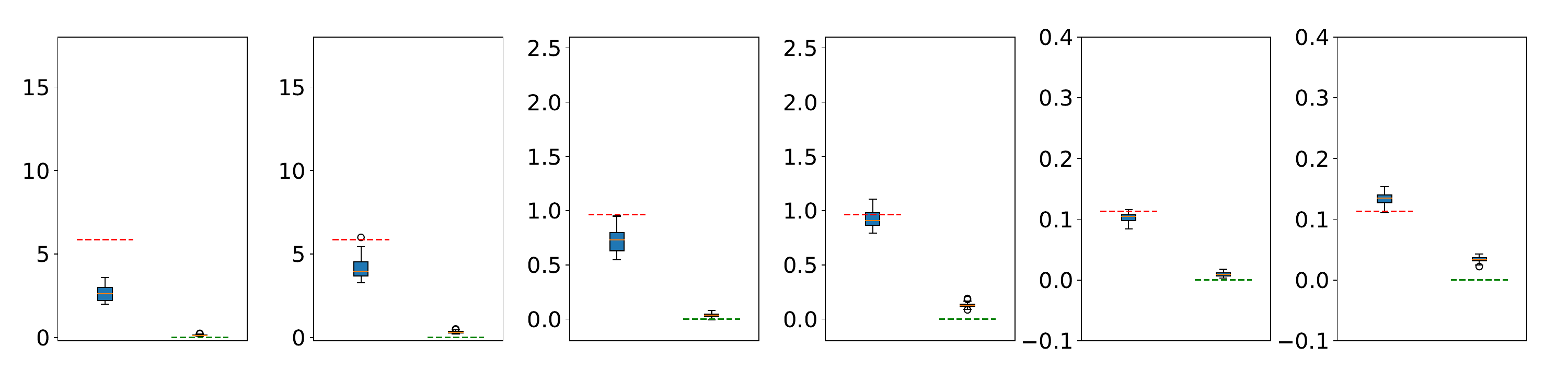}

\begin{figure}
\input{draw_figure2.tex}
\caption{\textbf{Experiment Synth.IIa}}\label{fig:results_2A_alphas}
\end{figure}

\renewcommand{\bracketHeight}{1.5}
\renewcommand{\firstFile}{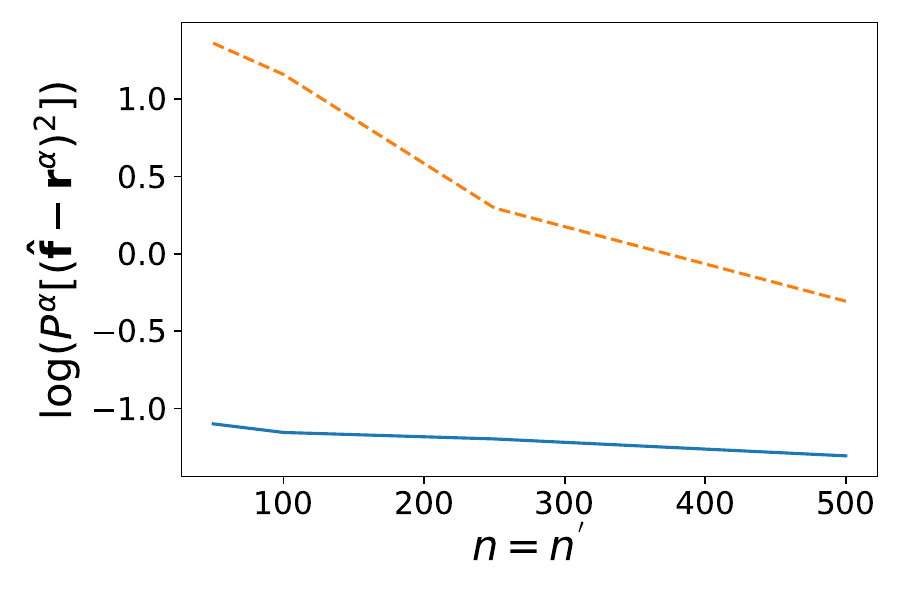}
\renewcommand{\secondFile}{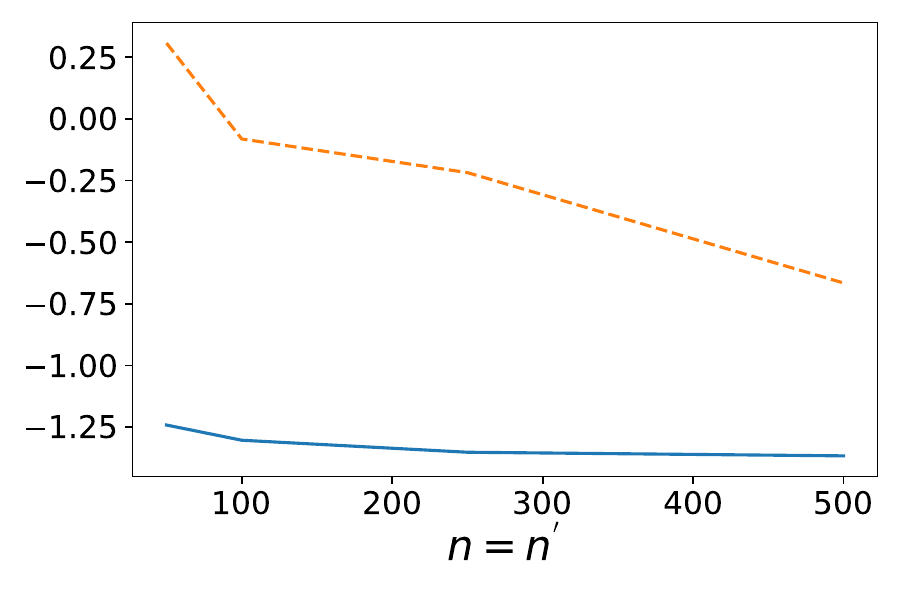}
\renewcommand{\thirdFile}{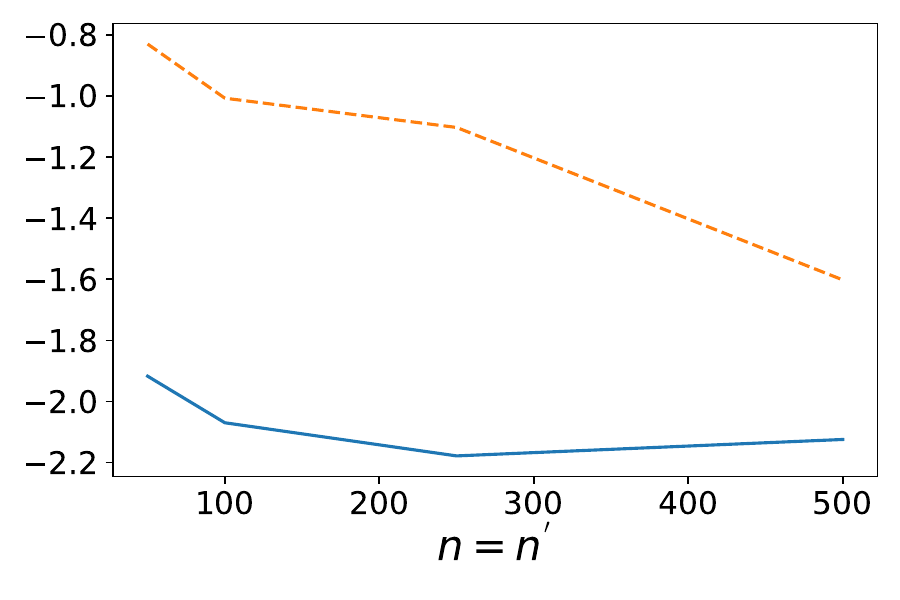}
\renewcommand{\fourthFile}{figures/legend_box_GRULSIF_POOL.png}
\renewcommand{\fifthFile}{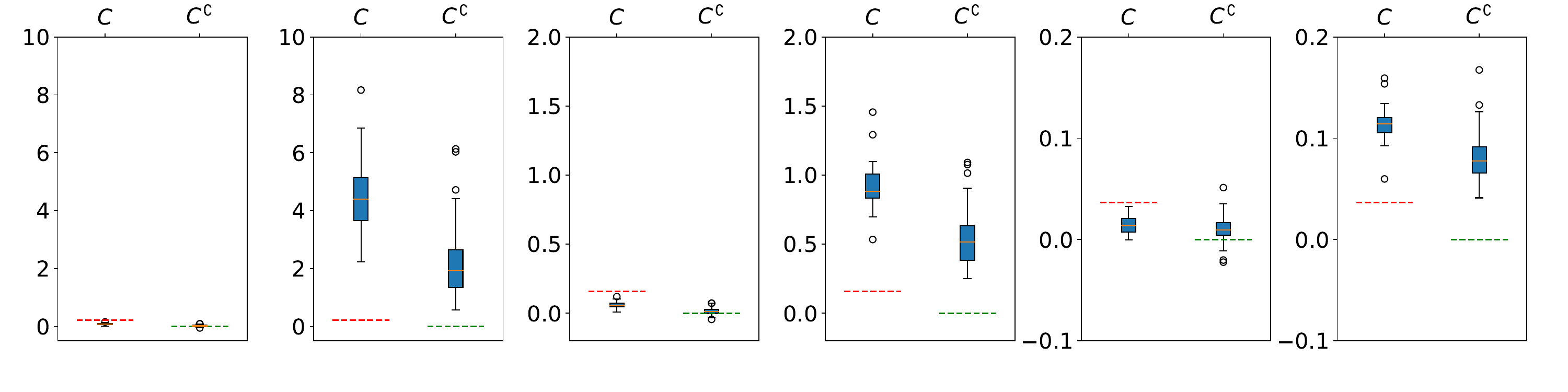}
\renewcommand{\sixthFile}{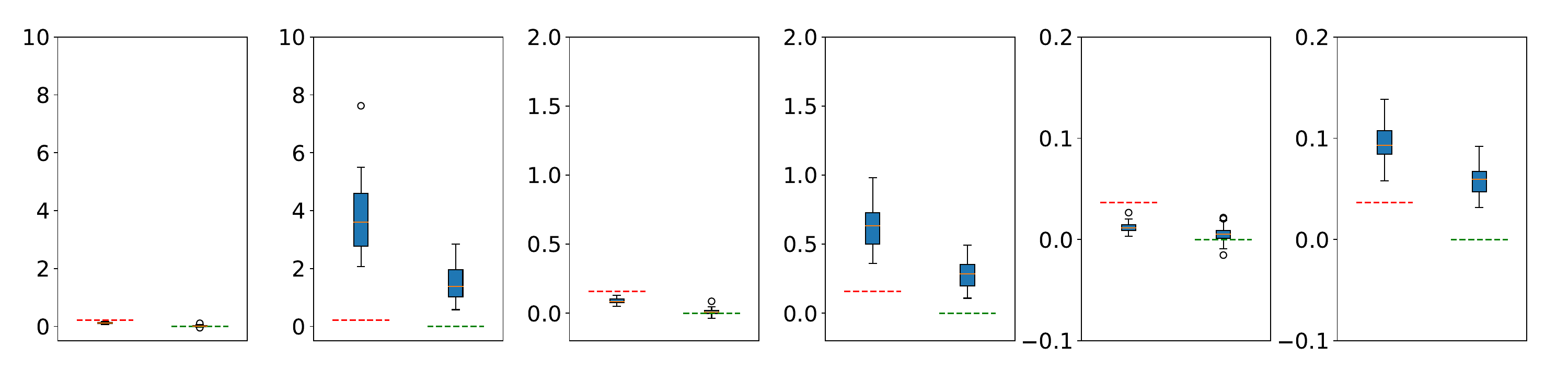}
\renewcommand{\seventhFile}{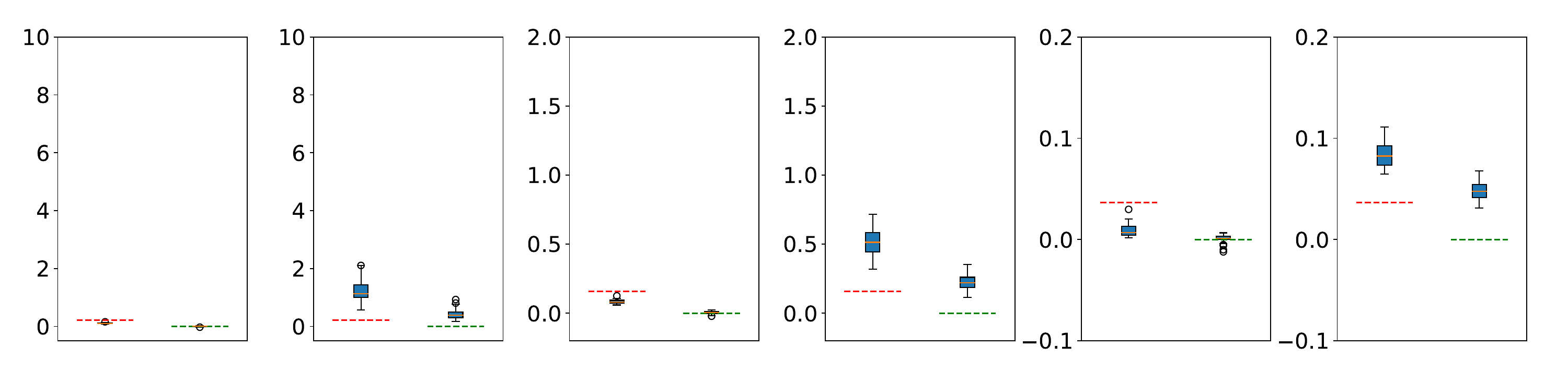}
\renewcommand{\eighthFile}{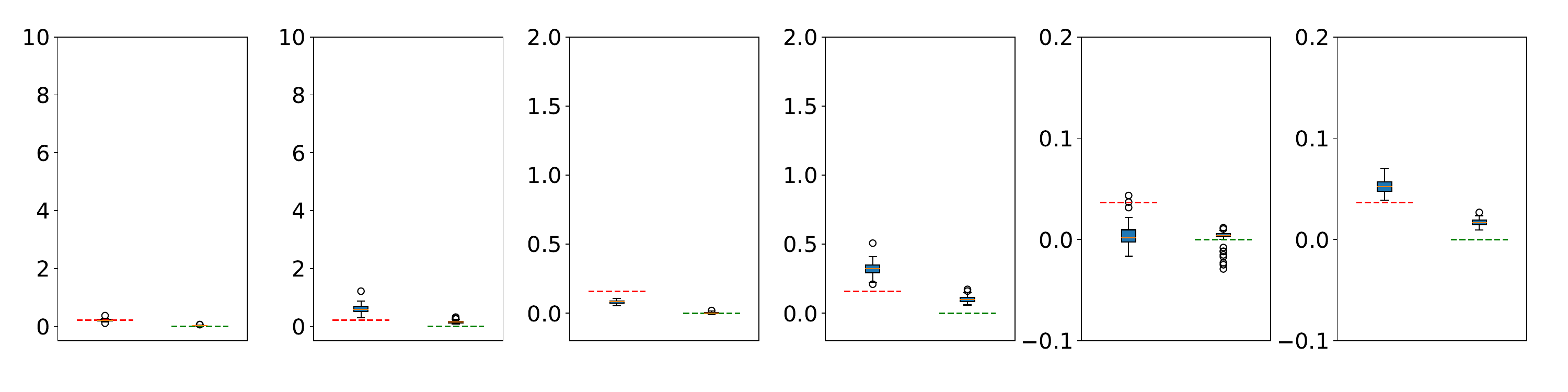}

\begin{figure}
\input{draw_figure2.tex}
\caption{\textbf{Experiment Synth.IIb}}\label{fig:results_2B_alphas}
\end{figure}

\section{Conclusions}{\label{sec:conclusions}}

In this paper, we presented GRULSIF: a novel collaborative likelihood-ratio estimation framework for multiple data sources whose similarity can be encoded as a graph.  Moreover, we provided a detailed convergence analysis that highlights the situations where collaboration is advisable and the role played by important variables as the complexity of the problem at hand, the available data for each local problem, the expressiveness of the graph structure chosen for estimation, and the number of nodes. A distributed implementation is provided as well as a computational complexity analysis that shows how the method conveniently scales for big graphs.

As future work, we ambition to develop the applications outlined in this article, a collaborative extension of transfer learning, multiple hypothesis testing and collaborative change-point detection. The theoretical guarantees in this paper could be latter be used to study the behavior of these tasks when exploiting GRULSIF.  

A topic that was left outside the discussion but is an important component of collaborative \LRE is the graph $G$. Such a topic has motivated some works in Multitasking Learning \citep{Zhang2021}. It is left as an open question to exploit the specifics of \LRE to design a graph learning algorithm with optimal generalization guarantees.

\section*{Acknowledgments}\label{sec:acknowledgments}
 
The authors acknowledge support from the Industrial Data Analytics and Machine Learning Chair hosted at ENS Paris-Saclay, Université Paris-Saclay.

\appendix

\section*{APPENDIX}

\section{Methodological aspects}\label{appendix:methodological}

\addcontentsline{toc}{section}{Appendices}

\subsection{Connection between Pearson's divergence and likelihood-ratio estimation
}\label{appendix:PE_LRE_link}

In this section, we explain the relationship between \PEdiv and likelihood-ratio estimation (\LRE). This connection justifies why we formulate the problem of comparing  probabilistic models defined over the nodes of a graph as a \LRE problem. In other words, we motivate and justify the way we define \Problem{eq:multitasking}.

Let us 
introduce the 
conjugate dual function associated with $\fdivsymb:\real \rightarrow \real$ as: 
\begin{equation*}
    \fdivsymb^*(s)=\sup_{\chi \in \R} s \chi -\fdivsymb(\chi).
\end{equation*}
In general, the connection between \fdiv and \LRE is a consequence of the \Lemma{lemma:Nguyen2008}, which comes from \cite{Nguyen2007}. %
In the case of our interest, \PEdiv is a \fdiv where $\fdivsymb(\chi)=\frac{(\chi-1)^2}{2}$, for which it is easy to verify that the conjugate dual function is:
\begin{equation*}
   \fdivsymb^*(s)= \frac{s^2}{2}+s.
\end{equation*}
Given $\fdivsymb$ is differentiable, we have for the subdifferential: $\partial  \fdivsymb\big(\!\frac{\q(x)}{p(x)}\!\big)=\{\frac{\q(x)}{p(x)}+1\}=\{r(x)-1\}$, where $r(x)=\frac{\q(x)}{p(x)}$. %
After this observation, we can conclude from \Lemma{lemma:Nguyen2008} that:
\begin{equation*}
\begin{aligned}
    \PE(p \Vert \q) & =\int \frac{1}{2}(r(x)-1)^2 p(x) dx = \sup_{g \in \mathcal{F}} \int g(x) \q(x)dx - \int \bigg( \frac{1}{2}g^2(x) + g(x)\bigg) p(x) dx \\
   & = \int  (r(x)-1) \q(x) dx - \int \left(\frac{1}{2}(r(x)-1)^2 + (r(x)-1) \right) p(x) dx 
   \\ 
    & = \int  r(x) \q(x) dx - \int \frac{1}{2}r^2(x)  p(x) dx - \frac{1}{2}.
\end{aligned}
\end{equation*}
By fixing $p(x)=p_v^{\alpha}(x)$ and $\q(x)=\q_v'(x)$, we recover the relative \PEdiv at node-level, denoted by $\PE(p_v^{\alpha} \Vert \q_v)$. 

\subsection{Building an efficient dictionary}\label{appendix:dictionary}

In this section, we describe in further detail how the dictionary is built for the GRULSIF implementation. Our strategy is an adaptation of the algorithm described in \cite{Cedric2009}. The main feature of this method is the definition of the \emph{dictionary coherence}, a metric computable in linear time that is able to measure the redundancy in a set of basis functions, with regard to their lineal dependency. In practice, this approach leads to the selection of a subset of observations (datapoints) forming a non-redundant subset of basis functions with good approximation performance.

This strategy is applicable to unit-norm kernel  ($\KernelFunc(x,x)=1$, $x \in \mathcal{X}$). The coherence of a dictionary $D_L=\{x_{l}\}_{l=1}^L$ of size $L$ is defined as: 
\begin{equation}
    \mu=\max_{l \neq l'} \abs{\dott{\varphi(x_{l})}{\varphi(x_{l'})}}=\max_{l \neq l'} \abs{\KernelFunc(x_{l},x_{l'})}.
\end{equation}
This quantity can be read as the largest level of cross-correlation in the dictionary. When the basis functions are orthogonal, such quantity equals to zero. The authors of that work proposed to integrate a new observation $x$ to the current dictionary $D$ if its coherence remains bellow a given threshold $\mu_0 \in (0,1)$, that is if: 
%
    $\max_{x_{l} \in D_L} \abs{\KernelFunc(x,x_l)} \leq \mu_0$.
%

In our context, we apply this strategy twice: i) first for each of the nodes we produce a dictionary $D_{v}$ at node-level, using a specific threshold coherence $\mu_{0,v}$
, then ii) we filter the elements of the node-level dictionaries with the help of a graph-level coherence threshold $\mu_{0,\G}$. We expect $\mu_{0,G}>\mu_{0,v}$, as we want to preserve as many elements as possible describing each node while eliminating redundancy. The full strategy for the Gaussian kernel is described in \Alg{alg:Dictionar_Building}. Notice that \Alg{alg:Dictionar_Building} may be sensitive to the order of the inputs, as the median heuristic is estimated only with the distance between the elements of $\mathbf{X}_v$.

\begin{algorithm}[!t]
\small
   \caption{\textbf{--} Dictionary building \!\!\!\!}\label{alg:Dictionar_Building}
\begin{algorithmic}[1]
\STATE {\bfseries Input:}  $\{\mathbf{X}_v\}_{v \in V}=\{\{x_{v,1},...,x_{v,\Npre_v}\}\}_{v \in V}$
  , $\{\mathbf{X}'_v\}_{v \in V}=\{\{x'_{v,1},...,x'_{v,\Npost_v}\}\}_{v \in V}$: the observations \\
	\hspace*{\algorithmicindent} \ \ \ \ \ \ \ \ \ \ \ \ indexed by the node of the graph $\G=(V,E,W)$ they belong;\\
  \STATE \hspace*{\algorithmicindent} $\mu_{0,v}$\,: a coherence threshold used to select the elements of the dictionary at the node-level; \\
  \STATE \hspace*{\algorithmicindent} $\mu_{0,\mathcal{G}}$: a coherence threshold used to produce a global dictionary at the graph-level. 
\hspace*{\algorithmicindent}\hspace*{\algorithmicindent}\quad 
\STATE {\bfseries Output:} $D$: a dictionary containing selected elements from $\{\mathbf{X}_v\}_{v \in V}$ and  $\{\mathbf{X}'_v\}_{v \in V}$. \\
\vspace{1.3mm}
\hrule
\vspace{1.3mm}

\raisebox{0.25em}{{\scriptsize$_\blacksquare$}}~\textbf{Create node-level dictionaries}
\FOR{$v \in V$}
\STATE Fix $\sigma_v$ based on the median heuristic with the distance between the elements of $X_v$ 

\STATE $D_{v}=\{x_{v,1}\}$

\raisebox{0.25em}{{\scriptsize$_\blacksquare$}}~\textbf{Select non-redundant elements from $\mathbf{X}_v$}

\FOR{$i \in \{2,..,n_v\}$}
\IF{$\max_{x \in D_{v}} \abs{\KernelFunc_{\sigma_v}}(x_{v,i},x)\leq \mu_{0,v}$}
\STATE $D_{v} = D_{v} \cup \{x_{v,i}\}$ 
\ENDIF
\ENDFOR

\raisebox{0.25em}{{\scriptsize$_\blacksquare$}}~\textbf{Select non-redundant elements from $\mathbf{X}'_v$} 
 \FOR{$i \in \{1,..,n'_v\}$}
\IF{$\max_{x \in D_{v,L_v}} \abs{\KernelFunc_{\sigma_v}(x'_{v,i},x)}\leq \mu_{0,v}$}
\STATE $D_{v} = D_{v} \cup \{x'_{v,i}\}$
\ENDIF
\ENDFOR
\ENDFOR  
\\
\raisebox{0.25em}{{\scriptsize$_\blacksquare$}}~\textbf{Select elements from the node-level dictionaries to generate the global dictionary}%
\STATE{Define the set $Z= \bigcup_{v \in V} D_{v}$} \hfill
\STATE{Fix $\sigma = \text{median}(\{\sigma_v\}_{v \in V})$}
\STATE $D=\{z_{1}\} \in Z$,
\FOR{$z \in Z$}
\IF{$\max_{x \in D} \abs{\KernelFunc_{\sigma}(z,x)}\leq \mu_{0,\mathcal{G}}$}
\STATE $D = D \cup \{z\}$ \hfill
\ENDIF
\ENDFOR 
\STATE \textbf{return} $D$
\end{algorithmic}
\end{algorithm}

In all the experiments reported in this paper, we fix $\mu_{0,v}=0.1$ for all nodes, and $\mu_{0,G}=0.99$. Larger values of $\mu_{0,v}$ did not show any performance improvement, however they increase the running time of our method.

\subsection{Analysis of the proposed optimization algorithm}\label{appendix:optimization}

In this section, we detail the Cyclic Block Gradient Descent (CBGD) strategy described in \Sec{sec:practical_implementation}. In particular, we prove an upper-bound for the number of interactions to attain a given precision $\epsilon$ in terms of the size of the dictionary $L$ and the number of nodes $N$. 

\Theorem{Th:convergence} is a particular case of the results appearing in \cite{Li2018}. In that work, the convergence of Cyclic Block Coordinate-type algorithms is analyzed. For completeness of the presentation, we present some of their main results. %
The objective functions analyzed in \cite{Li2018} takes the form: 
\begin{equation}{\label{eq:LU_opt_problem}}
\begin{aligned}
    \min_{\ThetaG \in \R^M} \Phi(\ThetaG) &=  \min_{\theta \in \R^M}  Q(\ThetaG) + R(\ThetaG),
\end{aligned}
\end{equation}
where $Q$ is a twice differentiable loss function, $R$ is a possibly non-smooth and strongly convex penalty function, and the variable $\ThetaG$ is of dimension $M=\sum_{v=1}^{\Gdims} M_v$ and is partitioned into disjoint blocks $\ThetaG=(\theta_1,\theta_2,...,\theta_\Gdims)$ each of them being of dimension $M_v$. It is supposed that the penalization term can be written as $R(\ThetaG)=\sum_{v=1}^\Gdims R_v(\theta_v)$.

\begin{assumption1}\label{ass_a:Lipschitz}
$Q(\cdot)$ is convex, and its gradient mapping $\nabla Q (\cdot)$ is Lipschitz-continuous and also block-wise Lipschitz-continuous, \ie there exist positive constants $C$ and $C_v$ such that for any $\ThetaG,\ThetaG' \in \R^M$ and $v=1,...,\Gdims$, we have:
\begin{equation}{\label{eq:Li_problem}}
\begin{aligned}
\norm{\nabla Q(\ThetaG') - \nabla Q(\ThetaG)}  & \leq C \norm{\ThetaG'-\ThetaG} \\
\norm{\nabla_v Q(\theta'_{u<v},\theta_v,\theta'_{u>v}) - \nabla_v Q(\ThetaG')} &\leq C_v \norm{\theta_v-\theta'_v}.
\end{aligned}
\end{equation}
\end{assumption1}
%

\begin{assumption1}\label{ass_b:strong_convexity}
$R(\cdot)$ is strongly convex and also blockwise strongly convex, i.e. there exist positive constants $\mu$ and $\mu_v's$ such that for any $\ThetaG,\ThetaG' \in \R^M$ and $v\in V$, we have:
\begin{equation}
\begin{aligned}
R(\ThetaG) & \geq R(\ThetaG') +(\ThetaG-\ThetaG')^{\top}\xi + \frac{\mu}{2}\norm{\ThetaG-\ThetaG'}^2, \\
R_v(\theta_v) & \geq R(\theta'_v) +(\theta_v-\theta'_v)^{\top}\xi_v + \frac{\mu_v}{2}\norm{\theta_v-\theta'_v}^2,
\end{aligned}
\end{equation}
for all $\xi \in \nabla R(\ThetaG')$.
\end{assumption1}
Under the aforementioned assumptions, the CBGD method, in which the cycle $i$ for block $v$, is defined as: 
\begin{equation}\label{eq:CCBGD_update}
\begin{aligned} 
\hthetanode{v}^{(i)} &  \mydef  \argmin_{\theta_v} (\theta_v - \hthetanode{v}^{(i-1)})^{\top} \nabla_v Q (\hthetanode{u<v}^{(i)}, \hthetanode{u \geq v}^{(i-1)})+ \frac{\eta_v}{2} \norm{\theta_v-\hthetanode{v}^{(i-1)}}^2 + R_v(\theta_v).
\end{aligned}
\end{equation}
Then, Theorem\,A.1 characterizes the maximum number of interaction required to achieve a pre-specified accuracy $\epsilon$.

\begin{theorem}\label{Th:Li2018}
(Theorem\,3 in \cite{Li2018}) -- Suppose that Assumptions \ref{ass_a:Lipschitz} and \ref{ass_b:strong_convexity} hold with $M \geq 2$. And that the optimization point is $\ThetaG^*$. We choose $\alpha_v=C_v$ for the CBGD method. Given a pre-specified accuracy $\epsilon>0$ of the objective value, we need at most %
\begin{equation*}
i_{\textup{max}}  \mydef  \bigg\lceil \frac{\mu C_{\textup{min}}^{\mu} + 16 C^2 \log^2(3 \Gdims M_{\textup{max}})}{\mu C_{\textup{min}}^{\mu}} \log\Big( \frac{\phi(\ThetaG^{(0)})-\phi(\ThetaG^*)}{\epsilon}\Big) \bigg\rceil 
\end{equation*}
iterations to ensure $\phi(\ThetaG^{(i)})-\phi(\ThetaG^*) < \epsilon $ for $i \geq i_{\textup{max}}$, where $C_{\textup{min}}^{\mu} \mydef \min_{v \in V} C_v + \mu_v$ and $M_{\textup{max}}=\max_{v \in V } M_v$.
\end{theorem}

\inlinetitle{Proof of \Theorem{Th:convergence}}{.}
\begin{proof}
For this section, we will assume that $\Gramm$ is positive-definite meaning its minimum eigenvalue $e_{\min}(\Gramm)>0$ is strictly positive. In practice, this will be in general not be the case, but we can transform the problem in order to guarantee this condition holds. A solution will be to define an approximation of the problem in terms of $\bar{\Gramm}=\Gramm+c\id_{L}$ with $c>0$. Alternatively we can lake $\Gramm$ positive-definite by selecting a dictionary with linear independent components as described in \cite{Cedric2009} or via Nyström approximations along with anchor points selected via Kernel-PCA as described in Section \ref{sec:Nyström_dictionary}.

\Problem{eq:thetas_cost_function1} takes the form of \Expr{eq:LU_opt_problem}, where we identify the functions $Q$ and $R$ as:
\begin{equation*}
\begin{aligned}
Q(\ThetaG) &=  \min_{\ThetaG \in \real^{\Gdims L }}    \frac{1}{\Gdims}  \left( \frac{(1-\alpha)}{2} \ThetaG ^\top  \mathbf{H} \ThetaG + \frac{\alpha}{2}  \ThetaG^\top \mathbf{H'} \ThetaG  -  \mathbf{h'}^{\top} \ThetaG \right) 
+  \frac{\lambda}{2}  \ThetaG^{\top} ( \id_{N} \otimes \Gramm^{\frac{1}{2}} )^{\top} 
 \left[\Lpc  \otimes \id_{L} \right](\id_{N} \otimes \Gramm^{\frac{1}{2}}) \ThetaG\\
&=  \frac{1}{\Gdims} \sum_{v \in V }  \left( \frac{(1-\alpha)}{2} \theta_v^\top  H_v \theta_v + \frac{\alpha}{2} \theta_v^\top  H'_v \theta_v -  h'^\top_v  \theta_v \right)  + \frac{\lambda}{4} \sum_{u,v \in V} \! W_{uv} (\theta_v-\theta_u)^{\top} \Gramm (\theta_v-\theta_u), \\
R(\theta) &=  \frac{\lambda \gamma}{2} \sum_{v \in V } R_v(\theta_v)  =
\frac{\lambda\gamma}{2} \sum_{v \in V } \theta_v ^\top \Gramm \theta_v.
\end{aligned}
\end{equation*}
That given, it is easy to verify that the updating scheme of \Eq{eq:CCBGD_update} takes the form of \Eq{eq:update}.

It is clear that, given our hypothesis,  $R(\ThetaG)$ and $R_v(\theta_v)$ are stronger convex functions of modulus $\lambda \gamma e_{\min}(K) $. Therefore, \Assumption{ass_b:strong_convexity} is satisfied. 

Second, the full gradient of $Q(\cdot)$ can be written as:
\begin{equation}
    \nabla Q(\ThetaG)=   \left( \frac{1-\alpha}{\Gdims} \mathbf{H}  + \frac{\alpha}{\Gdims} \mathbf{H'} +  \lambda ( \id_{N} \otimes \Gramm^{\frac{1}{2}} )^{\top} 
 \left[\Lpc  \otimes \id_{L} \right](\id_{N} \otimes \Gramm^{\frac{1}{2}}) \right) \ThetaG - \frac{1}{\Gdims} \mathbf{h'},
\end{equation}
which is Lipschitz-continuous with constant
%
$C=\eigmax{ \frac{1-\alpha}{\Gdims} H  + \frac{\alpha}{\Gdims} H' +  \lambda ( \id_{N} \otimes \Gramm^{\frac{1}{2}} )^{\top} 
 \left[\Lpc  \otimes \id_{L} \right](\id_{N} \otimes \Gramm^{\frac{1}{2}})}$.
%
From the node-level expression, it is easy to derive the partial derivative of $Q(\cdot)$:
\begin{equation}
\begin{aligned}
    \nabla_v Q(\ThetaG) & = \frac{1-\alpha}{\Gdims} H_{v} + \frac{\alpha}{\Gdims} H'_{v} \theta_v
    + \lambda \Gramm \Bigg( d_v  \theta_v - \sum_{u \in \nghood(v)\!\!\!\!\!\!\!\!} W_{uv} \big( \theta_u \one_{u<v} + \theta_u \one_{u > v} \big) \Bigg) - \frac{1}{\Gdims}h'_{v},\!\!\!\!\!\!
\end{aligned}
\end{equation}
where $d_v$ is the degree of node $v$. This means: 
\begin{equation}
\begin{aligned}
   \norm{\nabla_v Q(\theta'_{u<v},\theta_v,\theta'_{u>v}) - \nabla_v Q(\ThetaG')} & \leq \norm{ \left( \frac{1-\alpha}{\Gdims} H_{v,t} + \frac{\alpha}{\Gdims} H'_{v} + \lambda d_v \Gramm \right) \left(\theta_v-\theta'_v \right)} \\
    & \leq C_v \norm{(\theta_v-\theta'_v)},
\end{aligned}
\end{equation}
where $C_v=  \eigmax{ \frac{1-\alpha}{\Gdims} H_{v,t} + \frac{\alpha}{\Gdims} H'_{v} + \lambda d_v K} $. Then, \Assumption{ass_a:Lipschitz} is satisfied.  

With these elements, and by fixing $\eta_v \mydef C_v$, we can apply Theorem\,A.1 where
\begin{equation*}
\begin{aligned}
C & \mydef \eigmax{ \frac{1-\alpha}{\Gdims} \mathbf{H}  + \frac{\alpha}{\Gdims} \mathbf{H'} +  \lambda( \id_{N} \otimes \Gramm^{\frac{1}{2}} )^{\top} 
 \left[\Lpc  \otimes \id_{L} \right](\id_{N} \otimes \Gramm^{\frac{1}{2}}) }, \\
C_{v} & \mydef \eigmax{ \frac{(1-\alpha)}{\Gdims} H_{v} + \frac{\alpha}{\Gdims} H'_{v} + \lambda d_v \Gramm}, \\ 
C^{\mu}_{\textup{min}} & \mydef \min_{v \in V} C_v + \mu, \\
\mu & \mydef  \lambda \gamma c.
\end{aligned}
\end{equation*}
After substitution, we get the expression given in \Eq{eq:complexity}.
\end{proof}
\section{Practical use of GRULSIF and details for the empirical evaluation}\label{appendix:practical_use}

\subsection{Further details for the conducted experiments}{\label{appendix:hyperparemeters}}

In this section, we detail how the hyperparameters used in the experiments were chosen. 

For  RULSIF and ULSIF algorithms, we follow \citep{Sugiyama2011} and \citep{Yamada2011}.
We run a leave-one-out cross-validation procedure over the parameter associated with the Gaussian kernel and the penalization term $\gamma$. The parameter $\sigma$ is selected from the grid $\{0.6 \sigma_{\text{median}},0.8\sigma_{\text{median}},1\sigma_{\text{median}},1.2\sigma_{\text{median}},1.4\sigma_{\text{median}} \}$  where $\sigma_{\text{median}}$ is the parameter $\sigma$ found via the median heuristic over the observations in $X'_v$. On the other hand, the penalization parameter $\gamma$ is optimized from the grid $\{1e^{-5},1e^{-3},0.1,10\}$. 

The procedure for KLIEP is similar, but we use instead a $5$-fold cross-validation procedure, over the grids  $ \{ 0.6 \sigma_{\text{median}},0.8\sigma_{\text{median}},1\sigma_{\text{median}},1.2\sigma_{\text{median}},1.4\sigma_{\text{median}} \}$  for the width $\sigma$ of the Gaussian kernel, and $\{1e^{-5},1e^{-3},0.1,10\}$ for the penalization constant. 

Finally, for the GRULSIF and the POOL algorithms, we apply  $5$-fold cross-validation to select the hyperparameters $\sigma$, $\gamma$, and $\lambda$ (\Alg{alg:model_selection}). Since the POOL approach ignores the graph structure, we fix $\lambda=1$, and the penalization term related with the norm of each functional $f_v$ will depend just on the parameter $\gamma$. In order to select the width $\sigma$ for the Gaussian kernel, we first compute $\{\sigma_{v}\}_{v \in V}$ for each node via the median heuristic applied to the observations of $X_v$ (such quantities are available when generating the dictionary), and we define $\sigma_{\text{min}}= \argmin \{\sigma_{v}\}_{v \in V}$, $\sigma_{\text{median}}= \text{median}\{\sigma_{v}\}_{v \in V}$ and $\sigma_{\text{max}}= \argmax \{\sigma_{v}\}_{v \in V}$, we then chose the final parameter from the set $\{\sigma_{\text{min}},\frac{1}{2}(\sigma_{\text{min}}+\sigma_{\text{median}}),\sigma_{\text{median}},\frac{1}{2}(\sigma_{\text{max}}+\sigma_{\text{median}}),\sigma_{\text{max}}\}$. $\gamma$ is selected from the set $\{1e^{-5},1e^{-3},0.1,1\}$. Finally, we define the average node degree $\bar{d}$, 
and we identify the optimal $\lambda^*$ from the set $\{1e^{-3}\cdot\frac{1}{\bar{d}},1e^{-2}\cdot\frac{1}{\bar{d}},0.1\cdot\frac{1}{\bar{d}},1\cdot\frac{1}{\bar{d}},10\cdot\frac{1}{\bar{d}}\}$.

\section{GRULSIF convergence guarantees}{\label{app:convergence_guarantees}}

\subsection{Auxiliary concepts and results from Multitask Learning}

The excess risk bounds for Multitask  Learning  proposed by \cite{Yousefi2018} depend on the concept of Multitask Local Rademacher Complexity (MTLRC) which aims to quantify the complexity of classes of vector-valued functions.  It has the advantage of leading to sharper bounds when compared to classical Global Rademacher Complexity. Moreover, they are easy to compute for \VVRKHS. In this last scenario, the bounds are tight enough to 
explicit the role played by important variables such as the number of observations, the number of tasks, the smoothness of the vector-valued function to approximated in terms of the norm in the associated \VVRKHS.

For completeness of exposition and to clarify the content of the results, we rewrite important concepts and results appearing on the reference papers \cite{Barlett2005,Yousefi2018} adapted to the notation used in the main text.   

Let us denote by $\mathbf{Z}:=(z_{v,i})_{v \in V, i=1,...,n}$  a set of $n \Gdims$ independent observations such that for each $v \in \{1,...,\Gdims\}$ $\{z_{v,i}\}_{i=1}^{n}$ are identically distributed according to the measure $p_{z,v}$. 

Given a vector-valued function $\mathbf{h}=(h_1,...,h_v)$ we define the expressions: 
\begin{equation}
\begin{aligned}
P_z[\mathbf{h}]:= \frac{1}{\Gdims} \sum_{v \in V} \ExpecJointv{h_v(z)}  \ \ \ \text{and} \ \ \ P_{z,n}[\mathbf{h}]:= \frac{1}{n \Gdims} \sum_{v=1}^{\Gdims} \sum_{i=1}^{n}  h_v(z_{v,i}). 
\end{aligned}
\end{equation}
We start by introducing the concept of MTLRC \citep{Yousefi2018} and the sub-root function \citep{Barlett2005} that will appear in the 
upper-bounds of the excess risk. 
\begin{definition}(\textbf{Multitask Local Rademacher Complexity}) For a vector-value function class $\mathcal{F}=\{\mathbf{f}=(f_1,...,f_{\Gdims})\}$, the Multitask Local Rademacher Complexity (MTLRC) for $\rho>0$, $\mathcal{R}(\mathcal{F},\rho)$, is defined as: 
\begin{equation}{\label{eq:MT-LRC}}
\mathcal{R}(\mathcal{F},\rho)= \mathbb{E}_{z,\sigma} \Bigg[ \sup_{\substack{V(\mathbf{f}) \leq \rho \\ \mathbf{f}=(f_1,...,f_{\Gdims}) \in \mathcal{F}}}
\frac{1}{n \Gdims} \sum_{v=1}^{\Gdims} \sum_{i=1}^{n} \sigma_{v,i} f_v(z_{v,i}) \Bigg],
\end{equation}
where $\{\sigma_{v,i}\}_{v=1,...,\Gdims;i=1,...,n}$ is a sequence of independent Rademacher variables. We denote by  $\mathbb{E}_{z,\sigma}\left[\cdot\right]$ with respect to all the involved random variables. $V(\mathbf{f})$ is an upper bound on the variance of the function in $\mathcal{F}$. 
\end{definition}
\begin{definition}(\textbf{Sub-root function}) A function $\rhofunction:[0, \infty] \rightarrow [0, \infty] $ is sub-root iff 
it is non-decreasing and the function $\frac{\rhofunction(\rho)}{\sqrt{\rho}}$ is non-increasing for $\rho>0$.
\end{definition}
\begin{lemma}{\label{lemma:Lemma32Barlett}}(Lemma\,3.2 in \cite{Barlett2005}) If $\rhofunction$ is a sub-root function, then it is continuous on $[0,\infty]$, and the equation $\rhofunction(\rho)=\rho$ has a unique non-zero solution $\rho^*$, which is known as the fixed point of $\rhofunction$. Moreover, for any $\rho>0$, if holds that $\rho \geq \rhofunction(\rho)$ iff 
$\rho^* \leq \rho$. 
\end{lemma}
The following result establishes situations when the MTLRC is itself a sub-root function.
\begin{lemma}{\label{lemma:rademacher_and_root_functions}}(Lemma\,3.4 in \cite{Barlett2005}) If the class $\mathcal{F}$ is star-shaped around $\mathbf{f}_0$, and  $V:\mathcal{F} \rightarrow \R^{+}$ is a function that satisfies $V(a \mathbf{f}) \leq a^2 V(\mathbf{f})$ for any  $\mathbf{f} \in \mathcal{F}$ and any $a \in [0,1]$, then the function $\rhofunction$ defined for $\rho \geq 0$ by: 
\begin{equation}
\rhofunction(\rho)=\mathbb{E}_{\sigma} \Bigg[ \sup_{\substack{V(\mathbf{f}-\mathbf{f}_0) \leq \rho \\ \mathbf{f}\in \mathcal{F}}}
\frac{1}{n \Gdims} \sum_{v=1}^{\Gdims} \sum_{i=1}^{n} \sigma_{v,i} f_v(z_{v,i}) \Bigg]
\end{equation}
is sub-root and $r \rightarrow \mathbb{E}_{z}[\rhofunction(r)]$ is also sub-root. 
\end{lemma}
Being star-shaped around $\mathbf{f}_0$, means:
\begin{equation}
    \{\mathbf{f}_0+a(\mathbf{f}-\mathbf{f}_0): \mathbf{f} \in \mathcal{F}, a \in [0,1] \} \subset \mathcal{F}.
\end{equation}
Notice that when the class $\mathcal{F}$ is convex, it is star-shaped around each of its elements by definition. 

MTLRC will be used to obtain a global error bound to classes of vector-valued functions for which the variance is bounded $V(\mathbf{f}-\mathbf{f}_0) \leq \rho$. The goal is to identify models that attain a small generalization error and enjoy a small variance. There is a tradeoff between the size of the subset we consider (controlled by the parameter $\rho$)  and its complexity, the optimal choice is given by a fixed point of a root function. In order, to formalize the relationship between generalization error and variance 
we need to define the concept of Vector-Valued Berstein Class.
\begin{definition}(\textbf{Vector-Valued Bernstein Class}) Let $0<\beta \leq 1$ and $B>0$. A vector-valued function class $\mathcal{F}$ is said to be a $(\beta,B)$-Berstein class with respect to the probability measure $P$ if there exists a function $V:\mathcal{F} \rightarrow \R^{+}$ such that 
\begin{equation}
P\mathbf{f}^2 \leq V(\mathbf{f}) \leq B(P\mathbf{f})^{\beta}, \ \ \forall \mathbf{f} \in \mathcal{F}.
\end{equation}
\end{definition}

The following result describes the role of MTLRC to obtain upper bounds for Multitask Learning, it is the core component in the proof of \Theorem{thm:convergence_results}. 
\begin{theorem}{\label{thm:TheoremB3_Yousefi}}(Theorem\,B.3 in \cite{Yousefi2018}) Let $\mathcal{F}=\{\mathbf{f}=(f_1,...,f_{\Gdims})\}$ be a class of vector-valued functions satisfying $\max_{v \in V} \sup_{z \in \mathcal{Z}} \abs{f_v(z)} \leq b$. Let $\mathbf{Z} = \{\mathbf{Z}_v\}_{v\in V} = \big\{\{z_{v,1},...,z_{v,n}\}\big\}_{v \in V}$ be a vector of $n \Gdims$ random variables where  for each $v \in V$,$\{z_{v,1},...,z_{v,n}\}$ are identically distributed. Assume that $F$ is $(\beta,B)$-Bernstein class of vector-valued functions with $0< \beta \leq 1$ and $B \geq 1$. Let $\rhofunction$ be a sub-root function with fixed point $\rho^*$. If $B\mathcal{R}(\mathcal{F},r) \leq \rhofunction(\rho)$, $\forall \rho \geq \rho^*$, then for any $C > 1$, and $\delta \in (0,1)$, with probability at least $1-\delta$, every $\mathbf{f} \in \mathcal{F}$ satisfies: 

\begin{equation}
\begin{aligned}
    P_{z}[\mathbf{f}] \ \leq\  & \frac{C}{C-\beta} P_{z,n}[\mathbf{f}]+(2C)^{\frac{\beta}{2-\beta}}20^{\frac{2}{2-\beta}}\max\left( (\rho^*)^{\frac{1}{2-\beta}} ,(\rho^{*})^{\frac{1}{\beta}} \right) \\
    & + \left( \frac{2^{\beta+3} B^2 C^{\beta}}{n \Gdims} \log{\left(\frac{1}{\delta}\right)}\right)^{\frac{1}{2-\beta}}
    + \frac{24 B b}{(2-\beta) n \Gdims} \log{\left(\frac{1}{\delta}\right)}.
\end{aligned}
\end{equation}   
\end{theorem}

\subsection{Lemmata before \Theorem{thm:convergence_results}}

The general idea is to use \Theorem{thm:TheoremB3_Yousefi} to upperbound the excess risk associated with the optimization problem 
\ref{eq:functional_space_graph_smoothness}. To this end, we need to address the following subproblems: 
\begin{enumerate}[topsep=0.4em, itemsep=0.em]
    \item Define a class of vector-valued functions satisfying the hypotheses of \Theorem{thm:TheoremB3_Yousefi} (This is the context of \Lemma{lemma:Bbclass}).
    \item Identify the sub-root function $\rhofunction$ that upperbound the MTLRC of the class. (Provided in the last point of \Lemma{lemma:Bbclass})
    \item Upperbound the fixed point of $\rhofunction$.
    (\Lemma{lemma:rhofunction})
\end{enumerate}
Let us start by defining the instantaneous loss function for the scalar function $f \in \Hilbert$: 
\begin{equation}
\ell_v(f)(z_v)= \frac{(1-\alpha)f^2(x_v)+ \alpha f^2(x'_v)}{2} - f(x'_v). 
\end{equation}
Here, the variable $z_{v}$ denotes a pair of observations $z_{v}=(x_{v},x'_{v})$, where $x_{v} \sim p_v$ and $x'_v \sim \q_v$.  

Given a vector-valued function $\mathbf{f}=(f_1,f_2,...,f_{\Gdims}) \in \Ghilbert$  and pairs of observations $(z_1,z_2,...,z_{\Gdims})$, we define the vector-valued loss function:
\begin{equation}
\ell(\mathbf{f})=\left(\ell_1(f_1)(z_1),\ell_2(f_2)(z_2),...,\ell_{\Gdims}(f_{\Gdims})(z_{\Gdims})\right).
\end{equation}

To facilitate reading, we will introduce the following operators evaluated at vector-valued functions of the form $\mathbf{h}=(h_1,...,h_{\Gdims})$:
\begin{equation}
\begin{aligned}
P[\mathbf{h}]= \frac{1}{\Gdims} \sum_{v \in V} \Expecnv\left[ h_v(x)  \right], \ \  
Q[\mathbf{h}]= \frac{1}{\Gdims} \sum_{v \in V} \Expecav \left[ h_v(x') \right], \ \
P^{\alpha}[\mathbf{h}]= \frac{1}{\Gdims} \sum_{v \in V} \ExpecAlphav{h_v(y)}.
\end{aligned}
\end{equation}
We can easily verify the following expressions: 
\begin{equation}{\label{eq:expectantions_relationships}}
 P^{\alpha}[\mathbf{h}]= (1-\alpha) P[\mathbf{h}] + \alpha Q[\mathbf{h}], \text{ and } \ \ P^{\alpha}[\mathbf{r}^{\alpha}\mathbf{h}]= Q[\mathbf{h}],
\end{equation}
where, with an abuse of notation $\mathbf{r}^{\alpha}\mathbf{h}$, refers to point-wise multiplication of the vector-valued functions $\mathbf{r}^{\alpha}$ and $\mathbf{h}$. This convention will be used throughout the text. %
With this notation, we can define the cost function: 
\begin{equation}
L(\mathbf{f})=\sum_{v \in V} \Expecjointv\left[\ell_v(f_v)(z)\right]= P_z[\ell(\mathbf{f})].
\end{equation}
The following lemma identifies the connection between the excess risk $L(\mathbf{f})- L(\mathbf{r}^{\alpha})$ and the $L_2$ distance $P^{\alpha}[\mathbf{f}-\mathbf{r}^{\alpha}]^2$. In particular, this lemma makes evident the advantages of using \PEdiv as a surrogate loss function for \LRE .

\begin{lemma}{\label{lemma:why_pearson}}
Consider the vector-valued functional space $\mathcal{F}_G$ (\Expr{eq:functional_space_graph_smoothness}) and suppose the value of the scalar functions $f_v$ ranges in $[-b,b]$. Then the following statements hold.
\begin{enumerate}[topsep=0.4em, itemsep=0.em]
\item  There is a function $\mathbf{f}^*=(f^*_1,...,f^*_{\Gdims}) \in \mathbb{G}$ satisfying:%
    \begin{equation*}
    \mathbf{f}^*:=\argmin_{\mathbf{f} \in \mathcal{F}_G} \frac{1}{\Gdims} \sum_{v\in V} \left[ \frac{1}{2} \ExpecAlphav{f_v^2(x)} - \ExpecAm{f_v(x)}{v} \right] = \argmin_{\mathbf{f} \in \mathcal{F}_G}  L(f). 
    \end{equation*}
\item  For every $\mathbf{f} \in \mathbb{G}$, we have
 $P^{\alpha}\left[\mathbf{f}-\mathbf{f}^*\right]^2 = 2 (L(\mathbf{f})- L(\mathbf{f}^*))$.
 \item There exists $B_0>0$, such that $\forall \mathbf{f} \in \mathbf{G}$:
 \begin{equation*}
 P_{z}[\ell(\mathbf{f})-\ell(\mathbf{f}^*)]^2 \leq B_0 P^{\alpha}[\mathbf{f}-\mathbf{f}^*]^2 =  B_0 P_{z}[\ell(\mathbf{f})-\ell(\mathbf{f}^*)].    
 \end{equation*}
\end{enumerate}
\end{lemma}

\begin{proof}\mbox{}\\
\inlinetitle{First point}{.} Assumption \ref{ass:model_definition} says $\mathbf{r}^{\alpha} \in \mathcal{F}_G$. 
Following the line of reasoning  used to prove \Expr{eq:least_squares_node_formulation}, we can conclude: 
\begin{equation*}
\argmin_{\mathbf{f} \in \mathcal{F}_G} \frac{1}{\Gdims} \sum_{v\in V} \left[ \frac{1}{2} \ExpecAlphav{f_v^2(y)} - \ExpecAm{f_v(x')}{v} \right]=\argmin_{\mathbf{f} \in \mathcal{F}_G} \frac{1}{\Gdims} \sum_{v\in V} \frac{1}{2}\ExpecAlphav{(f_v(y)-r_v^{\alpha}(y))^2},
\end{equation*}
which implies $\mathbf{r}^{\alpha}=(r_1^{\alpha},...,r_{\Gdims}^{\alpha})$ is solution to the optimization problem.  

\inlinetitle{Second point}{.} The proof of the previous point implies $\mathbf{f}^*=\mathbf{r}^{\alpha}$. Then the second point of the lemma can restated in terms of $L(\mathbf{f})-L(\mathbf{r}^{\alpha})$ for $\mathbf{f} \in \mathcal{F}_G$: 
\begin{equation*}
\begin{aligned}
L(\mathbf{f})-L(\mathbf{r}^{\alpha}) &= \frac{1}{\Gdims} \sum_{v\in V} \left[ \frac{1}{2}\ExpecAlphav{f_v^2(y) - (r^{\alpha}_v)^2(y) } - \ExpecAm{\left(f_v(x')- r^{\alpha}_v(x')\right)}{v} \right] \\
&= \frac{1}{\Gdims} \sum_{v\in V} \Expecalphav \left[
\frac{1}{2} \left[f_v^2(y) - (r^{\alpha}_v)^2(y) \right]- r^{\alpha}_v(y)(f_v(y) - r^{\alpha}_v(y)) \right] \\
& = \frac{1}{\Gdims}  \sum_{v\in V} \frac{1}{2} \ExpecAlphav{[f_v -r^{\alpha}_v]^2(y)} \\
& = \frac{1}{2} P^{\alpha}[\mathbf{f}-\mathbf{r}^{\alpha}]^2,
\end{aligned}
\end{equation*}
where the second inequality comes from the expression $\ExpecAlphav{r^\alpha_v(y)g(y)}= \ExpecAm{g(x')}{v}$. 

\inlinetitle{Third point}{.} Notice by hypothesis over the functional space $\mathcal{F}_G$ and the upperbound of $r^{\alpha}_v$ with respect to the regularization parameter: 

\begin{equation}{\label{eq:inf_norm_sum_f_r}}
\norm{f_v+r^{\alpha}_v}_{\infty} \leq (b+\frac{1}{\alpha}) \ \ \norm{f_v-r^{\alpha}_v}_{\infty} \leq (b+\frac{1}{\alpha})
\end{equation}

\begin{equation*}
\begin{aligned}
P_{z}\left[\ell(\mathbf{f})-\ell(\mathbf{r}^{\alpha}) \right]^2
&= P_z \left[ \frac{1-\alpha}{2} [\mathbf{f}^2-(\mathbf{r}^{\alpha})^2](x)  + \frac{\alpha}{2} [\mathbf{f}^2-(\mathbf{r}^{\alpha})^2](x')  - [\mathbf{f}-\mathbf{r}^{\alpha}](x') \right]^2 \\ 
& \leq 2  P_{z}\left[ \frac{1-\alpha}{2} [\mathbf{f}^2-(\mathbf{r}^{\alpha})^2](x)  + \frac{\alpha}{2}[\mathbf{f}^2-(\mathbf{r}^{\alpha})^2](x')\right]^2 + 2 Q \left[ [\mathbf{f}-\mathbf{r}^{\alpha}]^2(x') \right]  (\text{\tiny{Inequality $(a+b)^2 \leq 2a^2+2b^2$}}) \\
&\leq 2  P_{z} \left[
\frac{(1-\alpha)}{4} \left(\left[\mathbf{f}^2-(\mathbf{r}^{\alpha})^2\right](x)\right)^2  + \frac{\alpha}{4} \left(\left[\mathbf{f}^2-(\mathbf{r}^{\alpha})^2\right](x')\right)^2  \right] + 2 Q \left[ \left[\mathbf{f}-\mathbf{r}^{\alpha}\right]^2(x') \right](\text{\tiny{Convexity of $x \rightarrow x^2$}}) \\
& = \frac{1}{2} P^{\alpha}  \left[\mathbf{f}^2-(\mathbf{r}^{\alpha})^2\right]^2 + 2 P^{\alpha}\left[\mathbf{r}^{\alpha}(\mathbf{f}-\mathbf{r}^{\alpha})^2\right] \qquad (\text{\tiny{\Expr{eq:expectantions_relationships}}}) 
\\
&\leq \frac{1}{2} P^{\alpha} \left[ \left[\mathbf{f}-\mathbf{r}^{\alpha}\right]^2 \left[\mathbf{f}+\mathbf{r}^{\alpha}\right]^2 \right] + \frac{2}{\alpha} P^{\alpha}\left[\left[\mathbf{f}-\mathbf{r}^{\alpha}\right]^2\right] \\
& \leq  \frac{1}{2}\left(\Big(b+\frac{1}{\alpha}\Big)^2 +\frac{4}{\alpha} \right) P^{\alpha}\left[\mathbf{f}-\mathbf{r}^{\alpha}\right]^2 \qquad (\text{\tiny{\Expr{eq:inf_norm_sum_f_r}}})
\\ &= \frac{1}{2} B_0 P^{\alpha}\left[\mathbf{f}-\mathbf{r}^{\alpha}\right]^2,
\end{aligned}
\end{equation*}
Moreover, the second point implies: 
\begin{equation*}
    P_{z}\left[\ell(\mathbf{f})-\ell(\mathbf{r}^{\alpha}) \right]^2 \leq B_0 \left[ L(\mathbf{f}) -L(\mathbf{r}^{\alpha}) \right] = P_z \left[\ell(\mathbf{f}) -\ell(\mathbf{r}^{\alpha}) \right].
\end{equation*}
\end{proof}
\begin{lemma}{\label{lemma:MTLCR_alpha_bound}} Let  $\mathbf{Y} = \{\mathbf{Y}_v\}_{v\in V} = \big\{\{y_{v,1},...,y_{v,n}\}\big\}_{v \in V}$ be a sample of $n \Gdims$ observations such that $\forall v,i: \ \ y_{v,i} \,\overset{\text{\iid}}{\sim}\, p_v^{\alpha}$. Then: 
\begin{equation}
\mathbb{E}_{\sigma} \Bigg[ \sup_{\substack{
P^{\alpha}\left[\mathbf{f}-\mathbf{r}^{\alpha}\right]^2 \leq \rho \\ \mathbf{f} \in \mathcal{F}_G}} \frac{1}{n \Gdims} \sum_{v \in V} \sum_{i=1}^{n} \sigma_{v,i} f_v(y_{v,i}) r^{\alpha}(y_{v,i}) \Bigg] \leq \frac{1}{\alpha}
\mathbb{E}_{\sigma} \Bigg[ \sup_{\substack{
P^{\alpha}\left[\mathbf{f}-\mathbf{r}^{\alpha}\right]^2 \leq \rho \\ \mathbf{f} \in \mathcal{F}_G}}  \frac{1}{n \Gdims} \sum_{v \in V} \sum_{i=1}^{n} \sigma_{v,i} f_v(y_{v,i})\Bigg] 
\end{equation}
\end{lemma}
\begin{proof}
Let us define $\mathbb{E}_{\sigma \setminus \sigma_{u,j}}[\cdot]$ the expectation with respect to all the Rademacher random variables $\{\sigma_{v,i}\}_{v=1,...,\Gdims;i=1,...,n}$ except $\sigma_{u,j}$, then: 
\begin{equation*}
\begin{aligned}
&\mathbb{E}_{\sigma} \Bigg[ \sup_{\substack{
P^{\alpha}\left[\mathbf{f}-\mathbf{r}^{\alpha}\right]^2 \leq \rho \\ \mathbf{f} \in \mathcal{F}_G}} \frac{1}{n \Gdims} \sum_{v \in V} \sum_{i=1}^{n} \sigma_{v,i} f_v(y_{v,i}) r^{\alpha}_v(y_{v,i}) \Bigg]  =
\mathbb{E}_{\sigma \setminus \sigma_{u,j}}  \left[ \mathbb{E}_{\sigma_{u,j}} \Bigg[ \sup_{\substack{
P^{\alpha}\left[\mathbf{f}-\mathbf{r}^{\alpha}\right]^2 \leq \rho \\ \mathbf{f} \in \mathcal{F}_G}}  \frac{1}{n \Gdims} \sum_{v \in V} \sum_{i=1}^{n} \sigma_{v,i} f_v(y_{v,i}) r^{\alpha}_v(y_{v,i}) \Bigg] \right]  \\
& = \frac{1}{n \Gdims} \mathbb{E}_{\sigma \setminus \sigma_{u,j}}  \left[ \mathbb{E}_{\sigma_{u,j}} \Bigg[ \sup_{\substack{
P^{\alpha}\left[\mathbf{f}-\mathbf{r}^{\alpha}\right]^2 \leq \rho \\ \mathbf{f} \in \mathcal{F}_G}} U_{V \setminus u;-j}(\mathbf{f}) + \sigma_{u,j} f_{u}(y_{u,j}) r^{\alpha}_u(y_{u,j})
\Bigg] \right],
\end{aligned}
\end{equation*}
where $U_{V \setminus u;-j}(\mathbf{f})=\sum_{v \in V } \sum_{i=1}^{n} \sigma_{v,i} f_v(y_{v,i}) r_v^{\alpha}(y_{v,i}) -  \sigma_{u,j} f_u(y_{u,j}) r_u^{\alpha}(y_{u,j})$. By definition of the supremum, for any $\epsilon>0$, there exists $\mathbf{g},\mathbf{h} \in \mathcal{F}_{G}$ such that 
$P^{\alpha}(\mathbf{g}-\mathbf{r}^{\alpha})^2 \leq \rho$ and $P^{\alpha}(\mathbf{h}-\mathbf{r}^{\alpha})^2 \leq \rho$, such that: 
\begin{equation*}
\begin{aligned}
U_{V \setminus u;-j}(\mathbf{g})+ g_{u}(y_{u,j}) r^{\alpha}_{u}(y_{u,j})  &\geq (1-\epsilon) \Bigg[ \sup_{\substack{
P^{\alpha}\left[\mathbf{f}-\mathbf{r}^{\alpha}\right]^2 \leq \rho \\ \mathbf{f} \in \mathcal{F}_G}} U_{V \setminus u;-j}(\mathbf{f}) +  f_{u}(y_{u,j}) r^{\alpha}_u(y_{u,j})
\Bigg] \\ 
U_{V \setminus u;-j}(\mathbf{h})- h_{u}(y_{u,j}) r^{\alpha}_{u}(y_{u,j})   & \geq (1-\epsilon) \Bigg[ \sup_{\substack{
P^{\alpha}\left[\mathbf{f}-\mathbf{r}^{\alpha}\right]^2 \leq \rho \\ \mathbf{f} \in \mathcal{F}_G}} U_{V \setminus u;-j}(\mathbf{f}) -  f_{u}(y_{u,j}) r^{\alpha}_u(y_{u,j})
\Bigg].
\end{aligned}
\end{equation*}
This latter implies:
\begin{equation*}
\begin{aligned}
&(1-\epsilon)\mathbb{E}_{\sigma_{u,j}} \Bigg[ \sup_{\substack{
P^{\alpha}\left[\mathbf{f}-\mathbf{r}^{\alpha}\right]^2 \leq \rho \\ \mathbf{f} \in \mathcal{F}_G}} U_{V \setminus u;-j}(\mathbf{f}) + \sigma_{u,j} f_{u}(y_{u,j}) r^{\alpha}_u(y_{u,j})
\Bigg] \\
&=(1-\epsilon) \Bigg[ \frac{1}{2} \sup_{\substack{
P^{\alpha}\left[\mathbf{f}-\mathbf{r}^{\alpha}\right]^2 \leq \rho \\ \mathbf{f} \in \mathcal{F}_G}}  \left[  U_{V \setminus u;-j}(\mathbf{f}) + f_{u}(y_{u,j}) r^{\alpha}_u(y_{u,j}) \right] + \frac{1}{2} \sup_{\substack{
P^{\alpha}\left[\mathbf{f}-\mathbf{r}^{\alpha}\right]^2 \leq \rho \\ \mathbf{f} \in \mathcal{F}_G}}  \left[  U_{V \setminus u;-j}(\mathbf{f}) - f_{u}(y_{u,j}) r^{\alpha}_u(y_{u,j}) \right]     \Bigg] \\
&\leq \frac{1}{2} \left[ U_{V \setminus u;-j}(\mathbf{g})+ g_{u}(y_{u,j}) r^{\alpha}_{u}(y_{u,v})  \right] +
\frac{1}{2} \left[ U_{V \setminus u;-j}(\mathbf{h})- h_{u}(y_{u,j}) r^{\alpha}_{u}(y_{u,v})   \right] \\
& \leq \frac{1}{2} \left[
U_{V \setminus u;-j}(\mathbf{g})+U_{V \setminus u;-j}(\mathbf{h})+ s r^{\alpha}_{u}(y_{u,j})\left(g_{u}(y_{u,j})-h_{u}(y_{u,j})\right)\right],
\end{aligned}
\end{equation*}
where $s=\operatorname{sgn}(g_u(y_{u,j})-h_u(y_{u,j}))$. Then, the upperbound on $r^{\alpha}$ implies:
\begin{equation*}
\begin{aligned}
&(1-\epsilon)\mathbb{E}_{\sigma_{u,j}} \Bigg[ \sup_{\substack{
P^{\alpha}\left[\mathbf{f}-\mathbf{r}^{\alpha}\right]^2 \leq \rho \\ \mathbf{f} \in \mathcal{F}_G}} U_{V \setminus u;-j}(\mathbf{f}) + \sigma_{u,j} f_{u,j}(y_{u,j}) r^{\alpha}_u(y_{u,j})
\Bigg] \\ 
&\leq \frac{1}{2} \left[
U_{V \setminus u;-j}(\mathbf{g})+U_{V \setminus u;-j}(\mathbf{h})+ \frac{1}{\alpha} s\left(g_{u}(y_{u,j})-h_{u}(y_{u,j})\right)\right] \\
& = \frac{1}{2} \left[ U_{V \setminus u;-j}(\mathbf{g}) + \frac{1}{\alpha} sg_{u}(y_{u,j}) \right] + \frac{1}{2} \left[ U_{V \setminus u;-j}(\mathbf{h}) - \frac{1}{\alpha} sh_{u}(y_{u,j}) \right] \\
& \leq \frac{1}{2} \sup_{\substack{
P^{\alpha}\left[\mathbf{f}-\mathbf{r}^{\alpha}\right]^2 \leq \rho \\ \mathbf{f} \in \mathcal{F}_G}} \left[ U_{V \setminus u;-j}(\mathbf{f}) + \frac{1}{\alpha} sf_{u}(y_{u,j}) \right] + \frac{1}{2} \sup_{\substack{
P^{\alpha}\left[\mathbf{f}-\mathbf{r}^{\alpha}\right]^2 \leq \rho \\ \mathbf{f} \in \mathcal{F}_G}} \left[ U_{V \setminus u;-j}(\mathbf{f}) - \frac{1}{\alpha} sf_{u}(y_{u,j}) \right] \\
&= \mathbb{E}_{\sigma_{u,j}} \Bigg[ \sup_{\substack{
P^{\alpha}\left[\mathbf{f}-\mathbf{r}^{\alpha}\right]^2 \leq \rho \\ \mathbf{f} \in \mathcal{F}_G}} U_{V \setminus u;-j}(\mathbf{f}) + \frac{1}{\alpha}\sigma_{u,j} f_{u,j}(y_{u,j}),
\Bigg],
\end{aligned}    
\end{equation*}
where in the last inequality, we have used the definition of $\sigma_{u,j}$. As the inequality is satisfied for all $\epsilon>0$, we have
\begin{equation}
\mathbb{E}_{\sigma_{u,j}} \Bigg[ \sup_{\substack{
P^{\alpha}\left[\mathbf{f}-\mathbf{r}^{\alpha}\right]^2 \leq \rho \\ \mathbf{f} \in \mathcal{F}_G}} U_{V \setminus u;-j}(\mathbf{f}) + \sigma_{u,j} f_{u,j}(y_{u,j}) r^{\alpha}_u(y_{u,j})
\Bigg]  \leq \mathbb{E}_{\sigma_{u,j}} \Bigg[ \sup_{\substack{
P^{\alpha}\left[\mathbf{f}-\mathbf{r}^{\alpha}\right]^2 \leq \rho \\ \mathbf{f} \in \mathcal{F}_G}} U_{V \setminus u;-j}(\mathbf{f}) + \frac{1}{\alpha}\sigma_{u,j} f_{u,j}(y_{u,j}),
\Bigg] 
\end{equation}
We can use the same argument for all the remaining $\sigma_{v,i}$ for $v \neq u$, $i \neq j$, which leads to the result of the lemma.
\end{proof}

The following result identifies the vector-valued function class satisfying the hypotheses of \Lemma{lemma:Bbclass}. 
\begin{lemma}{\label{lemma:Bbclass}}
Let us define the class of functions: 
\begin{equation}{\label{eq:H_class}}
\mathcal{H}_{\Ghilbert}=\{ h_\mathbf{f}=(h_{f_1},...,h_{f_\Gdims}), h_{f_v}:(x_v,x'_v) \rightarrow \ell_v(f_v)(x_v,x'_v)-\ell_v(r_v^{\alpha})(x_v,x'_v), \mathbf{f} \in \mathcal{F}_G\},
\end{equation}
which satisfies the following points: 
\begin{enumerate}[topsep=0.4em, itemsep=0.em]
\item  $\max_{v \in V} \sup_{(x,x') \in \mathcal{X}} \abs{h_{f_v}(x,x')} \leq B_1$. 
\item  $\mathcal{H}_{\Ghilbert}$ is a ($\beta$,$B$)-Bernstein class with $\beta=1$  and $B=B_0:= \frac{1}{2}\left((b+\frac{1}{\alpha})^2 +\frac{4}{\alpha} \right)$.
\item  Let us define the following MTLRCs: 
\begin{equation}
\begin{aligned}
\mathcal{R}(\mathcal{H}_{\Ghilbert},\rho) 
&= \mathbb{E}_{z,\sigma} \left[  
\sup_{V(h_{\mathbf{f}}) \leq \rho, \mathbf{f} \in \mathcal{F}_{G}} \frac{1}{n \Gdims}
\sum_{v \in V} \sum_{i=1}^n \sigma_{v,i} h_{f_v}(x_{v,i},x'_{v,i}),
\right] \\
\mathcal{R}(\mathcal{F}_{G},\rho) &=\mathbb{E}_{p^{\alpha},\!\sigma}  \Bigg[
\sup_{\substack{
P^{\alpha}\mathbf{f} ^2 \leq \rho \\ \mathbf{f} \in \mathcal{F}_G}} \frac{1}{n \Gdims}
\sum_{v \in V} \sum_{i=1}^n \sigma_{v,i} f_v(y_{v,i}) \Bigg].
\end{aligned}
\end{equation}
Then, the following inequality is satisfied: 
\begin{equation}
\mathcal{R}(\mathcal{H}_{\Ghilbert},\rho) \leq   2 \left(b+\frac{1}{\alpha} \right) \mathcal{R}\left(\mathcal{F}_{G},\frac{\rho}{2B_0}\right).
\end{equation}
\end{enumerate}
\end{lemma}
\begin{proof} \mbox{}\\
\inlinetitle{First point}{:}%
\begin{equation*}{\label{eq:uniform_bound_G_f}}
\begin{aligned}
\max_{v \in V} \sup_{(x,x') \in \mathcal{X}} \abs{h_{f_v}(x,x')} &= \max_{v \in V} \sup_{(x,x') \in \mathcal{X} \times \mathcal{X} } \abs{\ell_v(f_v)(x,x')-\ell_v(r_v^{\alpha})(x,x')} \\
&  \leq \max_{v \in V} \sup_{(x,x') \in \mathcal{X} \times \mathcal{X}}  
\frac{(1-\alpha)}{2} \abs{\left[f^2_v-(r_v^{\alpha})^2\right](x)} + \frac{\alpha}{2} \abs{\left[f^2_v-(r_v^{\alpha})^2\right](x')} + \abs{\left[f_v-r_v^{\alpha}\right](x')}  \\
& =\frac{1}{2} (b+\frac{1}{\alpha})^2 +
(b+\frac{1}{\alpha}) := B_1. \qquad (\text{\tiny{\Expr{eq:inf_norm_sum_f_r}}})
\end{aligned}
\end{equation*}

\inlinetitle{Second point}{:}
Due to the properties listed in \Lemma{lemma:why_pearson}, we have the following 
inequalities: 
\begin{equation}{\label{eq:bernstein_class_proof}}
\begin{aligned}
P_{z}\left[h_{\mathbf{f}}\right]^2&=P_{z}\left[\ell(\mathbf{f})-\ell(\mathbf{r^\alpha})\right]^2  \leq  \frac{B_0}{2} P^{\alpha}\left[\mathbf{f}-\mathbf{r^\alpha}\right]^2 = B_0 P_z \left[ \ell(\mathbf{f})-\ell(\mathbf{r^\alpha}) \right],
\end{aligned}
\end{equation}
which means $\mathcal{H}_{\Ghilbert}$ is a $(\beta,B)$-Bernstein class of vector-value functions, with $\beta=1$ and $B=B_0$, and the function controlling the variance of the class 
 is defined as: $V(h_{\mathbf{f}})= \frac{B_0}{2} P^{\alpha}(\mathbf{f}-\mathbf{r^\alpha})^2$. 

\inlinetitle{Third point}{:} Lets fix $\rho \in \R^{+}$, then we can verify:
\begin{equation*}
\begin{aligned}
& B\mathcal{R}(\mathcal{H}_{\Ghilbert},\rho)  = B \mathbb{E}_{z,\sigma} \left[  
\sup_{V(h_{\mathbf{f}}) \leq \rho, \mathbf{f} \in \mathcal{F}_G} \frac{1}{n \Gdims}
\sum_{v \in V} \sum_{i=1}^n \sigma_{v,i} \ell_v(f_v)(x_{v,i},x'_{v,i})
\right] \qquad (\text{\tiny{\Eq{eq:MT-LRC}}})  \\
& = B \mathbb{E}_{z,\sigma} \left[  
\sup_{V(G_{\mathbf{g}}) \leq r, \mathbf{f} \in \mathcal{F}_{G}} \frac{1}{n \Gdims}
\sum_{v \in V} \sum_{i=1}^n \sigma_{v,i} \left( \frac{(1-\alpha)}{2}f^2_v(x_{v,i}) +
\frac{\alpha}{2} f^2_v(x'_{v,i}) -f_v(x'_{v,i}) \right)
\right]  \\
& \leq B \mathbb{E}_{z,\sigma} \left[
\sup_{V(h_{\mathbf{f}}) \leq \rho, \mathbf{f} \in \mathcal{F}_{G}} \frac{1}{n \Gdims}
\sum_{v \in V} \sum_{i=1}^n \sigma_{v,i} \left( \frac{(1-\alpha)}{2}  f^2_v(x_{v,i}) +
\frac{\alpha}{2}  f^2_v(x'_{v,i}) \right) \right] \\
& + B \mathbb{E}_{z,\sigma} \left[  
\sup_{V(h_{\mathbf{f}}) \leq r, \mathbf{f} \in \mathcal{F}_{G}} \frac{1}{n \Gdims}
\sum_{v \in V} \sum_{i=1}^n  \sigma_{v,i}  f_v(x'_{v,i})  
\right] \ \ \ \
(\text{\tiny{Subadditivity of the supremum and symmetry of the Redemacher variables}})  \\ 
&=  B \mathbb{E}_{p^{\alpha}\!\!,\sigma} \left[
\sup_{V(h_{\mathbf{f}}) \leq \rho, \mathbf{f} \in \mathcal{F}_{G}} \frac{1}{n \Gdims}
\sum_{v \in V} \sum_{i=1}^n \frac{1}{2} \sigma_{v,i}  f^2_v(y_{v,i}) \right] + \mathbb{E}_{p^{\alpha}\!\!,\sigma} \left[  
\sup_{V(h_{\mathbf{f}}) \leq r, \mathbf{f} \in \mathcal{F}_{G}} \frac{1}{n \Gdims}
\sum_{v \in V} \sum_{i=1}^n \sigma_{v,i}  f_v(y_{v,i}) r^{\alpha}_v(y_{v,v})  
\right],
\end{aligned}
\end{equation*}
\noindent where the last expression is a consequence of $\Expecalphav[h(y)]=(1-\alpha) \Expecnv[h(x)] +\alpha\Expecav[h(x')]$ and $\Expecalphav[f(y) r^{\alpha}(y)]=\Expecav[g(x')]$. Notice, $x^2$ is a Lipschitz function with Lipschitz constant $2b$ when $x \in [-b,b]$. We can apply the Contraction property of Rademacher Complexity, which holds for Local Rademacher Complexities for vector-valued class of functions (Theorem\,17 in \cite{Maurer2006}). This result leads to the inequality:
\begin{equation*}
\mathbb{E}_{p^{\alpha}\!\!,\sigma} \left[
\sup_{V(h_{\mathbf{f}}) \leq \rho, \mathbf{f} \in \mathcal{F}_{G}} \frac{1}{n \Gdims}
\sum_{v \in V} \sum_{i=1}^n \sigma_{v,i}  f^2_v(y_{v,i}) \right] \leq  2 b \mathbb{E}_{p^{\alpha}\!\!,\sigma} \left[
\sup_{V(h_{\mathbf{f}}) \leq \rho, \mathbf{f} \in \mathcal{F}_{G}} \frac{1}{n \Gdims}
\sum_{v \in V} \sum_{i=1}^n \sigma_{v,i}  f_v(y_{v,i}) \right] 
\end{equation*}
By combining this inequality and \Lemma{lemma:MTLCR_alpha_bound} we obtain:
\begingroup
\allowdisplaybreaks%
\begin{equation}
\begin{aligned}
B_0 \mathcal{R}(\mathcal{H}_{\Ghilbert},\rho)  
& \leq B_0 \left(b+\frac{1}{\alpha} \right)\mathbb{E}_{p^{\alpha}\!\!,\sigma} \left[
\sup_{V(h_{\mathbf{f}}) \leq \rho, \mathbf{f} \in \mathcal{F}_{G}} \frac{1}{n \Gdims}
\sum_{v \in V} \sum_{i=1}^n \sigma_{v,i}  f_v(y_{v,i}) \right] \\
&= B_0 \left(b+\frac{1}{\alpha} \right)\mathbb{E}_{p^{\alpha}\!\!,\sigma} \left[
\sup_{\substack{
\frac{B_0}{2}P^{\alpha}\left[\mathbf{f}-\mathbf{r}^{\alpha}\right]^2 \leq \rho \\ \mathbf{f} \in \mathcal{F}_G}} \frac{1}{n \Gdims}
\sum_{v \in V} \sum_{i=1}^n \sigma_{v,i}  f_v(y_{v,i}) \right] \\
&=  B_0 \left(b+\frac{1}{\alpha} \right)\mathbb{E}_{p^{\alpha}\!\!,\sigma}  \left[
\sup_{\substack{
\frac{B_0}{2}P^{\alpha}\left[\mathbf{f}-\mathbf{r}^{\alpha}\right]^2 \leq \rho \\ \mathbf{f} \in \mathcal{F}_G}} \frac{1}{n \Gdims}
\sum_{v \in V} \sum_{i=1}^n \sigma_{v,i} [ f_v(y_{v,i}) -r_v^{\alpha}(y_{v,i})] \right] \quad (\text{\tiny{By independence of $\sigma$ and $\mathbf{Y}$.}}) \\
&\leq  B_0 \left(b+\frac{1}{\alpha} \right)\mathbb{E}_{p^{\alpha}\!\!,\sigma}  \left[
\sup_{\substack{
\frac{B_0}{2}P^{\alpha}\left[\mathbf{f}-\mathbf{g} \right]^2 \leq \rho \\ \mathbf{f},\mathbf{g} \in \mathcal{F}_G}} \frac{1}{n \Gdims}
\sum_{v \in V} \sum_{i=1}^n \sigma_{v,i} [ f_v(y_{v,i}) -g_v(y_{v,i})] \right]  \\ 
& = B_0 \left(b+\frac{1}{\alpha} \right)\mathbb{E}_{p^{\alpha}\!\!,\sigma}  \left[
\sup_{\substack{
2B_0P^{\alpha}\mathbf{f} ^2 \leq \rho \\ \mathbf{f} \in \mathcal{F}_G}} \frac{1}{n \Gdims}
\sum_{v \in V} \sum_{i=1}^n \sigma_{v,i} f_v(y_{v,i}) \right] = 2 B_0 \left(b+\frac{1}{\alpha} \right) \mathcal{R}(\mathcal{F}_{G},\frac{\rho}{2B_0}).
\end{aligned}
\end{equation}
\endgroup
In the last inequality we have used the symmetry of the Rademacher variables and the fact $f \in \mathcal{F}_{G}$ is symmetric and convex. 
\end{proof}

\Lemma{lemma:rademacher_and_root_functions} implies that $2 B_0 \left(b+\frac{1}{\alpha} \right) \mathcal{R}(\mathcal{F}_{G},\frac{\rho}{2B_0})$ is sub-root function. The goal now is to upperbound its fixed point $\rho^*$. This point requires us to exploit the properties of the graph regularization and the capacity condition associated with the covariance operators $\{\CO_v\}_{v \in V}$. A big part of this analysis has been already done in \cite{Yousefi2018}. We rewrite the most relevant results of this work and rework the upperbounds to obtain more clear expressions adapted to our problem.

\begin{theorem}{\label{thm:Theorem11_Yousefi}}
(Theorem\,11 in \cite{Yousefi2018})
    Let the regularizer be $\norm{\mathbf{f}}^2_{G}$ as defined in \Eq{eq:penalized_term_G}, and denote its dual norm  by $\norm{\cdot}_{*}$. Let the kernels be uniformly bounded, and define the sample $\mathbf{Y} = \{\mathbf{Y}_v\}_{v\in V}= \big\{\{y_{v,1},...,y_{v,\Npre}\}\big\}_{v \in V}$ and where $\forall v \in V$ $\{y_{v,1},...,y_{v,\Npre}\}$ is a i.i.d sample drawn from $p^{\alpha}_{v}$. Assume that for each $v \in V$, the associated covariance operator admits an eigenvector decomposition $\CO_v=\ExpecAlphav{\varphi(y)\otimes\varphi(y)} = \sum_{i \in \N} \mu_{v,i} \tilde{\varphi}_{v,i} \otimes \tilde{\varphi}_{v,i}$, where $\{\tilde{\varphi}_{v,i}\}_{v \in V}$ forms an orthonormal basis of $\Hilbert$ and $\{\mu_{v,i}\}_{i=1}^{\infty}$ are the corresponding eigenvalues in non-increasing order. Then, for any given positive operator $\mathcal{D}$ on $\R^{\Gdims}$, any $\rho>0$ and any non-negative integers $h_1,...,h_{\Gdims}$: 
\begin{equation}
    R(\mathcal{F}_{G},\rho) \leq 
   \sqrt{ \frac{\rho \sum_{v \in V} h_v}{n \Gdims} } + \frac{\sqrt{2} \Lambda}{\Gdims}   \mathbb{E}_{p^{\alpha}\!\!,\sigma}\left[\norm{\mathcal{D}^{-\frac{1}{2}} \mathbf{V}}_* \right] ,
\end{equation}
where $\mathbf{V}=\left\{\sum_{j>h_v} \dotH{\frac{1}{n}\sum_{i=1}^{n} \sigma_{v,i} \phi(y_{v,i})}{\tilde{\varphi}_{v,j}} \tilde{\varphi}_{v,j}\right\}_{v\in V}$.
\end{theorem}
\begin{lemma}{\label{thm:Corollary_Yousefi}}
(Expr.\,C.9 of Corollary\,22 \citep{Yousefi2018})
Under the hypotheses of the previous theorem, we have the following inequality: 
\begin{equation}
\mathbb{E}_{p^{\alpha}\!\!,\sigma}\left[\norm{\mathcal{D}^{-\frac{1}{2}} \mathbf{V}}_* \right] \leq \sqrt{\frac{1}{n} \sum_{v \in V} \Big|\mathcal{D}^{-1}_{vv} \sum_{j >h_v}   \mu_{v,j}\Big|},
\end{equation}
where $\{D^{-1}_{vv}\}_{v \in V}$ are the diagonal elements of $\mathcal{D}^{-1}$.
\end{lemma}
\begin{lemma}{\label{lemma:rhofunction}} If Assumptions\,\ref{ass: independence}-\ref{ass:model_definition} are satisfied and that $\mathcal{F}_G$ is a class of functions with ranges in $[-b,b]$, then  $2B_0(b+\frac{1}{\alpha}) \mathcal{R}(\mathcal{F}_{G},\frac{\rho}{2B_0})$  is a sub-root function whose fixed point $\rho^*$ is upperbounded by: 
\begin{equation}
\rho^* \leq 8B_0\sqrt{ \frac{\zeta^{*}+1}{\zeta^{*}-1}} \left[\left(b+\frac{1}{\alpha}\right)^{\zeta_{\min}} \Lambda^2 \mathcal{D}^{-1}_{\max}\right]^{\frac{1}{1+\zeta^{*}}}   n^{\frac{-\zeta^*}{1+\zeta^{*}}} N^{\frac{-1}{1+\zeta^{*}}} s_{\max}^{\frac{1}{1+\zeta^{*}}}, 
\end{equation}
where $\zeta^*=\min_{v \in V} \zeta_{v}$, $s_{\max}=\max_{v \in V} s_v$, and $\mathcal{D}^{-1}_{max}= \max_{v \in V} \mathcal{D}^{-1}_{vv}=\abs{(\Lpc+\gamma)^{-1}_{vv}}$.
   \end{lemma}

\begin{proof}
Combining \Theorem{thm:Theorem11_Yousefi} and \Lemma{thm:Corollary_Yousefi} lead us to the following inequality: 
\begin{equation}
\begin{aligned}
2 B_0 (b+\frac{1}{\alpha})  \mathcal{R}(\mathcal{F}_{G},\frac{\rho}{2B_0}) &\leq 2 B_0 (b+\frac{1}{\alpha}) \left[  \sqrt{  \frac{\rho \sum_{v \in V} h_v}{2B_0n \Gdims} } +   \sqrt{\frac{2 \Lambda^2}{\Gdims^2 n} \sum_{v \in V} \bigg|\mathcal{D}^{-1}_{vv} \sum_{j >h_v}   \mu_{v,j}\bigg|} \right] \\
&= \sqrt{2B_0 \left(b+\frac{1}{\alpha}\right)^2   \frac{\rho \sum_{v \in V} h_v}{n \Gdims}} +  2 B_0 \left(b+\frac{1}{\alpha}\right)  \sqrt{\frac{2 \Lambda^2}{\Gdims^2 n} \sum_{v \in V} \bigg|\mathcal{D}^{-1}_{vv} \sum_{j >h_v}   \mu_{v,j}\bigg|}  \\
& = \sqrt{a_1 \rho} +a_2, 
\end{aligned}
\end{equation}
where in the last expression we have introduced the variables: 
\begin{equation}
    a_1= 2B_0 \left(b+\frac{1}{\alpha}\right)^2 \bigg(\frac{\sum_{v \in V} h_v}{n \Gdims}\bigg),  \ \ \ a_2= 2 B_0 (b+\frac{1}{\alpha})  \sqrt{\frac{2 \Lambda^2}{\Gdims^2 n} \sum_{v \in V} \bigg|\mathcal{D}^{-1}_{vv} \sum_{j >h_v}   \mu_{v,j}\bigg|}. 
\end{equation}

Now, we will look for the solution to the equation $\sqrt{a_1\rho}+a_2=\rho$, which is equivalent to solve $\rho^2- (a_1+2a_2)\rho + a_2^2=0$, that is 
\begin{equation}
    \rho = \frac{(a_1+2a_2)\pm \sqrt{a_1^2+4a_2}}{2} \leq a_1+2a_2.
\end{equation}
As $\rho^*$ is the fixed point of $2 B_0 (b+\frac{1}{\alpha})  \mathcal{R}(\mathcal{F}_{G},\frac{\rho}{2B_0})$, then by \Lemma{lemma:Lemma32Barlett}, we have: 
\begin{equation}
    \rho^* \leq \rho \leq a_1+2a_2.
\end{equation}
The goal now is to upperbound both terms $a_1$ and $a_2$ by exploiting \Assumption{ass:capacitycondition}.
Observe that by the capacity condition of \Assumption{ass:capacitycondition}, we have:
\begin{equation*}
    \sum_{j>h_v} \mu_{v,j} \leq \sum_{j>h_v} s_v^2 j^{-\zeta_v} \leq s_v \int_{h_v}^{\infty} x^{-\zeta_v} dx = -\frac{s_v}{1-\zeta_v} h_{v}^{1-\zeta_v},
\end{equation*}
which implies:\quad
%
$\displaystyle a_2 \leq  2 B_0 \left(b+\frac{1}{\alpha}\right)  \sqrt{\frac{- 2 \Lambda^2}{\Gdims^2 n} \sum_{v \in V} \abs{\mathcal{D}^{-1}_{vv}}\frac{s_v}{1-\zeta_v} h_{v}^{1-\zeta_v}}$.

\noindent Moreover, by the Cauchy–Schwarz inequality: 
\begin{equation*}
    a_1 \leq 2 B_0 \left(b+\frac{1}{\alpha}\right)^2 \sqrt{\Gdims} \sqrt{ \frac{\sum_{v \in V} h^2_v}{(n \Gdims)^2} }=  B_0 (b+\frac{1}{\alpha})^2 \sqrt{ \frac{\sum_{v \in V} h^2_v}{n^2 \Gdims}}.
\end{equation*}
After putting together both inequalities, we get: 
\begin{equation*}
\begin{aligned}
\rho^* &\leq  a_1+2a_2 \\
   &\leq 2 B_0 \left(b+\frac{1}{\alpha}\right) \left[ \sqrt{(b+\frac{1}{\alpha})^2 \left(\frac{\sum_{v \in V} h^2_v}{n^2 \Gdims}\right)} + \sqrt{\frac{- 8 \Lambda^2}{\Gdims^2 n} \sum_{v \in V} \abs{\mathcal{D}^{-1}_{vv}}\frac{s_v}{1-\zeta_v} h_{v}^{1-\zeta_v}} \right] \\
   &\leq 2 B_0 \left(b+\frac{1}{\alpha}\right)\sqrt{\sum_{v \in V}   2(b+\frac{1}{\alpha})^2 \left(\frac{h^2_v}{n^2 \Gdims}\right) -  \frac{16 \Lambda^2}{\Gdims^2 n} \abs{\mathcal{D}^{-1}_{vv}}\frac{s_v}{1-\zeta_v} h_{v}^{1-\zeta_v}}  \\
   & = 2 B_0 \left(b+\frac{1}{\alpha}\right) \sqrt{\sum_{v \in V} c h^2_v - c_v h_{v}^{1-\zeta_v}},
\end{aligned}
\end{equation*}
where
\begin{equation*}
c= \frac{2 (b+\frac{1}{\alpha})^2}{n^2N}, \ \ \ c_v= \frac{16 \Lambda^2}{\Gdims^2 n} \abs{\mathcal{D}^{-1}_{vv}}\frac{s_v}{1-\zeta_v}.
\end{equation*}
By taking the partial derivative \wrt $h_v$ and setting it to zero, yields to an optimal value: 
\begin{equation}
   h_v^*= \left( \frac{(1-\zeta_v)c_v}{2c} \right)^{\frac{1}{1+\zeta_v}}= \left( \frac{4 \Lambda^2\abs{\mathcal{D}^{-1}_{vv}} s_v n }{(b+\frac{1}{\alpha})^{2} N} \right)^{\frac{1}{1+\zeta_v}}.
\end{equation}
Then, after substitution: 
\begin{equation}
\begin{aligned}
\rho^* & \leq 2 B_0 (b+\frac{1}{\alpha}) \sqrt{\sum_{v \in V} c (h^*_v)^2 - c_v (h^*_{v})^{1-\zeta_v}} \\
& =  2 B_0 \left(b+\frac{1}{\alpha}\right) \sqrt{\sum_{v \in V}  (h^*_v)^2 \left(c - \frac{2c}{(1-\zeta_v)}\right)} \\
& =  2 B_0 \left(b+\frac{1}{\alpha}\right) \sqrt{c \sum_{v \in V}
 \left( \frac{\zeta_v+1}{\zeta_v-1} \right) (h_v^*)^2}.
\end{aligned}
\end{equation}
If we denote $\zeta^*=\min_{v \in V} \zeta_{v}$ ,$s_{\max}=\max_{v \in V} s_v$ and $\Lpc^{-1}_{max}= \max_{v \in V} \abs{\mathcal{D}^{-1}_{vv}}=\abs{(\Lpc+\gamma)^{-1}_{vv}}$:
\begin{equation}
\begin{aligned}
\rho^* & \leq 8 B_0\frac{(b+\frac{1}{\alpha})^2}{n} \sqrt{ \frac{\zeta^*+1}{\zeta^*-1}}
 \left( \frac{\Lambda^2\Lpc^{-1}_{max} s_{\max} n }{(b+\frac{1}{\alpha})^{2} N} \right)^{\frac{1}{1+\zeta^{*}}} \\
&  = 8 B_0\sqrt{ \frac{\zeta^{*}+1}{\zeta^{*}-1}} \left[\left(b+\frac{1}{\alpha}\right)^{2\zeta^{*}} \Lambda^2 \Lpc^{-1}_{\max}\right]^{\frac{1}{1+\zeta^{*}}}   n^{\frac{-\zeta^*}{1+\zeta^{*}}} N^{\frac{-1}{1+\zeta^{*}}} s_{\max}^{\frac{1}{1+\zeta^{*}}}.
\end{aligned}
\end{equation}
\end{proof}

\subsection{Proof \Theorem{thm:convergence_results}} 
\begin{proof}
\Lemma{lemma:Bbclass} implies that $\mathcal{H}_{\Ghilbert}$ is a $(\beta,B)$-Bernstein class of vector-valued functions with $\beta=1$ and $B=B_0$, and  $\max_{v \in V} \sup_{(x,x') \in \mathcal{X}} \abs{h_{f_v}(x,x')} \leq B_1$. By \Lemma{lemma:rhofunction} we have that there exists a sub-root function such that $B\mathcal{R}(\mathcal{H}_{\Ghilbert},\rho) \leq \rhofunction(\rho)$. Then, the hypotheses of \Theorem{thm:TheoremB3_Yousefi} are satisfied, which implies that with probability at least $1-\delta$, every $f \in \Ghilbert$ satisfies:
\begin{equation}{\label{eq:convergence_loss_function}}
\begin{aligned}
&\frac{1}{\Gdims} \sum_{v \in V}\Expecjointv\left[\ell_v(f_v)(z)-\ell_v(r_v^{\alpha})(z)\right]\\  & 
\leq \frac{B_1}{B_1-1} \left[ \frac{1}{N} \sum_{v \in V} \left( \frac{1-\alpha}{n} \sum_{i=1}^{n} \frac{\left[f^2_v-(r^{\alpha}_v)^2\right](x_{v,i})}{2} +  \frac{\alpha}{n} \sum_{i=1}^{n} \frac{\left[f^2_v-(r^{\alpha}_v)^2\right](x'_{v,i})}{2}  -\frac{1}{n}\sum_{i=1}^{n} \left[f_v-r^{\alpha}_v\right](x'_{v,i})  \right) \right] \\
&+ 2C(20^2) B_1  \rho^* + \frac{16 B_0^2 C}{n \Gdims} \log{\left(\frac{1}{\delta}\right)} + \frac{24B_0B_1}{n \Gdims} \log{\left(\frac{1}{\delta}\right)}.
\end{aligned}
\end{equation}
In particular for the minimum $\hat{\mathbf{f}}$ of problem \ref{eq:problem_collaborative_FG} we have that the term involving the empirical expectations is less than zero. 

As detailed in \Appendix{appendix:PE_LRE_link}, we can easily verify that $\PE{}(p^{\alpha} \Vert \q)= \Expecjointv\left[-\ell_v(r_v^{\alpha})(z)\right]-\frac{1}{2}$, and by \Expr{eq:PE_div_f_v} $\PE{}^{\alpha}_v(f_v)=\Expecjointv\left[-\ell_v(f_v)(z)\right]-\frac{1}{2}$. Then for $\hat{\mathbf{f}}$ we can rewrite \Expr{eq:convergence_loss_function} as: 
\begin{equation*}
\begin{aligned}
&\frac{1}{\Gdims} \sum_{v \in V} \left[ \PE(p^{\alpha}_v \Vert \q_v)-\PE{}^{\alpha}_v(\hat{f}_v) \right]
\leq  2C(20^2) B_1  \rho^* + \frac{16 B_0^2 C}{n \Gdims} \log{\left(\frac{1}{\delta}\right)} + \frac{24B_0B_1}{n \Gdims} \log{\left(\frac{1}{\delta}\right)}
\end{aligned}
\end{equation*}
Alternatively, after applying the second point of \Lemma{lemma:why_pearson} we can conclude:
\begin{equation*}
\begin{aligned}
\frac{1}{\Gdims} \sum_{v \in V} \Expecalphav\left[ \left[\hat{f}_v-r^{\alpha}_{v}\right]^2(y)  \right] &\leq   4C(20^2) B_1  \rho^* + \frac{32 B_0^2 C}{n \Gdims} \log{\left(\frac{1}{\delta}\right)} + \frac{48B_0B_1}{n \Gdims} \log{\left(\frac{1}{\delta}\right)},
\end{aligned}
\end{equation*}%
where \Lemma{lemma:rhofunction} implies: 
\begin{equation*}
\rho^* \leq 8B_0\sqrt{ \frac{\zeta^{*}+1}{\zeta^{*}-1}} \left[(b+\frac{1}{\alpha})^{2\zeta^{*}} \Lambda^2 \Lpc^{-1}_{\max}\right]^{\frac{1}{1+\zeta^{*}}}   n^{\frac{-\zeta^*}{1+\zeta^{*}}} N^{\frac{-1}{1+\zeta^{*}}} s_{\max}^{\frac{1}{1+\zeta^{*}}}.
\end{equation*}  
\end{proof}

{\small \bibliography{GRULSIF}}

\newpage

\end{document}